\RequirePackage{fix-cm}
\documentclass[twocolumn]{svjour3}          
\smartqed  
\usepackage{graphicx}
\usepackage{amsmath,amssymb,euscript}
\usepackage{latexsym}
\usepackage{color}
\usepackage{longtable}
\usepackage{cite}  
\newcommand{\Proofend}{\hfill$\diamondsuit$}
\newcommand\n{{\mathbf{n}}}
\newcommand\m{{\mathbf{m}}}
\newcommand\f{{\mathbf{f}}}

\def\NN{{\mathbb N}}
\def\RR{{\mathbb R}}

\def\sspan{\textrm{span}}

\newcommand{\PSNR}{\rm{PSNR}}
\newcommand{\MSE}{\rm{MSE}}
\newcommand{\SSIM}{\rm{SSIM}}
\newcommand{\cov}{\rm{cov}}
\begin{document}
\title{Image scaling by de la Vall\'ee-Poussin filtered interpolation
}

\author{Donatella Occorsio \and Giuliana Ramella* \and Woula Themistoclakis}
\institute{D. Occorsio \at
                Department of Mathematics and Computer Science, University of Basilicata, Viale dell'Ateneo Lucano 10, 85100 Potenza, Italy
               \email{donatella.occorsio@unibas.it}           
           \and
         G. Ramella (*corresponding author) \at
         C.N.R. National Research Council of Italy, Institute for Applied Computations  ``Mauro Picone", Via P. Castellino, 111, 80131 Naples, Italy
         \email{giuliana.ramella@cnr.it}
 	\and
         W. Themistoclakis \at
        C.N.R. National Research Council of Italy, Institute for Applied Computations  ``Mauro Picone", Via P. Castellino, 111, 80131 Naples, Italy
          \email{woula.themistoclakis@cnr.it}
}
\date{29 June 2022}
\maketitle
\begin{abstract}
We present a new image scaling method both for downscaling and upscaling, running with any scale factor or desired size.
The resized image is achieved by sampling  a bivariate polynomial which globally interpolates the data at the new scale. The method's particularities lay in both the sampling model and the interpolation polynomial we use. Rather than classical uniform grids, we consider an unusual sampling system based on Chebyshev zeros of the first kind. Such optimal distribution of nodes permits to consider near--best  interpolation polynomials defined by a filter of de la Vall\'ee Poussin type. The action ray of this filter provides an additional parameter that can be suitably regulated to improve the approximation.  The method has been tested on a significant number of different image datasets. The results are evaluated in qualitative and quantitative terms and compared with other available competitive methods. The perceived quality of the resulting scaled images is such that important details are preserved, and the appearance of artifacts is low. Competitive quality measurement values, good visual quality, limited computational effort, and moderate memory demand make the method suitable for real-world applications.
\keywords{Image downscaling \and Image upscaling \and de la Vall\'ee-Poussin Interpolation \and Chebyshev nodes }

\subclass{94A08 \and MSC 68U10 \and 65D05 \and 62H35}
\end{abstract}
\section{Introduction}
Image scaling aims to get the image at a different size preserving the original content as much as possible, with minor loss of quality,  in two opposite ways: downscaling and upscaling.
Downscaling is a compression process by which the size of the high-resolution (HR) input image  is reduced to recover the low-resolution  (LR) target image. Conversely, upscaling is an enhancement process in which the size of the LR input image  is enlarged  to regain the HR target image.

Image scaling (also termed image resampling or image resizing) is a widely used tool in several fields such as medical imaging, remote sensing, gaming, electronic publishing, autonomous driving, and aerial photography
 \cite{Atkinson, Meijering, App1, App2, App3, App4}. For example, upscaling allows highlighting of important details of the image in remote sensing and medical applications \cite{Atkinson, Meijering}, while downscaling is a fundamental operation for fast browsing or sharing purposes \cite{App1, App2}. Other applicative examples regard scenarios like deforestation, monitoring, traffic, surveillance, and many other engineering tasks. Sometimes image scaling is used for illicit purposes, e.g., to automatically generate camouflage images whose visual semantics change dramatically after scaling \cite{Xiao}. In these cases, it is very important to detect the scaling effects in order to defend against such attacks and adopt suitable countermeasures \cite{Lin, Bruni}.

From a computational point of view, image scaling can be addressed by different numerical methods (see Section 2), whose main critical points typically are:  a) undesired effects, such as ringing artifacts and aliasing, due to the increase/decrease in the number of pixels which introduces/reduces information to the image; b) computational efficiency in performing the resampling task in real-world applications. Moreover, most existing methods treat the resampling in only one direction since downscaling and upscaling are often considered separate problems in literature \cite{scaling1}.

We aim to propose a scaling method that works in both downscaling and upscaling directions. To this aim, looking at the scaling problem as an approximation problem, we employ an interpolation polynomial based on an {\it adjustable} filter of de la Vall\'ee Poussin (briefly VP) type, which can be suitably modulated to improve the approximation (see, e.g., \cite{Them_Van_Barel,Occo_Them_LNCS,Themistoclakis_2011}).

 Indeed, the VP type interpolation has been introduced in literature as a valid alternative to Lagrange interpolation to provide a better pointwise approximation, especially when the Gibbs phenomenon occurs \cite{Occo_Them_LNCS,Occo_Them_AMC,Occo_Them_Apnum}. In fact, an interesting feature of VP filtered approximation is the presence of a free additional degree--parameter, which is responsible for the localization degree of the polynomial interpolation basis (the so--called  fundamental VP polynomials) around the nodes.
By changing this parameter, we may modulate the typical oscillatory behavior of the fundamental Lagrange polynomials according to the data, improving the approximation without destroying the interpolation property and keeping fixed the number of the interpolation nodes.
Moreover, it is also worth noting that VP interpolation can be embedded in a wavelet scheme with decomposition and reconstruction algorithms very fast since based on fast cosine transforms \cite{CaThwave}.

From a theoretical point of view, the literature concerning VP filtered approximation provides many convergence theorems, also in the uniform norm. They estimate an error comparable with the error of the best polynomial approximation \cite{Themistoclakis_2011,WoulaL1} and allow to predict the convergence order from the regularity of the function to approximate \cite{Occo_Them_DRNA}.
Due to such nice behavior, VP approximation has been usefully applied as a demonstration tool to carry out proofs of different theorems \cite{mastrorussothem,Them_gauss,Mastro_them_acta,OccoRusso2011,prandtl_occo_db}.

From a more applicative point of view, it has been used to solve singular integral and integro-differential equations \cite{Air_ThMastro}, \cite{Prandtl_nostra} or derive good quadrature rules for the finite Hilbert transform \cite{Hilbert_MG,Hilbert_MG2}. However, to our knowledge, it has never been applied to Image Processing. Hence, the present paper represents the first step in investigating how the VP interpolation scheme can be usefully employed in image scaling.

To explain the proposed scaling method (shortly denoted by VPI method or simply by VPI), as a starting point, we consider that the input RGB image is represented at a continuous scale by a vectorial function (with separate channels for each color) whose sampling yields the pixels values. We globally approximate such function using suitable VP interpolation polynomials, modulated by a free parameter $\theta\in ]0,1]$ \cite{Themistoclakis_2011, Them_Van_Barel, Occo_Them_DRNA, Occo_Them_Apnum, Occo_Them_AMC}. Hence, we get the resized image by evaluating such VP polynomials in a denser (upscaling) or coarser (downscaling) grid of sampling points. 

Being designed both for downscaling and upscaling, VPI method is flexible and implementable for any scale factor. The rescaling can be obtained by specifying the scale factor or, alternatively, the desired size of the image. We point out that, in the following, to distinguish between upscaling and downscaling mode, we use the notation u-VPI and d-VPI, respectively.

Both in upscaling and downscaling, for the limiting parameter choice $\theta=0$, VPI coincides with the LCI method proposed in \cite{Lagrange} and based on classical Lagrange interpolation at the same nodes. Moreover, for any choice of the parameter  $\theta$, d-VPI with odd scale factors also produces the same output resized image of LCI, which results by a direct assignment without any computation. In these cases, if the LR image satisfies the Nyquist-Shannon sampling theorem \cite {Shannon}, d-VPI produces a MSE not greater than input MSE times the scale factor squared. Thus, we can get a null MSE and the best visual quality measures in case of {\it exact} input data (cf. Proposition 1). However, we point out that in cases where the downscaling size violates the sampling theorem, aliasing effects occur. Experiments in the paper also deepen this aspect, and a partial solution is proposed, remaining the problem open to further investigations.

A further contribution of this paper includes a detailed quantitative and qualitative analysis of the obtained results on several publicly available datasets commonly used in Image Processing. The experimental results confirm the effectiveness and utility of employing the VP interpolation scheme, achieving on average a good compromise between visual quality and processing time: the resized images present few blurred edges and artifacts, and the implementation is computationally simple and rather fast.

On average, VPI has a competitive and satisfactory performance, with quality measures generally higher and more stable than other existing scaling methods considered as a benchmark. In general, we have  satisfactory performance, also for high scale factors, compared to the benchmark methods. Specifically, in downscaling, when the free parameter is not equal to zero, VPI improves the LCI performance and results to be more stable than the latter due to the uniform boundedness of Lebesgue constants corresponding to the de la Vall\'ee Poussin type interpolation. Moreover, VPI results much faster than the methods specialized in only downscaling or upscaling. 

At a visual level, VPI captures the object's visual structure by preserving the salient details, the local contrast, and the luminance of the input image, with well--balanced colors and a limited presence of artifacts. To about the same extent as the other benchmark methods, in downscaling VPI exhibits aliasing effects. 

Overall, due to its features, we consider VPI suitable for real--world applications, and, at the same time, we look at it as a complete method because it can also perform upscaling and downscaling with adequate performance.

The remainder of this paper is as follows. In Section \ref{rel_work}, we outline the related work, briefly explaining the benchmark scaling methods we employ in the experimental phase. In Section \ref{math_prel} we provide the mathematical background. In Section \ref{method}, we describe the VPI method and state its main properties. In Section 5 we provide the most relevant implementation details and the qualitative/quantitative evaluations of the experimental results taken over a significant numebr of different image datasets. Finally, conclusions are drawn in Section \ref{concl}.
\section{Related work}
\label{rel_work}
Image scaling has received great attention in the literature of the past decades, during which many methods based on different approaches have been developed. An overview containing pros and cons for some of them can be found in \cite{rev1, rev2}.

Traditionally, image scaling methods are grouped into two categories \cite{book1}: non-adaptive \cite{NN-NIL2, Lanczos, B-splines, Burger, ScSR} and adaptive \cite{Stentiford, Setlur, Ramella2, Zhou, Yang}. In the first category, all the pixels are equally treated. In the second, suitable changes are arranged, depending on image features and intensity values, edge information, texture, etc. The non-adaptive category includes many of the most commonly used algorithms such as the nearest neighbor, bilinear,  bicubic and  B-splines interpolation, Lanczos method \cite{book1, NN-NIL2, Lanczos, B-splines, Burger, CN}. Adaptive methods are designed to maximize the quality of the results. They are also employed in most common approaches such as context-aware computing \cite{Stentiford}, segmentation techniques\cite{Setlur}, and adaptive bilinear schemes \cite{Ramella2}. Machine Learning (ML) methods can be  ascribed to the latter category,even if they are often considered as a separate problem \cite{Zhou, Yang}. The learning paradigm of ML methods aims to compensate for complete (missing) information of the downscaled (upscaled) image using a relationship between HR (LR) and LR (HR) images. Mostly, this paradigm is implemented by a training step, in which the relationship is learned, followed by a step in which the learned knowledge is applied to unseen HR (LR) images.

Usually, non-adaptive scaling methods have problems of blurring or artifacts around edges and only store the low-frequency components of the original image. On the other hand, adaptive scaling methods generally provide better image visual quality and preserve high-frequency components. However, adaptive methods take more computational time as compared to non-adaptive ones. In turn, the ML methods ensure high-quality results but, at the same time, require extensive learning based on a huge number of parameters and labeled training images.

In this section, we limit to describe shortly the methods considered in the validation phase of the VPI method (see  Section 5), namely  DPID \cite{Weber}, L$_0$ \cite{Liu}, SCN \cite{SCN}, LCI \cite{Lagrange} and BIC \cite{BIC}. The source code of such methods is made available by the authors themselves in a common language (Matlab). Except for BIC and LCI, these methods are designed and tested considering the problem of resizing in one direction, i.e., in downscaling (DPID and L$_0$) or upscaling mode (SCN).

DPID is based on the assumption that the Laplacian edge detector and adaptive low-pass filtering can be useful tools to approximate the behavior of the Human Visual System. Important details are preserved in the downscaled image by employing convolutional filters and by selecting the input pixels that contribute more to the output image the more their color deviates from their local neighborhood.

In L$_0$, an optimization framework for image downscaling, focusing on two critical issues, is proposed: salient features preservation and downscaled image construction. Accordingly, two L$_0$-regularized priors are introduced and applied iteratively until the objective function is verified. The first, based on gradient ratio, allows preserving the most salient edges and the visual perceptual  properties of the original image. The second optimates the downscaled image by the guidance of the original one, avoiding undesirable artifacts.

SCN (Sparse Coding based Network) adopts a neural network based on sparse coding, trained in a cascaded structure from end to end. It introduces some improvements in terms of both recovery accuracy and human perception employing a CNN (Convolutional Neural Network) model.

In LCI, the input RGB image is globally approximated by the bivariate Lagrange interpolating polynomial at a suitable grid of first kind Chebyshev zeros. The output RGB image is obtained by sampling this polynomial at the Chebyshev grid of the desired size. Since the LCI method works both in upscaling and downscaling, according to the notation in \cite{Lagrange}, we use the notation u-LCI and d-LCI in upscaling and in downscaling, respectively.

BIC, one of the  most commonly used rescaling methods, employs bicubic interpolation. It computes the unknown pixel value as a weighted average of $4\times 4$  pixels closest to it. Note that BIC produces noticeably sharper images than the other classical non-adaptive methods such as  bilinear and nearest neighbor,  offering  w.r.t. them  a favorable quality image and processing time ratio.

We remark that in the following, BIC is implemented by the Matlab built-in function \texttt{imresize} with \texttt{bicubic} option. For the other methods, we used the publicly available Matlab codes provided by the authors with the default parameters settings.
\section{Mathematical preliminaries}
\label{math_prel}
Let $I$ denote any color image of $n_1\times n_2$ pixels, with $n_1,n_2\in\NN$.
As is well--known, in the RGB space $I$ is represented by means of a triad of $n_1\times n_2$ matrices that we indicate using the same letter of the image they compose, namely $I_\lambda$, with $\lambda=1:3$ (i.e., $\lambda=1,2,3$). The entries of these matrices  are integers from $0$ to $\max_f$ that denotes the maximum possible value of the image pixel, (e.g., $\max_f= 255$ if the pixels are represented using 8 bits per sample).
On the other hand, such discrete values can be embedded in a vector function of the spatial coordinates, say $\f(x,y)=[f_1(x,y), f_2(x,y), f_3(x,y)]$, which represents the image at a continuous scale and whose sampling yields its digital versions of any finite size.

Hence, once fixed the sampling model, that is the system of nodes
\begin{equation}\label{X}
X_{\mu\times \nu}=\{(x_i^\mu, y_j^\nu)\}_{i=1:\mu, j=1:\nu}, \quad \mu,\nu\in\NN,
\end{equation}
we suppose that the digital image $I=[I_1,I_2,I_3]$  has behind the  function $\f=[f_1,f_2,f_3]$ such that
\begin{equation}\label{I}
I(i,j)=\f(x_i^{n_1},y_j^{n_2}),\quad i=1:n_1,\quad j=1:n_2.
\end{equation}
In both downscaling and upscaling, the goal is getting an accurate reconstruction of  $I$ at  a different (reduced and  enhanced, resp.) size. Denoting by $N_1\times N_2$ the new size that we aim to get and denoting by $R=[R_1,R_2,R_3]$  the target resized image of $N_1\times N_2$ pixels, according to the previous settings, we have
\begin{equation}\label{R}
R(i,j)=\f(x_i^{N_1},y_j^{N_2}),\quad i=1:N_1,\quad j=1:N_2.
\end{equation}
From this viewpoint, the scaling problem becomes a typical approximation problem: how to approximate the values of $\f$ at the grid $X_{N_1\times N_2}$ once known the values of $\f$ at the finer (in downscaling) or coarser (in upscaling) grid $X_{n_1\times n_2}$.

Within this setting, the choice of the nodes system (\ref{X}) as well as the choice of the approximation tool are both decisive for the success of a scaling method. In the next subsections we  introduce these two basic ingredients and  the evaluation metrics we use for our scaling method.
\subsection{Sampling system}
Since it is well--known that any finite interval $[a,b]$ can be mapped onto $[-1,1]$, in the following, we suppose that  each spatial coordinate belongs to the reference interval $[-1,1]$, so that the sampling system in (\ref{X}) is included in the square $[-1,1]^2$.

In literature, the equidistant nodes model is usually adopted  for sampling. According to such traditional model, in (\ref{X}) the coordinates $\{x_i^{\mu}\}_i$ and $\{y_j^{\nu}\}_j$ are those nodes that divide the segment $[-1,1]$ into $(\mu+1)$ and $(\nu+1)$ equal parts, respectively.
On the other hand, we recall that other coherent choices of the sampling system (\ref{X}) have been recently investigated, for instance, in \cite{demarchi1}, \cite{demarchi2} for Magnetic Particle Imaging.

Here we follow the sampling model recently introduced in \cite{Lagrange}. According to this model, we assume that (\ref{X}) is the Chebyshev grid where the coordinates $\{x_i^{\mu}\}_i$  and $\{y_j^{\nu}\}_j$ are the zeros of the Chebyshev polynomial of first kind of degree $\mu$ and $\nu$, respectively. This means that in (\ref{X}) we are going to assume that
\begin{equation}\label{xy}
x_i^\mu= \cos (t_i^\mu) \qquad \mbox{and}\qquad
y_j^\nu= \cos (t_j^\nu)
\end{equation}
where, for all $n\in\NN$, it is
\begin{equation}\label{t}
t_k^n=\frac{(2k-1)\pi}{2n}, \qquad k=1:n.
\end{equation}

Hence, supposed that $\f$ is the vector function representing the image at a continuous scale,  at a discrete scale we interpret the digital version of the image with size $\mu\times\nu$, as resulting from the sampling of $\f$ at the Chebyshev grid $X_{\mu\times\nu}$ defined by (\ref{X}) and (\ref{xy})-(\ref{t}).

We point out that both the coordinates in (\ref{xy}) are not equidistant in $[-1,1],$ but they are arcsine distributed and become denser approaching  the extremes $\pm 1$. Such nodes distribution is optimal from the approximation point of view but rather unusual in image sampling. Nevertheless, from a certain perspective, our sampling model is related to the traditional sampling at equidistant nodes since the nodes in (\ref{t}) are equally spaced in $[0,\pi]$.
Indeed, the idea behind our sampling model is to transfer the sampling question from the segment to the unit semicircle, which is divided into equal arcs by the nodes system equation (\ref{t}).

The main advantage of adopting this unusual point of view is the possibility of globally approximating the image, in a stable and near--best way, by the interpolation polynomials introduced in the next subsection.

\subsection{Filtered VP interpolation}
Regarding the approximation tool underlying our method, we consider some filtered interpolation polynomials recently studied in \cite{Occo_Them_AMC}. Such kind of interpolation is based on a generalization of the trigonometric VP means (see \cite{Filbir_Them, Them_Van_Barel}) and, besides the number of nodes, it depends on two additional parameters which can be suitable modulated in order to reduce the Gibbs phenomenon (see \cite{Occo_Them_AMC, Themistoclakis_2011}).

More precisely, for any  $n_i,m_i\in\NN$ such that $m_i\le n_i$, $i=1,2$, let
\[
\n=(n_1,n_2),\quad\mbox{and}\quad \m=(m_1,m_2),
\]
and let $n,m$ denote indifferently the first components (i.e. $n_1,m_1$ resp.) or  the second components (i.e. $n_2,m_2$ resp.) of such vectors. Corresponding to these parameters,
 for any $r=0:(n-1)$, we define the following {\it orthogonal VP polynomials}
\begin{equation}\label{qr}
q_{m,r}^n(\xi)=\hspace{-.1cm}\left\{
\begin{array}{l}
\cos(r t) 
\qquad\qquad\mbox{if $0\le r\le (n-m)$},\\ [.1in]
\frac{n+m-r}{2m}\cos(r t)+\frac{n-m-r}{2m}\cos((2n-r)t)
\\ [.07in]
\qquad\qquad\qquad\mbox{if $n-m<r<n$},
\end{array}\right.
\end{equation}
where here and in the following $\xi\in [-1,1]$ and $t\in[0,\pi]$ are related by $\xi=\cos t$.

We recall the polynomial system in (\ref{qr}) consists of $n$ univariate algebraic polynomials of degree at most $(n+m-1)$ that are orthogonal with respect to the scalar product
\[
<F, G>=\int_{-1}^1F(\xi)G(\xi)\frac{d\xi}{\sqrt{1-\xi^2}}.
\]
They generate the space (of dimension $n$)
\[
S_m^n:=\sspan\{q_{m,r}^n:\ r=0:(n-1)\}
 \]
that is an intermediate polynomial space nested between the sets of all polynomials of degree at most $n-m$ and $n+m-1$.

The space $S_m^n$  has also an interpolating basis consisting of the so--called {\it fundamental VP polynomials} that, in terms of the orthogonal basis (\ref{qr}), have the following expansion \cite{Themistoclakis_2011},\cite{Occo_Them_AMC}
\begin{equation}\label{fundVP}
\Phi_{m,k}^{n}(\xi)=\frac 2n\left[\frac 12 +\hspace{-.2cm}\sum_{r=1}^{n-1} \cos(r t_k^n) q_{m,r}^n(\xi)\right],\quad k=1:n.
\end{equation}

\begin{figure*}[!htbp]
\begin{center}
\includegraphics[height=8cm, keepaspectratio]{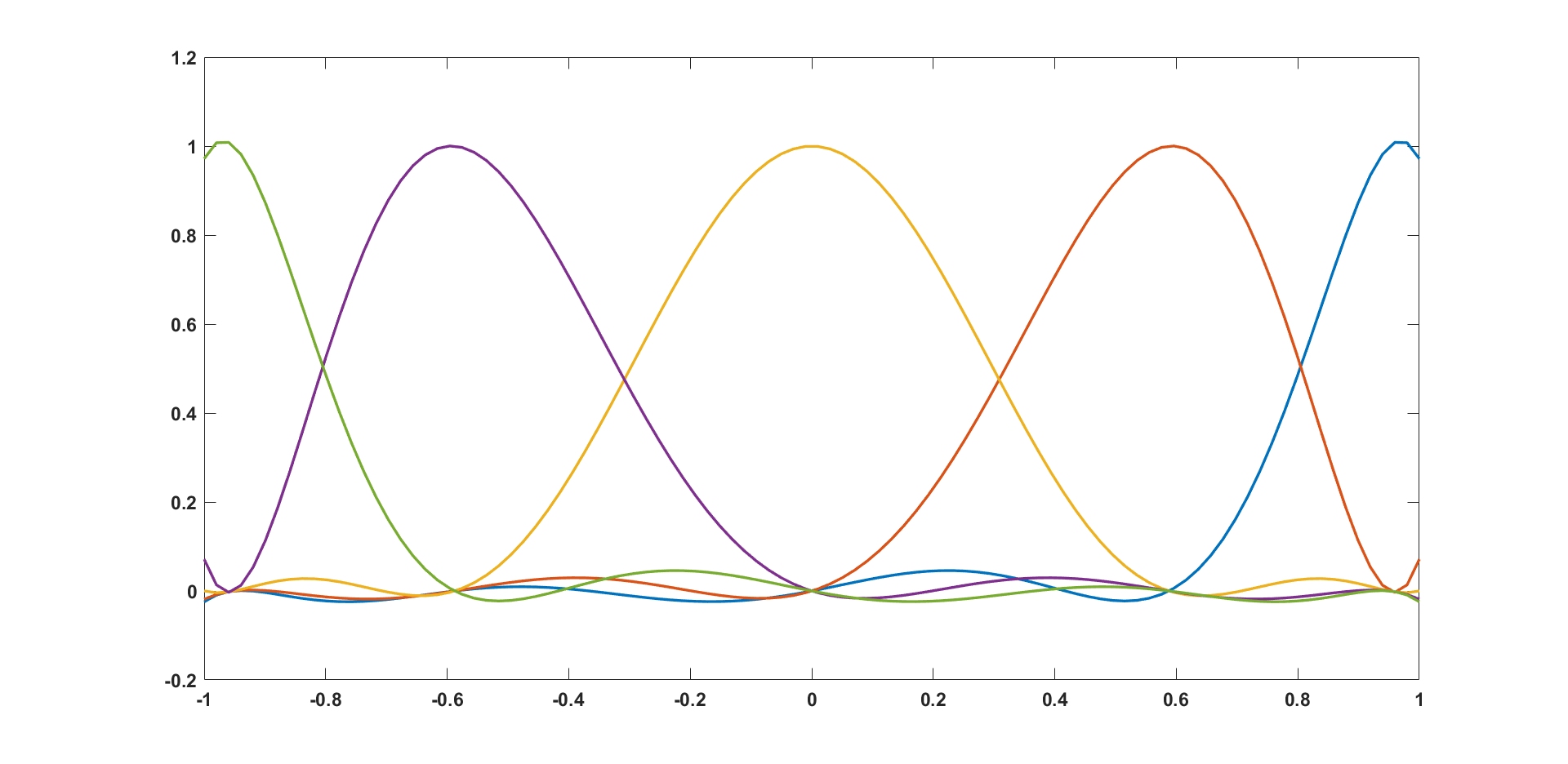}
\caption{Fundamental VP polynomials $\{\Phi_{m,k}^n\}_{k=1}^n$  for $n=5$ and $m=4$}
\end{center}
\end{figure*}

\begin{figure*} [!htbp]
\begin{center}
\includegraphics[height=8cm, keepaspectratio]{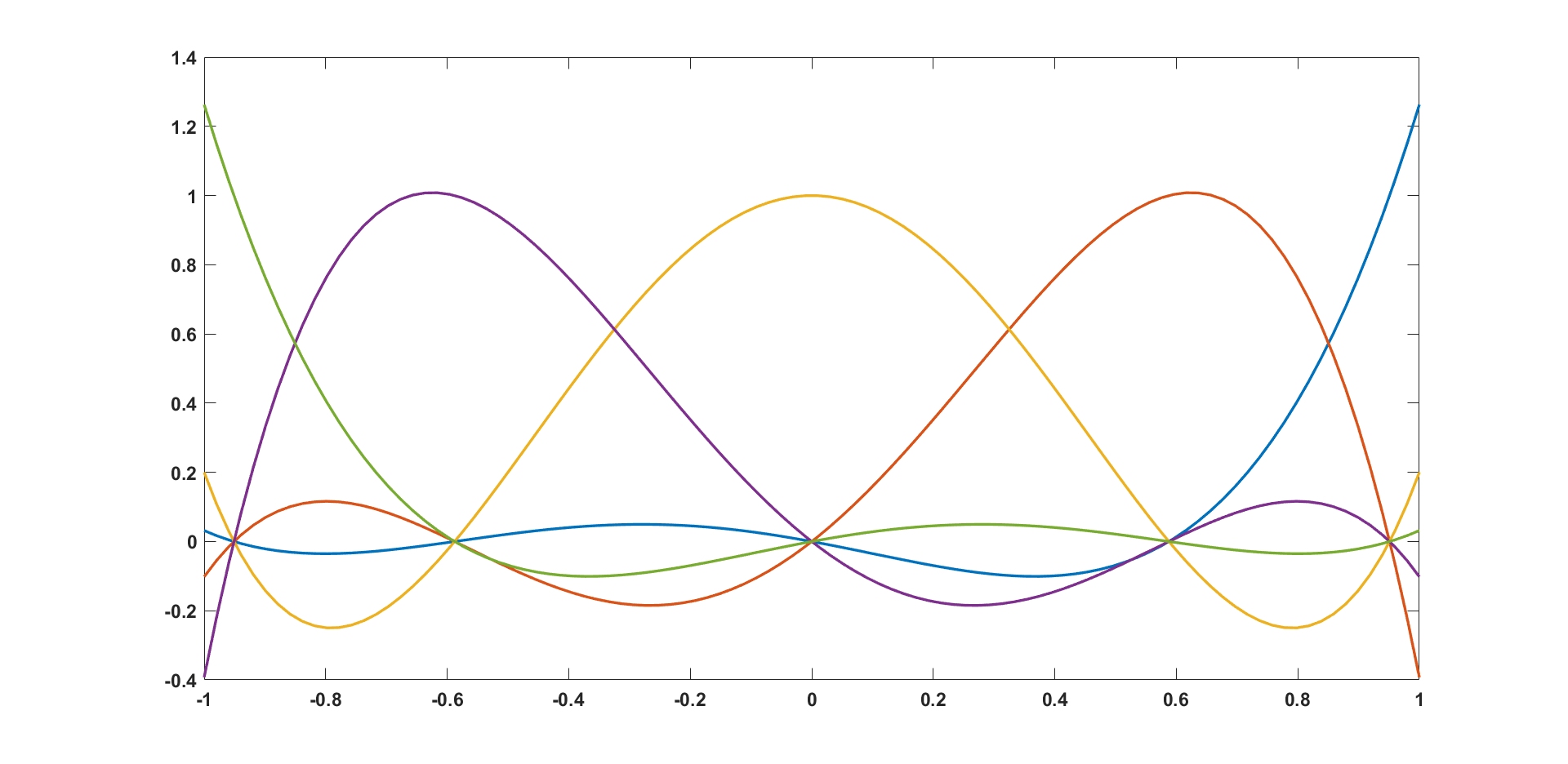}
\caption{Fundamental  Lagrange polynomials $\{\ell_{n,k}\}_{k=1}^n$  for $n=5$}
\end{center}
\end{figure*}


In Figures 1 and 2 we show respectively the fundamental VP polynomials for $n=5$ and $m=4$, and for the same $n=5$, the well-known fundamental Lagrange polynomials, defined as \begin{equation}\label{fundLAG}
l_{n,k}(\xi)=\frac 2n\left[\frac 12 +\hspace{-.2cm}\sum_{r=1}^{n-1} \cos(r t_k^n) \cos(r t)\right],\quad k=1:n.
\end{equation}
We see that, similarly to $\{l_{n,k}(\xi)\}_{k=1}^n$
also the fundamental VP polynomials satisfy the interpolation property
\begin{equation}\label{inter}
\Phi_{m,k}^{n}(\cos t_h^n)=l_{n,k}(\cos t_h^n)=\left\{
\begin{array}{lr}
1 & h=k\\
0 & h\ne k
\end{array}\right.
\end{equation}
for all $h,k=1:n$.

In addition to the number $n$ of nodes, we also have the free parameter $m$ which can be arbitrarily chosen ($m=1:n$ being possible) without loosing the interpolation property (\ref{inter}), as stated in \cite{Themistoclakis_2011}. Moreover, we note that also the limiting choice $m=0$ is possible, being, in this case, $S_0^n$ equal to the space of  polynomials  of degree at most $(n-1)$ and
\begin{equation}\label{limite}
\Phi_{0,k}^{n}(\xi)=l_{n,k}(\xi), \qquad  \forall |\xi|\le 1, \qquad k=1:n .
\end{equation}

We recall in \cite{CaThwave} both $n,m$ are chosen depending on a resolution level $\ell\in\NN$ and the fundamental VP polynomials constitute the scaling functions generating the multiresolution spaces $V_\ell=S_m^n$.

Another choice of $m$, often suggested in literature, is the following (see e.g. \cite{Occo_Them_Apnum},\cite{Occo_Them_DRNA})
\begin{equation}\label{mteta}
m=\lfloor\theta n\rfloor, \qquad\mbox{with $\theta\in ]0,1[$},
\end{equation}
where, $\forall a\in\RR^+$,  $\lfloor a \rfloor$ denotes the largest integer not greater than $a$.

\begin{figure*} [!htbp]
\begin{center}
\includegraphics[height=8cm, keepaspectratio]{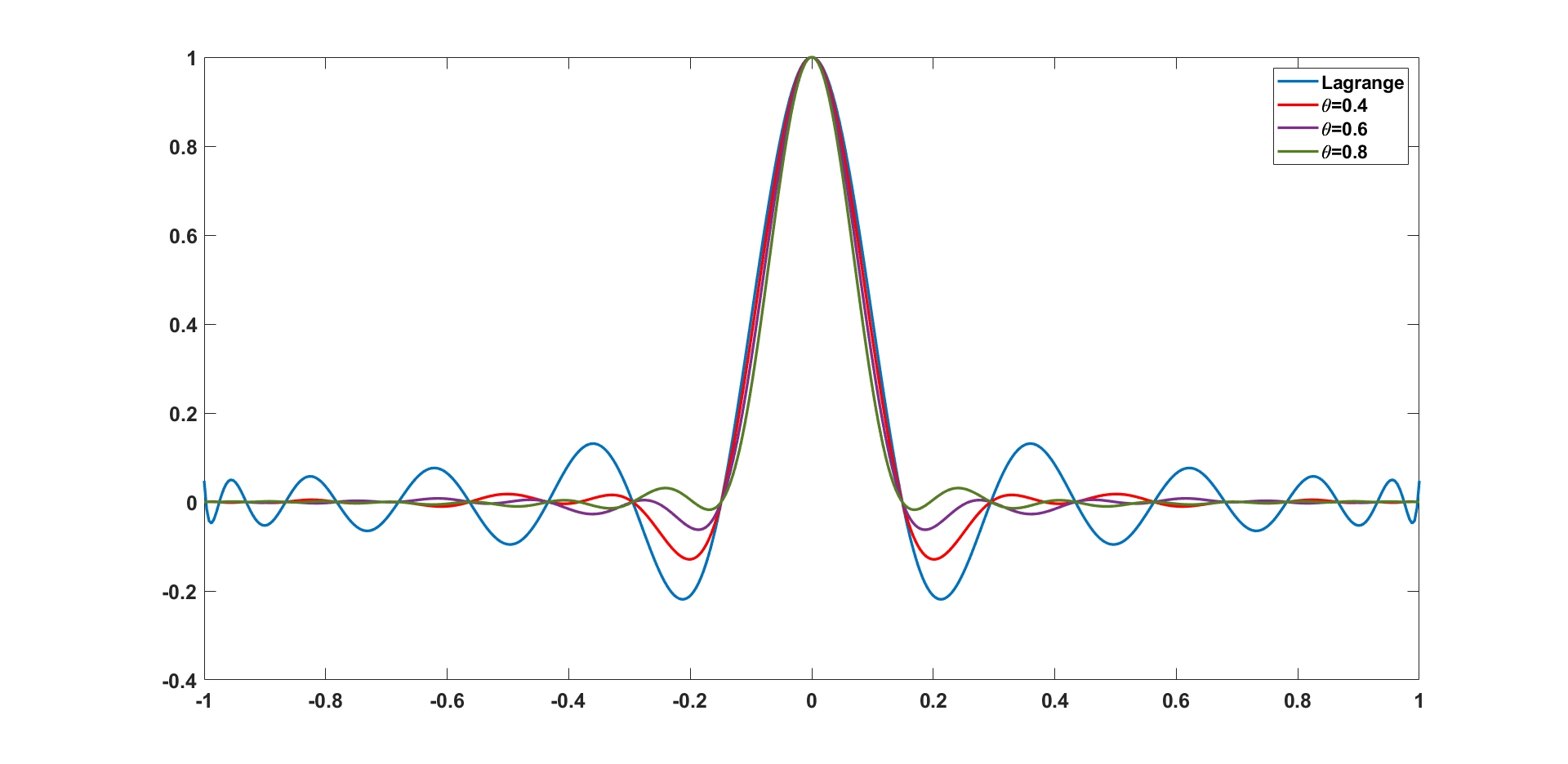}
 \caption{Fundamental  polynomials $\ell_{n,k}$ and $\Phi_{m,k}^n$ for $n=21,\ k=11$ and  $m=\lfloor n\theta\rfloor,$ with $\theta \in \{\ 0.4,\ 0.6,\ 0.8\}.$}
\label{fig:1}
\end{center}
\end{figure*}

Figure \ref{fig:1} displays the plots of the fundamental VP polynomials corresponding to fixed $n,k$ and $m$ given by (\ref{mteta}) with different values of $\theta$. Indeed such parameter (and more generally $m$) is responsible for the localization of the fundamental VP polynomial $\Phi_{n,k}^{m}(\xi)$ around the node $\xi_k^n=\cos t_k^n$. In fact, in Figure \ref{fig:1} we can see how those oscillations typical of the fundamental Lagrange polynomial $\ell_{n,k}$ (plotted too) are very dampened by suitable choices of $\theta$.

By using the fundamental VP polynomials (\ref{fundVP}) we can approximate any  function $g(x,y)$ on the square $[-1,1]^2$ by means of its samples at the Chebyshev grid (\ref{X}) as follows
 \begin{equation}\label{VP}
V_{\n}^{\m}g(x,y):=\sum_{i=1}^{n_1}\sum_{j=1}^{n_2} g(x_i^{n_1}, y_j^{n_2})\Phi_{n_1,i}^{m_1}(x)\Phi_{n_2,j}^{m_2}(y).
\end{equation}
This is the definition of the  {\it VP polynomial of $g$} and the approximation tool we use in  our method.

By virtue of (\ref{inter}), such polynomial coincides with $g$ at the grid $X_{n_1\times n_2}$, i.e.
\begin{equation}\label{inter-g}
V_{\n}^{\m}g(x_i^{n_1},y_j^{n_2})=g(x_i^{n_1},y_j^{n_2}), \quad i=1:n_1,\ j=1:n_2,
\end{equation}
Moreover, it has been proved that for any $(x,y)\in [-1,1]^2$, if (\ref{m-tested}) holds with an arbitrarily fixed $\theta\in ]0,1[$ then for all continuous functions $g$, the following limit holds uniformly on $[-1,1]^2$
\[
\lim_{\n\rightarrow \infty}\left|V_{\n}^{\m}g(x,y)-g(x,y)\right|=0 
\]
with the same convergence rate of the error of best polynomial approximation of $g$ \cite[Th.3.1]{Occo_Them_AMC}.
\subsection{Quality metrics}
Similar to most of the existing methods in literature, the performance of our method is quantitatively evaluated and compared with other scaling methods in terms of the Peak-Signal-to-Noise-Ratio (PSNR) and the Structural Similarity Index (SSIM). For our method such metrics will give a measure of the error between the target resized image $R=[R_1,R_2,R_3]$ and the output resized image that, in the following, we denote by $\tilde R=[\tilde R_1, \tilde R_2,\tilde R_3]$ .

The definition of PSNR is based on the standard definition of the Mean Squared Error between two matrices
\begin{equation}\label{def_mse}
\MSE(A,B)= \displaystyle\frac{1}{\nu\mu}\|A-B\|_F^2,\qquad \forall A,B\in \RR^{\nu\times \mu} \end{equation}
being  $\|\cdot \|_F$  the Frobenius norm defined as
\[
\|A\|_F:=\left(\sum_{h=1}^{\nu}\sum_{k=1}^{\mu} a_{h,k}^2\right)^\frac 12, \ \forall A=(a_{h,k})\in \RR^{\nu\times \mu}.
\]
The extension of such definition to the case of color digital images of $\nu\times\mu$ pixels can be performed in different ways giving rise to different measures of the related PSNR (see e.g. \cite{Ramella1, Matlab}). More precisely, for the color images $R$ and $\tilde R$, defining their MSE as follows
\begin{equation}\label{MSE-mean}
\MSE(\tilde R, R)=\frac 13 \sum_{\lambda=1}^3 \MSE(\tilde R_\lambda,R_\lambda),
\end{equation}
the first, usually adopted definition of PSNR (used for instance in \cite{Lagrange}) is the following
\begin{equation}\label{PSNR}
\PSNR(\tilde R, R)=20 \log_{10}\left(\frac{\max_f}{\sqrt{\MSE(\tilde R,R)}}\right).
\end{equation}
Another common way to measure the PSNR (also used in \cite{SCN}) is given  by converting to the color space YCrCb both the color RGB images $R=[R_1,R_2,R_3]$ and $\tilde R=[\tilde R_1,\tilde R_2,\tilde R_3]$, and separating the intensity Y (luma) channels that we denote by $R_Y$ and $\tilde R_Y$, respectively. We recall they are defined by the following weighted average of the respective RGB components
\begin{equation}\label{RY}
R_Y=\sum_{\lambda=1}^3 \alpha_\lambda R_\lambda+\alpha_4,\quad \tilde R_Y=\sum_{\lambda=1}^3 \alpha_\lambda \tilde R_\lambda+\alpha_4,
\end{equation}
 with $\alpha_i,\ i=1:4$ coefficients of the  ITU -R BT.601 standard (see e.g. \cite{Burger}).
Hence, taking the MSE of the matrices $R_Y$ and $\tilde R_Y$, the second, commonly used, definition of PSNR is referred to only such luma channel as follows
\begin{equation}\label{psnr}
\PSNR(\tilde R,R)=20 \displaystyle \log_{10}\left( \frac{\max_f}{\sqrt{\MSE(\tilde R_Y, R_Y)}}\right),
\end{equation}
We point out that in our experiments the PSNR has been computed using both the previous definitions. However, for brevity, in this paper we report only the values achieved by definition (\ref{psnr}), giving no new insight the results obtained by using the other definition (\ref{PSNR}).

Finally, also  the  SSIM metric is defined via the luma channel as follows \cite{SSIM}
\begin{equation}\label{def_ssim}
\SSIM(\tilde R, R)=  
\frac{\left[2\tilde\mu  \mu+c_1\right]\left[2\cov +c_2\right]}
{\left[{\tilde \mu}^2+\mu^2+c_1\right]\left[{\tilde\sigma}^2+\sigma^2+c_2\right]},
\end{equation}
where  $\tilde \mu, \mu$ and $\tilde\sigma, \sigma$ denote the average and variance of the matrices $\tilde R_Y, R_Y$, respectively, $\cov $  indicates their covariance, and the constants are usually fixed as $c_1=(0.01\times L), c_2=(0.03\times L)$ with the dynamic range of the pixel values $L=255$ in the case of 8-bit images.

\section{VPI scaling method}
\label{method}
According to the  notation introduced in the previous section, both $I$ and $R$ are digital versions (with $n_1\times n_2$ and $N_1\times N_2$ pixels, respectively) of the same continuous image represented by the vector function $\f=(f_1,f_2,f_3)$ (cf. (\ref{I}), (\ref{R})).
Nevertheless, to be more general, in view of the finite representation of the data and  the accuracy used to store the image, we suppose the effective input image of our method is a more or less corrupted version of $I$. We denote it  by $\tilde I=[\tilde I_1,\tilde I_1,\tilde I_3]$ and assume that  there exists a corrupted function $\widetilde \f=(\widetilde f_1,\widetilde f_2,\widetilde f_3)$ such that
\begin{equation}\label{I-tilde}
\tilde I(i,j)=\tilde\f(x_i^{n_1},y_j^{n_2}),\quad i=1:n_1,\quad j=1:n_2.
\end{equation}
Starting from these initial data, VPI method computes the output image $\tilde R$ having the desired size $N_1\times N_2$ and defined as follows
\begin{equation}\label{R-tilde}
\tilde R(i,j)=V_{\n}^{\m}\tilde \f(x_i^{N_1},y_j^{N_2}), \ i=1:N_1,\ j=1:N_2.
\end{equation}
In terms of the RGB components  $\tilde R=[\tilde R_1,\tilde R_2, \tilde R_3]$, by (\ref{VP}), this means that for any $i=1:N_1,\ j=1:N_2$ and $\lambda=1:3$ we have
\begin{eqnarray}\label{Rk-tilde}
\tilde R_\lambda(i,j)&=&V_{\n}^{\m}\tilde f_\lambda(x_i^{N_1},y_j^{N_2})=\\
\nonumber
&=& \sum_{u=1}^{n_1}\sum_{v=1}^{n_2} \tilde I_\lambda(u,v)\Phi_{n_1,u}^{m_1}(x_i^{N_1})\Phi_{n_2,v}^{m_2}(y_j^{N_2}),
\end{eqnarray}
that is
\begin{equation}\label{R-prod}
\tilde R_\lambda=V_1^T \tilde I_\lambda V_2, \quad \lambda=1:3,
\end{equation}
where the matrices $V_1\in \RR^{n_1\times N_1}$ and $V_2\in \RR^{n_2\times N_2}$ have the following entries
\begin{eqnarray}\label{V1}
V_1(i,j)&=&\Phi_{n_1,i}^{m_1}(x_j^{N_1}),\qquad i=1:n_1,\ j=1:N_1\\
\label{V2}
V_2(i,j)&=&\Phi_{n_2,i}^{m_2}(y_j^{N_2}),\qquad i=1:n_2,\ j=1:N_2
\end{eqnarray}

To compute $V_1,V_2$ efficient algorithms based on Fast Fourier transform can be implemented (see, e.g., \cite {FFT}). Moreover, by pre-computing matrices $V_i$, the representation (\ref{R-prod}) allows to reduce the computational effort when we have to resize many images for the same fixed sizes.
 
Now, we note that in the previous formulas, the integers $n_\ell$ and $N_\ell$  for $\ell=1,2$  are determined by the initial and final size of the scaling problem at hand, while the parameter $m_\ell$ is free.  Theoretically, it can be arbitrarily chosen from the set of integers between 1 and

$n_\ell $. According to (\ref{mteta}), for our method we fix
\begin{equation}\label{m-tested}
m_\ell=\lfloor\theta n_\ell\rfloor, \qquad \ell=1,2,\qquad \mbox{with}\qquad \theta\in ]0,1]
\end{equation}
including the limit case $\theta =1$ too.
Moreover, we also allow $\theta=0$, but, in this case, we remark that, by virtue of (\ref{limite}), VPI reduces to Lagrange--Chebyshev Interpolation (LCI) scaling method recently proposed in \cite{Lagrange}. In this sense, VPI can be considered a generalization of the LCI method.

Regarding the choice of the parameter $\theta$, in the  experimental validation of VPI, we consider two modes that we indicate in the sequel:  "supervised VPI" and "unsupervised VPI".  In the latter case, $\theta$ must be supplied by the user as an input parameter, arbitrarily chosen in $[0,1]$, where the choice $\theta=0$ means to select the LCI method.

Nevertheless,  if a target resized  image is available, we have structured VPI method in a supervised mode that requires  the target image as input argument, instead of the parameter $\theta$. In this case, we take several choices of $\theta\in [0,1]$ and, consequently, we get several matrices $V_1, V_2$ that determine, using (\ref{R-prod}), several resized images. Among these images, the one that,  once compared with the target image, gives the smallest MSE is chosen as the output image of the supervised VPI method.

In the sequel of this Section, we focus on  d-VPI method with odd scale factors $s=n_1/N_1=n_2/N_2$.  In the following proposition, we suppose that the lower resolution sampling satisfies the Nyquist limit so that the continuous image $\f$ can be uniquely reconstructed from both digital images $I$ and $R$ without any error. In this ideal case, we prove d-VPI method produces a MSE that it is not greater than $s^2$ times the MSE of the input data (cf. (\ref{eq-mse})) and, in particular, we get a null MSE if the input image is {\it exact} or, at least, only some {\it crucial} pixels of it are {\it exact} (cf. Remark 1). 

\begin{proposition}\label{prop}
Let $I$ and $R$  be the initial and resized true images given by (\ref{I}) and (\ref{R}) respectively, and let $\tilde I$ and $\tilde R$ by input and output images of d-VPI method, respectively given by (\ref{I-tilde}) and (\ref{R-tilde}), with arbitrarily fixed integer parameters $m_1<n_1$ and $m_2<n_2$. If there exists $\ell\in\NN$ that relates the initial size $n_1\times n_2$ with the final size $N_1\times N_2$ as follows
\begin{equation}\label{hp}
\frac{n_1}{N_1}=\frac{n_2}{N_2}=(2\ell-1),
\end{equation}
  then we have
 \begin{equation}\label{eq-mse}
\MSE(R,\tilde R)\le s^2 \MSE(I, \tilde I),\qquad s=(2\ell -1).
 \end{equation}
The same estimate holds also for the luma channel  and,  if in addition $I=\tilde I$ holds too, then we get
  \begin{equation}\label{dim_ssim}
  \PSNR(R,\tilde R)=\infty,\qquad\mbox{and}\qquad \SSIM (R,\tilde R)=1.\end{equation}
\end{proposition}
{\it Proof. } Recalling the definition (\ref{MSE-mean}), to prove (\ref {eq-mse}) it is sufficient to state the same inequality holds for the respective RGB components. Hence, according to our notation, let us state that for all $\lambda=1:3$ we have
 \begin{equation}\label{dim-mse}
\MSE(R_\lambda,\tilde R_\lambda)< s^2 \MSE(I_\lambda, \tilde I_\lambda).
 \end{equation}
To this aim, by using the short notation $n$ and $N$ to denote $n_i$ and $N_i$, respectively, for any  $i=1,2$, we note that whenever we have $n= s N$ with $s=(2\ell -1)$ and $\ell\in\NN$, all the zeros of the first kind Chebyshev polynomial of degree $N$ (i.e. $\cos(t_h^{N})$ with $h=1:N$) are also zeros of the first kind Chebyshev polynomial of degree $n$. More precisely, we have
\begin{equation}\label{nodi_3}
\cos(t_h^{N})=\cos(t_{i(h)}^{n}) \quad \mbox{with\ }  i(h)=\frac{s(2h-1)+1}2,
\end{equation}
where we point out that for all $h=1:N$ the index $i(h)=\frac{s(2h-1)+1}2$ is an integer between 1 and $n$ thanks to the hypothesis $s$ is odd.

By virtue of (\ref{nodi_3}), for all $h_1=1:N_1$ and $h_2=1:N_2$, recalling (\ref{R})--(\ref{t}) we get
\begin{equation}\label{RI}
\begin{array}{rl}
R_\lambda(h_1,h_2)=&f_\lambda\left(x_{h_1}^{N_1},y_{h_2}^{N_2}\right)\\
=& f_\lambda\left(x_{i(h_1)}^{n_1},y_{i(h_2)}^{n_2}\right)\\
=& I_\lambda\left(i(h_1),\ i(h_2)\right), \qquad \lambda=1:3.
\end{array}
\end{equation}
Similarly,  from (\ref{R-tilde}), (\ref{nodi_3}), (\ref{inter-g}) and (\ref{I-tilde}), we deduce
\begin{equation}\label{RI-tilde}
\begin{array}{rl}
\tilde R_\lambda(h_1,h_2)=&V_{\n}^{\m}\tilde f_\lambda\left(x_{h_1}^{N_1},y_{h_2}^{N_2}\right)\\ [.1in]
=& V_{\n}^{\m}\tilde f_\lambda\left(x_{i(h_1)}^{n_1},y_{i(h_2)}^{n_2}\right)
=\tilde f_\lambda\left(x_{i(h_1)}^{n_1},y_{i(h_2)}^{n_2}\right)\\ [.1in]
=& \tilde I_\lambda\left(i(h_1),\ i(h_2)\right), \qquad \lambda=1:3.
\end{array}
\end{equation}
Therefore, by (\ref{RI}) and (\ref{RI-tilde}), for any $\lambda=1:3$ we deduce (\ref{dim-mse}) as follows
\begin{eqnarray*}
&&\MSE(R_\lambda,\tilde R_\lambda)=\\
&=& \frac 1{N_1N_2}\sum_{h_1=1}^{N_1}\sum_{h_2=1}^{N_2}
\left[R_\lambda(h_1,h_2)-\tilde R_\lambda(h_1,h_2)\right]^2\\
&=&\frac 1{N_1N_2}\sum_{h_1=1}^{N_1}\sum_{h_2=1}^{N_2}
\left[I_\lambda\left(i(h_1),\ i(h_2)\right)-\tilde I_\lambda\left(i(h_1),\ i(h_2)\right)\right]^2\\
&\le & \frac 1{N_1N_2}\sum_{i=1}^{n_1}\sum_{j=1}^{n_2}
\left[I_\lambda(i,j)-\tilde I_\lambda(i,j))\right]^2\\
&=&\frac {s^2}{n_1n_2}\sum_{i=1}^{n_1}\sum_{j=1}^{n_2}
\left[I_\lambda(i,j)-\tilde I_\lambda(i,j))\right]^2\\
&=&s^2 \MSE(I_\lambda,\tilde I_\lambda).
\end{eqnarray*}
As regards the luma chanel, we note that by (\ref{RI})--(\ref{RI-tilde}) and (\ref{RY}) we easily deduce that
\begin{equation}\label{RY-1}
\begin{array}{rl}
R_Y(h_1,h_2)=&I_Y(i(h_1), \ i(h_2))\\
\tilde R_Y(h_1,h_2)=&\tilde I_Y(i(h_1), \ i(h_2))
\end{array}
\end{equation}
and such identities, similarly to the case of the RGB components, easily imply
\begin{equation}\label{RY-2}
\MSE(R_Y, \tilde R_Y)\le s^2 \MSE(I_Y,\tilde I_Y)
\end{equation}
Finally, in the case that $I=\tilde I$, from (\ref{RI}) and (\ref{RI-tilde}) we deduce that
\[
R_\lambda(h_1,h_2)=\tilde R_\lambda(h_1,h_2), \quad \lambda=1:3
\]
holds for any $h_1=1:N_1$  and $h_2=1:N_2$. Consequently we get the best result (\ref{dim_ssim}).
\Proofend

\begin{remark}
The previous proof shows that the hypothesis $I=\tilde I$ can be relaxed requiring that these images coincide only on some suitable pixels.

More precisely, in order to get (\ref{dim_ssim}) it is sufficient that
\begin{equation}\label{hp-reduced}
\begin{array}{l}
I_\lambda(i(h_1),\ i(h_2))=\tilde I_\lambda(i(h_1),\ i(h_2)), \\ [.1in]
\qquad \lambda=1:3, \quad h_1=1:N_1,\quad h_2=1:N_2
 \end{array}
\end{equation}
holds, where we defined
\begin{equation}\label{ih}
i(h)=\frac{s(2h-1)+1}2.
\end{equation}
\end{remark}

\begin{remark}
From (\ref{RI-tilde}), we deduce that in all cases of downscaling with odd scale factors, the choice of the parameter $\theta$, and more generally, the values we assign to $ \ m $ do not matter as d-VPI always returns the same output image which coincides with the one produced by the d-LCI method. Moreover, by virtue of (\ref{RI-tilde}), starting from a given image and using the same odd scale factor in opposite directions, for any $\m$ ones may sequentially run u-VPI first and then d-VPI, getting back the initial image without any error.
In all downscaling with odd scale factors, formula (\ref{RI-tilde}) is used instead of \eqref{R-prod} to get the output image by d-VPI.
\end{remark}

In conclusion, we point out that the theoretical results of Proposition \ref{prop} do not exclude the possible occurrence of aliasing effects when we are downscaling input images  with high-frequency details. In this case, even starting from not corrupted HR images, d-VPI produces LR images with aliasing effects that, under a certain size, become more and more visible. The experimental results in the next section show that aliasing also occurs when the downscaling factor (if any) does not satisfy the hypothesis of Proposition \ref{prop}.

Following the sampling theorem \cite {Shannon}, the standard approach for minimizing aliasing artifacts involves limiting the spectral bandwidth of the HR input image by filtering the image via convolution with a kernel before subsampling. As a well-known side effect, the resulting LR output image might suffer from loss of fine details and blurring of sharp edges. Thus, many filters have been developed \cite{Wolberg} to balance mathematical optimality with the perceptual quality of the downsampled result. However, these filter-based methods can introduce undesirable ringing or over-smoothing artifacts. 

Both L$_0$ and DPID have focused on the aspects of detail preserving. Specifically, to solve the aliasing problem  L$_0$ proposes an L$_0$-regularized optimization framework where the gradient ratio and reconstruction prior are iteratively optimized in an alternative way. In contrast, DPID uses an inverse bilateral filter to emphasize the differences between areas with small detail and bigger ones with similar colors. However, the aliasing reduction process of these methods influences their performance results both in quality and CPU time terms, as it is possible to see in the next Section \ref{ER}.

In this paper, our attention is focused on studying the effect of VP interpolation applied to image resizing in both upscaling and downscaling. Consequently, we limit to suggest the employment of suitable convolutional filtering for high downscaling scale factors whenever the aliasing effects are too evident (see Subsection \ref{PE}). In the meantime, we are working on finding better solutions to reduce the possible aliasing effects in d-VPI.

\section{Experimental Results}
\label{ER}

In this section, we describe the experimental validation of VPI. We test it on some publicly available image datasets, and we compare it with the methods described in Section 2; namely, we compare d-VPI with BIC, d-LCI, L$_0$, DPID, and u-VPI with BIC, u-LCI, SCN.  Although DPID and L$_0$ (SCN) can also be applied in upscaling (downscaling) mode, we do not force the comparison with them in unplanned way to avoid an incorrect experimental evaluation.

All methods have run on the same computer with the configuration Intel Core i7 3770K CPU @350GHz in Matlab 2018a. In the following, Subsection \ref{Data}  introduces  the considered datasets while Subsections \ref{QE} and 5.3 are respectively devoted to  quantitative and qualitative performance evaluation, both in downscaling and upscaling.
 
\subsection{Datasets}
\label{Data}
To be more general, besides the datasets used by the benchmark methods  \cite{Weber,Liu,SCN,Lagrange,BIC}, we also consider some datasets  offering different characteristics and extensively employed in Image Processing. \\
Specifically, the d-VPI performance evaluation is carried out on some publicly available datasets comprising 1026 color images in total. In particular, we consider BSDS500 dataset \cite{Martin}, available at \cite{Berkeley} which includes 500 color images having the same size (481$\times$321 or 321$\times$481). This set, also used in \cite{Liu, Lagrange}, is sufficiently general and provides a large variety of images often employed  also in other different image analysis tasks, such as in image segmentation \cite{seg_survey, seg1, seg2, seg3} and color quantization \cite{Chaki, CQ1, CQ2, CQ3}.
We also consider the following datasets to favor the comparison with the benchmark methods.
\begin{itemize}
\item
 The 13 natural-color images of the user study in \cite{Oztireli} available at \cite{13US} and  here denoted by 13US. They are originally taken from the MSRA Salient Object Database \cite{Liu_2011},  used in a previous study \cite{Kopf} and also employed in \cite{Weber}.
These images  have sizes ranging from 241$\times$400 to 400$\times$310 pixels.  \item
The extensive two sets  selected in \cite{Weber} from the Yahoo 100Mimage dataset \cite{Thomee} and the NASA Image Gallery \cite{NASA}, available at \cite{NASA_set}. We denote by NY17 and NY96 the corresponding sets of color images extracted from them. These sets comprise 17 and 96 color images, with sizes ranging from 500$\times$334 to 6394$\times$3456, respectively.
\item The Urban100 dataset \cite{Urban100} including 100 color images related to an urban context, with one  dimension  at most equal to 1024 and the other ranging from 564 to 1024 pixels. It has also been employed in \cite{Liu}.
\item The dataset PEXELS300 considered in \cite{Lagrange} and available with VPI code. It consists of 300 color images randomly selected  from \cite{pex}  and  originally having different large sizes that  we centrally cropped to 1800$\times$1800 pixels.
\end{itemize}

Regarding  the u-VPI performance evaluation, in addition to the previous datasets, we have also used the following well--known datasets, commonly used by the Super Resolution community \cite{Hayat, Li-DL} for a total of 1943 color images.
\begin{itemize}
\item
The 5 images, known  in the literature as Set5 and originally taken from \cite{Bevilacqua},  with sizes ranging from 256$\times$256 to 512$\times$512 available at \cite{Set5}.
\item
The 12 color images belonging to the Set14  \cite{Zeyde}, with sizes ranging from 276$\times$276 to 512$\times$768 available at \cite{Set14}.
\item
The image dataset DIV2k (DIVerse 2k) consisting of high-quality resolution images used for the NTIRE 2017 SR challenge (CVPR 2017 and CVPR 2018) \cite{Timofte} available at \cite{DIV2k}. It comprises the train set (DIV2k-T) and the valid set (DIV2k-V), with 800 and 100 color images, respectively. Such images have one dimension equal to 2040, while  the other  ranges from 768 to 2040.  DIV2k has permitted testing the performance of all the benchmark methods on input images characterized by different degradations. Such input images are included in DIV2k and  collected as follows:
\begin{itemize}
\item
 DIV2k-T-B ( DIV2k-V-B), generated by BIC (-B);\item
 DIV2k-T-u ( DIV2k-V-u), classified as unknown (-u); \item
DIV2k-T-d ( DIV2k-V-d), classified as difficult (-d);\item
DIV2k-T-m ( DIV2k-V-m), classified as mild (-m).
\end{itemize}
 \end{itemize}

Since DIV2k is the only one to contain both the input image and the target  image, in order to implement supervised VPI with the other datasets, we fix the images of these datasets as target images with $N_1\times N_2$ pixels. Hence, we generate the respective input images, with size $n_1\times n_2$, by a scaling method which we assume to be BIC in most cases, since it can be used both in downscaling (for testing supervised u-VPI) and upscaling (for testing supervised d-VPI). However, in Subsection 5.2.3, we also analyze the implementation of the other scaling methods from Section 2 to generate the input images.

For simplicity, in the following, we distinguish how the input image is generated by premising the name of the generating method. For instance, the input image generated by BIC is indicated as BIC input image.

\vspace{.2cm}

\subsection{Quantitative evaluation}
\label{QE}
For the quantitative evaluation, we compute, both in upscaling and downscaling, the visual quality measures PSNR, SSIM  (cf. Subsection 3.3), and the CPU time for VPI and the benchmark methods starting from the same input image. Here we focus on the  quality measures while the CPU time is analyzed in Subsection 5.2.3.

 Since the target image is necessary to compute PSNR and SSIM, we employ the supervised VPI both in upscaling and downscaling. Specifically, we let the free parameter $\theta$ vary from 0.05 to 0.95 with step 0.05. In this way, we get 19 resized images, and we take as output image the one with minimum MSE.

First, we test supervised VPI and the benchmark methods on the DIV2k image dataset. As above specified, this is the only dataset that, for certain fixed scaling factors, includes both the input $n_1\times n_2$ images and the target $N_1\times N_2$ image. Since in DIV2k only the cases
\[
(N_1, N_2)=s (n_1, n_2), \qquad\mbox{with}\qquad s=2,3,4,
\]
are present, on DIV2k we test upscaling methods (supervised u-VPI, BIC, u-LCI, and SCN) for these scale factors.
Tables \ref{DIV2k-up234} and \ref{DIV2k-up4} show the average results of PSNR and SSIM computed with target images from both the train and valid sets (DIV2k-T and DIV2k-V, resp.), taking as input images the respective ones classified in DIV2k as BIC (-B), unknown (-u), difficult (-d), and mild (-m).
We remark that Table \ref{DIV2k-up4} concerns only the case $s=4$ since for  $s=2,3$ the input LR images are not present in DIV2k-T-d/m and DIV2k-V-d/m  datasets.

To test supervised VPI and the benchmark methods on the other datasets detailed in Subsection 5.1, as previously announced, we take as target $N_1\times N_2$ images the ones in the datasets and apply BIC to them in order to generate the input $n_1\times n_2$ images. For brevity, in both upscaling and downscaling,
 we show only the performance results for the scale factors s=2,3,4, which means the input size $n_1\times n_2$ is computed from the target size according to the following formula
 \begin{equation}\label{input-size}
 n_i=\left\{\begin{array}{ll}
 sN_i & \mbox{to test downscaling methods}\\ [.1in]
 \left\lfloor 
 \frac{N_i}{s}\right\rfloor & \mbox{to test upscaling methods}
 \end{array}\right.\ i=1,2.
 \end{equation}
Tables \ref{up234} and \ref{down234} concern upscaling and downscaling, respectively, and show the average results of PSNR, SSIM values obtained for the datasets and methods specified in the first columns. Note that for all methods we  provide the input images generated from those in the datasets, with size $N_1\times N_2$, by applying BIC according to (\ref{input-size}).

We remark that, in upscaling, the comparison with SCN is limited to DIV2k since, for the images in the datasets of Table \ref{up234}, the SCN demo version does not always produce the exact size of the HR image, making it impossible to compute the quality measure values.

\begin{figure*}[!htbp]
\includegraphics[height=4.5cm,width=8.5cm]{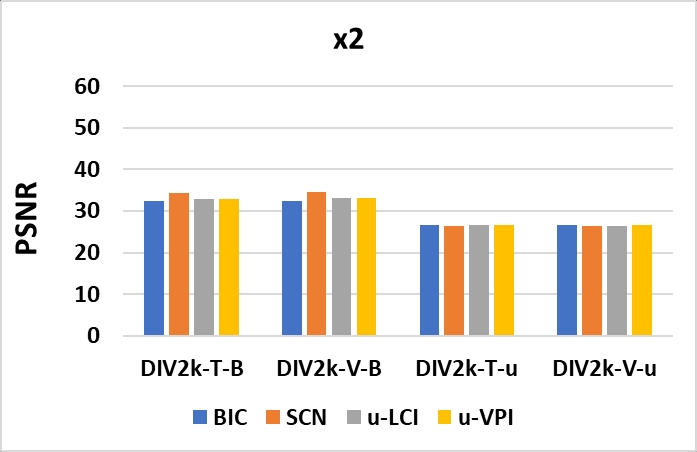}
\includegraphics[height=4.5cm,width=8.5cm]{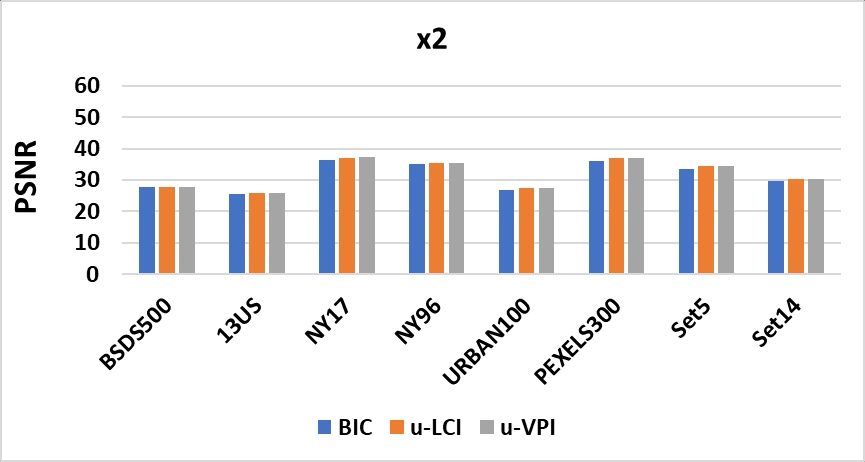}
\newline
\newline
\includegraphics[height=4.5cm,width=8.5cm]{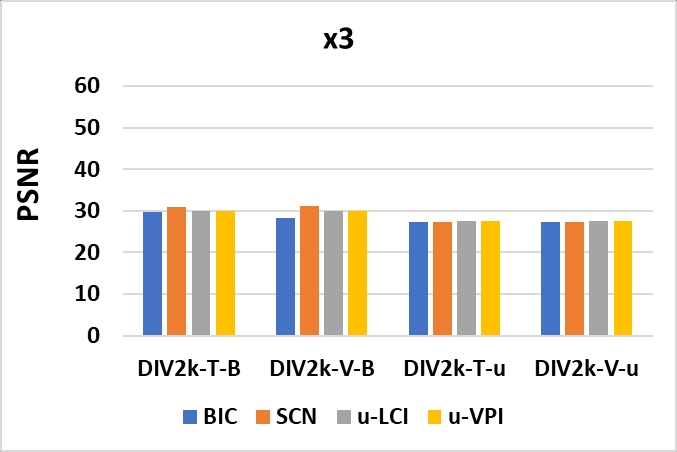}
\includegraphics[height=4.5cm,width=8.5cm]{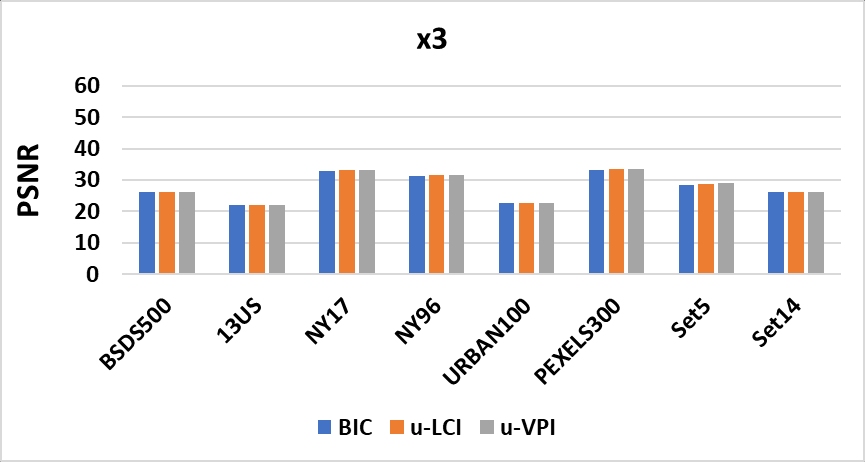}
\newline
\newline
\includegraphics[height=4.5cm,width=8.5cm]{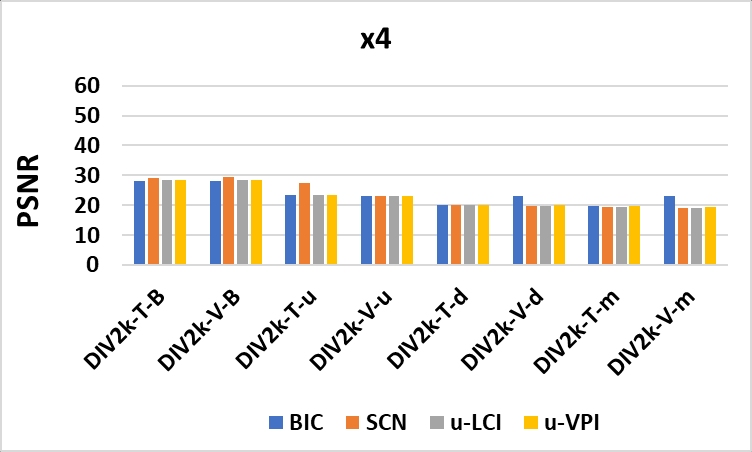}
\includegraphics[height=4.5cm,width=8.5cm]{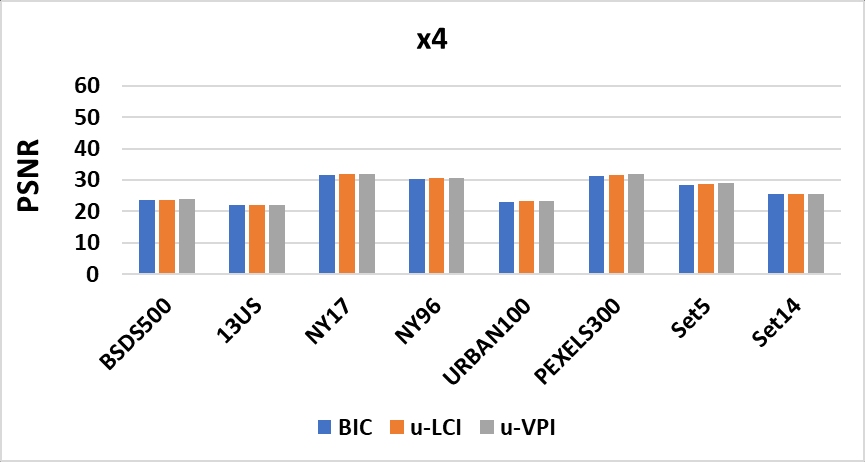}
\newline
\newline
\includegraphics[height=4.5cm,width=8.5cm]{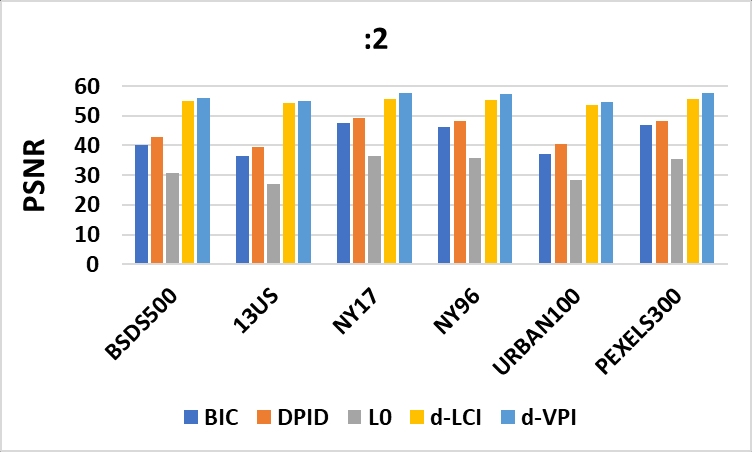}
\includegraphics[height=4.5cm,width=8.5cm]{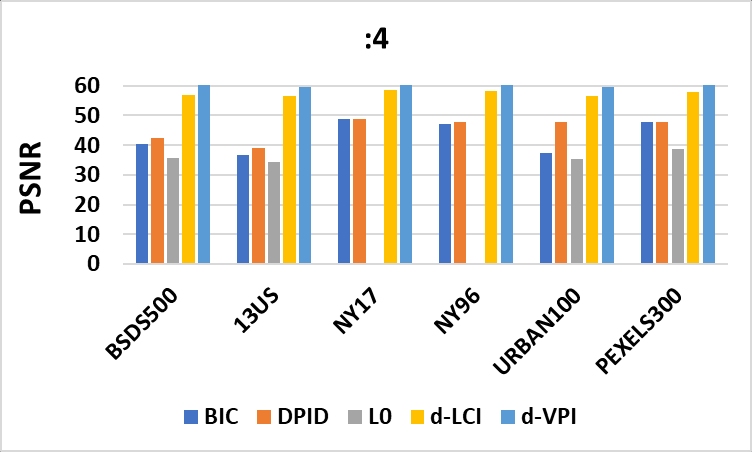}
%
\caption {PSNR values extracted from Tables \ref{DIV2k-up234}--\ref{down234}}
\label{fig:12a}
\end{figure*}

\begin{figure*}[!htbp]
\includegraphics[height=4.5cm,width=8.5cm]{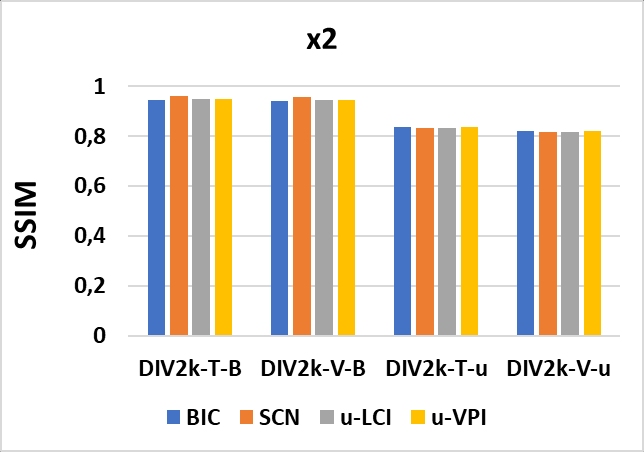}
\includegraphics[height=4.5cm,width=8.5cm]{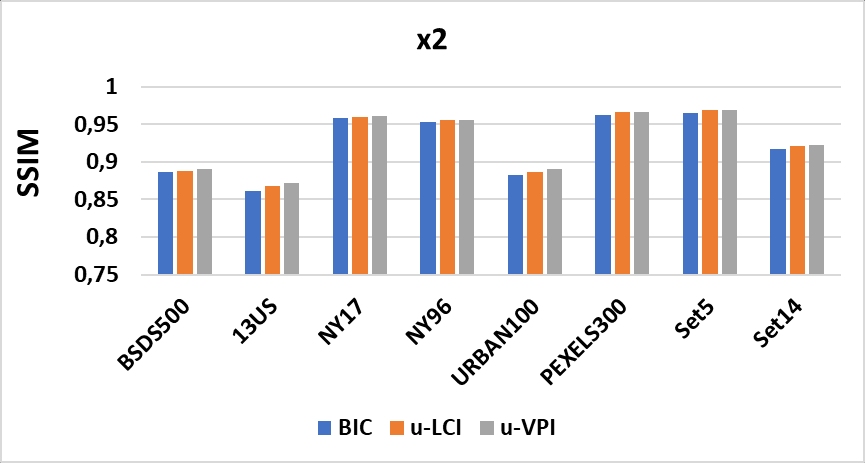}
\newline
\newline
\includegraphics[height=4.5cm,width=8.5cm]{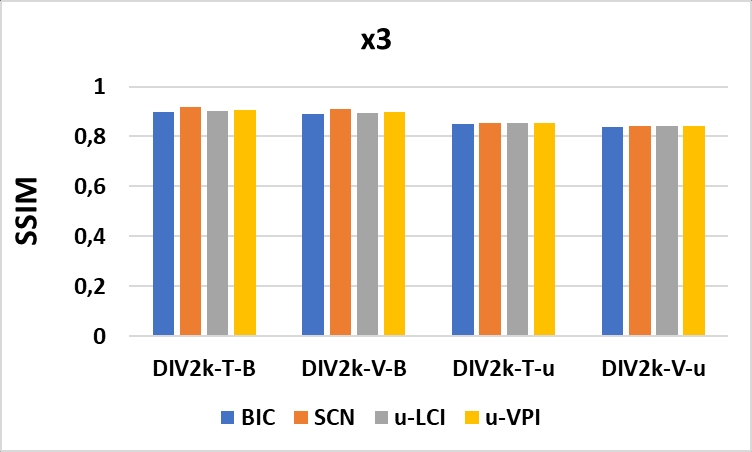}
\includegraphics[height=4.5cm,width=8.5cm]{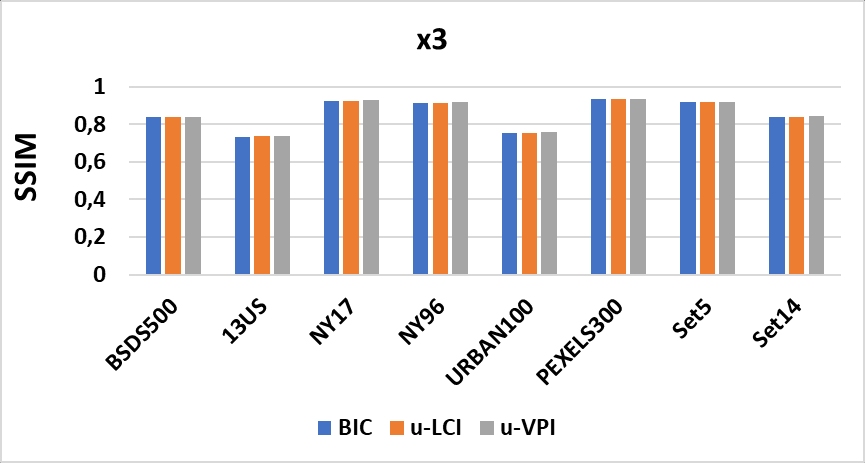}
\newline
\newline
\includegraphics[height=4.5cm,width=8.5cm]{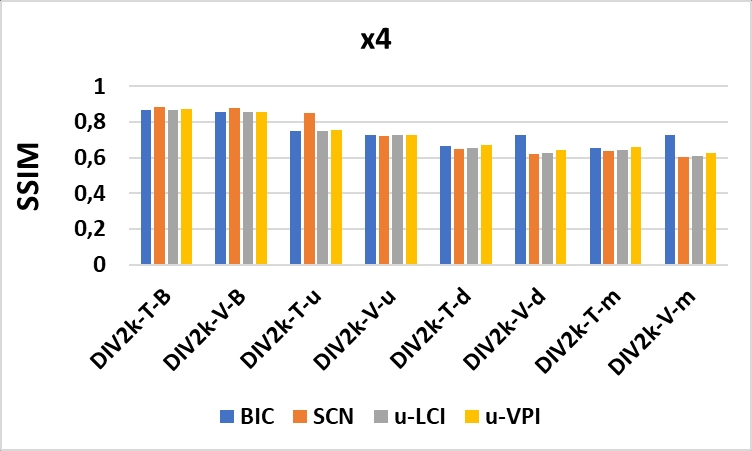}
\includegraphics[height=4.5cm,width=8.5cm]{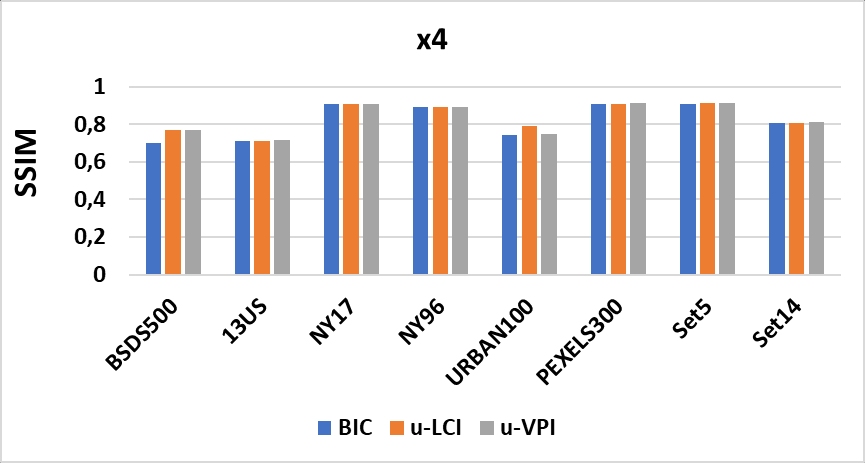}
\newline
\newline
\includegraphics[height=4.5cm,width=8.5cm]{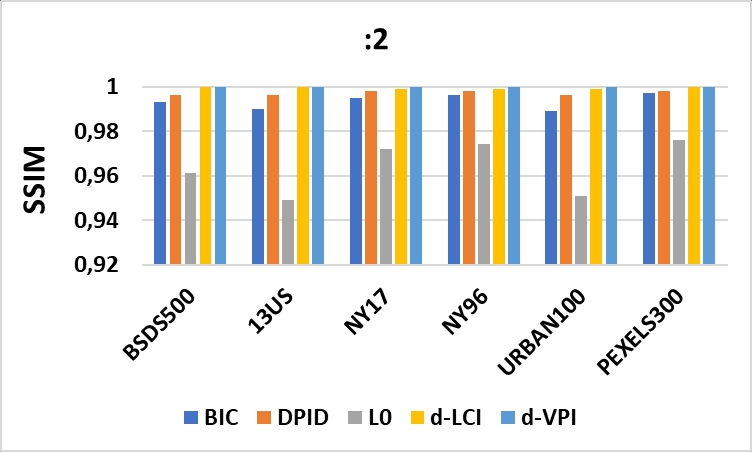}
\includegraphics[height=4.5cm,width=8.5cm]{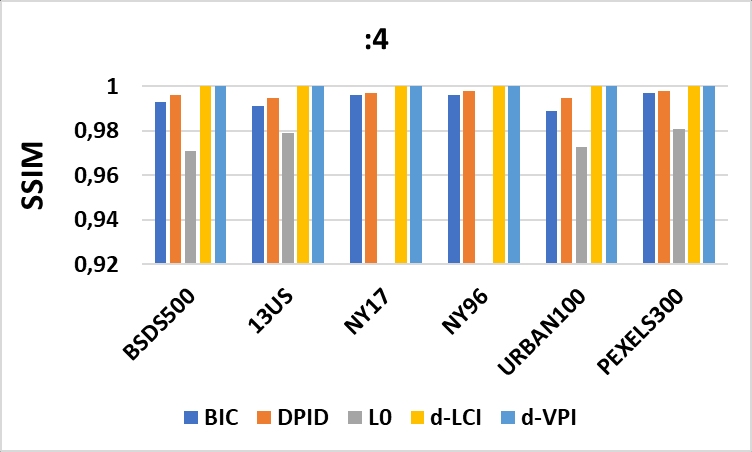}
%
\caption{SSIM values  extracted from Tables \ref{DIV2k-up234}--\ref{down234}}
\label{fig:11a}
\end{figure*}

 The bar graphs describing the trend discovered by Tables \ref{DIV2k-up234}--\ref{down234} are shown in Figures \ref{fig:12a} and \ref{fig:11a} for PSNR and SSIM values, respectively.

\begin{table}[!htbp]
\caption{Average performance of upscaling methods on DIV2k  with input images generated by BIC (-B) and classified as unknown (-u)}
\label{DIV2k-up234}
\tiny{
\begin{tabular}{r|rr|rr|rr}
\hline
  &\multicolumn{2}{c}{x2} & \multicolumn{2}{|c}{x3} & \multicolumn{2}{|c}{x4}\\ \hline
& PSNR & SSIM & PSNR  & SSIM  & PSNR  & SSIM \\
\hline
\textbf {DIV2k-T-B}& & & & & & \\
BIC   & 32,369          &	0,944  & 29,623  &	0,899  & 28,094 &	0,865 \\
SCN   & \textbf{34,336} &	\textbf{0,960}  &  \textbf{30,924} &	\textbf{0,918}    & \textbf{29,170} &	\textbf{0,884}\\
u-LCI & 32,969          &	0,948            & 29,967 &	0,903  & 28,381                                             &	0,868 \\
u-VPI & 33,003          &	0,949            & 30,013 &	0,905  & 28,419  &	0,870 \\
\hline
\textbf {DIV2k-V-B}& & & & & & \\
BIC & 32,411&	0,940   &	29,647 &	0,891  & 28,108 &	0,853 \\
SCN & \textbf{34,513} &	\textbf{0,958 }& \textbf{31,078	}&	\textbf{0,912}& \textbf{29,312}&\textbf{0,875}\\
u-LCI & 33,010	&	0,944& 29,989  &	0,895  & 28,396  &	0,856 \\
u-VPI & 33,072	&	0,946 &30,036  &	0,897 & 28,433 &	0,859\\
\hline
\textbf {DIV2k-T-u}& & & & & & \\
BIC & 26,566&	0,835 & 27,292  &	0,849  & 23,409 	& 0,751 \\
SCN & 26,444  &	0,831  & 27,378 &	0,853  & \textbf{27,292} &	\textbf{0,849}\\
u-LCI &26,525	 &	0,834  & 27,430&	0,853  & 23,357&	0,749\\
u-VPI &\textbf{26,586}	 &	\textbf{0,836} & \textbf{27,433}&	\textbf{0,854}&23,435&	0,752\\
\hline
\textbf {DIV2k-V-u}& & & & & &  \\
BIC &26,536 &	0,820 & 27,279  &	0,836 & 23,233 &0,726\\
SCN & 26,412  &	0,816  & 27,368  &	0,840  & 23,074 &	0,720 \\
u-LCI & 26,496 &	0,818	 &	27,418 &0,840  &23,180  &	0,724 \\
u-VPI &\textbf{26,557}	 &	\textbf{0,821}  & \textbf{27,421}&	\textbf{0,841}& \textbf{23,262}&	 \textbf{0,727}\\
\hline
\end{tabular}
}
\end{table}
\begin{table}[!htbp]
\caption{Average performance of upscaling methods on DIV2k  with input images classified as difficult (-d) and mild (-m)}
\label{DIV2k-up4}
\tiny{
\begin{tabular}{r|rr|r|rr}
\hline
\multicolumn{3}{c}{x4} & \multicolumn{3}{c}{x4}\\ \hline
& PSNR  & SSIM  & & PSNR  & SSIM \\
\hline
\textbf {DIV2k-T-d}& & &  \textbf {DIV2k-V-d} &  \\
BIC & 20,056 &0,665  &                BIC & \textbf{23,233} & \textbf{0,726 }\\
SCN &  19,956 &0,649          &               SCN  &  19,824  & 0,621  \\
u-LCI & 20,010  & 0,656        &              u-LCI  & 19,876 & 0,628 \\
u-VPI& \textbf{20,138}  & \textbf{0,672}    & u-VPI & 20,012  & 0,644\\
\hline
\hline
\textbf {DIV2k-T-m}& & &  \textbf {DIV2k-V-m} &  \\
BIC & 19,589& 0,652   &    BIC & \textbf{23,233}& \textbf{0,726} \\
SCN &  19,475 &0,636   &                           SCN &  19,047  & 0,601 \\
u-LCI & 19,530  & 0,643     &                       u-LCI & 19,095  & 0,608 \\
u-VPI & \textbf{19,735}  & \textbf{0,661}     &    u-VPI& 19,332  & 0,627 \\
\hline
\hline
\end{tabular}
}
\end{table}

\begin{table}[!htbp]
\caption{Average performance of upscaling methods with BIC input images}
\label{up234}
\tiny{
\begin{tabular}{r|rr|rr|rr}
\hline
  &\multicolumn{2}{c}{x2} & \multicolumn{2}{|c}{x3} & \multicolumn{2}{|c}{x4}\\ \hline
& PSNR & SSIM  & PSNR  & SSIM & PSNR  & SSIM \\
\hline
\textbf {BSDS500}& & & & & & \\
BIC & 27,665 &	0,886	 &26,148		&0,837	&23,678		&0,701	 \\
u-LCI & 27,707	&0,888	 &26,196	&0,839	 &23,793 &0,769	\\
u-VPI& \textbf{27,748}		&\textbf{0,890}	 &\textbf{26,237}	&\textbf{0,841}	 &\textbf{23,865} &\textbf{0,770}\\
\hline
\textbf {13US}& & & & & & \\
BIC & 25,429 &	0,861	 &22,125		&0,734&21,906		&0,710	  \\
u-LCI & 25,800&0,868	 &22,127	&0,738	 &22,010&0,713	\\
u-VPI& \textbf{25,859}		&\textbf{0,872}	 &\textbf{22,194}	&\textbf{0,739}	 &\textbf{22,045} &\textbf{0,716}\\
\hline
\textbf {NY17}& & & & & &  \\
BIC & 37,638&	0,958	 & 32,298&	0,924	&32,232	&	0,907  \\
u-LCI & 38,485 &0,960	&34,910&	0,924  &32,793&	0,908  \\
u-VPI & \textbf{38,540} &	\textbf{0,961}	& \textbf{34,976}&	 \textbf{0,927}1&\textbf{32,830}	 &	 \textbf{0,910}  \\
\hline
\textbf {NY96}& & & & & & \\
BIC & 34,979 &	0,953& 31,354 &	0,913	&30,368	&	0,891 \\
u-LCI & 35,507	&	0,955	&31,556 &0,914&30,607&	0,891 \\
u-VPI & \textbf{35,573}	&	\textbf{0,956}  & \textbf{31,602} &\textbf{0,916}&\textbf{30,654}&	 \textbf{0,894}  \\
\hline
\textbf {URBAN100}& & & & & & \\
BIC & 26,860 &	0,882& 22,737 &	0,755	&23,135	&	0,741 \\
u-LCI & 27,321	&	0,886	 &22,755 &0754&23,350&	0,793 \\
u-VPI & \textbf{27,387}	&	\textbf{0,891} & \textbf{22,802} &\textbf{0,759}&\textbf{23,388}&	 \textbf{0,748}\\
\hline
\textbf {PEXELS300}& & & & & &  \\
 BIC & 36,249 & 0,96 & 33,147 & 0,932&  31,374	&	0,908\\
 u-LCI & 37,067	& \textbf{0,966}&  33,622 &0,935 &31,741	 &0,910\\
 u-VPI &  \textbf{37,128}	&	 \textbf{0,966}&\textbf{33,671}& \textbf{0,936}&  \textbf{31,786} & \textbf{0,912}\\
\hline
\textbf {Set5}& & & & & &  \\
BIC & 33,646&	0,965 & 28,596	&0,916 & 28,425	&	0,908\\
u-LCI & 34,499	&	\textbf{0,969 }& 28,881 &\textbf{0,918}&28,915	 &0,912\\
u-VPI&\textbf{34,540}	&	\textbf{0,969} & \textbf{28,924} &\textbf{0,918} &\textbf{28,946} &\textbf{0,913}\\
\hline
\textbf {Set14}& & & & & &  \\
BIC &  30,375	 &	 0,917	& 27,597 &0,839 &26,022	&	0,807\\
u-LCI &31,020	&	0,921& 28,026 	&0,841&26,333	 &	0,809\\
u-VPI & \textbf{31,080}	&	\textbf{0,923}& \textbf{28,064} 	&\textbf{0,843}&\textbf{26,371}	 &	 \textbf{0,812}\\
\hline
\end{tabular}
}
\end{table}

\begin{table}[!htbp]
\caption{Average performance of downscaling methods with BIC input imagess (oom is the short way to indicate "out of memory").}
\label{down234}
\tiny{
\begin{tabular}{r|rr|rr|rr}
\hline
&\multicolumn{2}{c}{:2} & \multicolumn{2}{|c}{:3} & \multicolumn{2}{|c}{:4}\\ \hline
& PSNR  & SSIM  & PSNR  & SSIM  & PSNR & SSIM \\
\hline
\textbf {BSDS500} & & & & & &  \\
BIC    & 40,152  &	0,993	     &40,526	&0,993	&40,456	&0,993	  \\
DPID   & 43,011	 &	0,996           &43,090  &0,996	 &42,481		&0,996	 \\
L$_0$  & 30,742	 &  0,961	     &34,304	&0,971	&35,585	&0,971	\\
d-LCI  & 54,852	 &\textbf{1,000} &\textbf{$\infty$}	&\textbf{1,000}	 &56,890	 &\textbf{1,000}	 \\
d-VPI  & \textbf{56,025} &\textbf{1,000}	&\textbf{$\infty$}	&\textbf{1,000}	&\textbf{60,928}	 &\textbf{1,000}	\\
\hline
\textbf {13US}& & & & & &  \\
BIC & 36,419	 &	0,990	 & 36,755  &	0,991	&36,683	&	0,991  \\
DPID & 39,405	&	0,996  &39,676	 &	0,996	&38,988	&0,995	\\
L$_0$     & 27,008	&0,949 & 31,637 &0,972	&34,467	&0,979	\\
d-LCI &54,172 	&\textbf{1,000} & \textbf{$\infty$} &\textbf{1,000}	&56,541 &\textbf{1,000}	 \\
d-VPI & \textbf{55,039} 	&\textbf{1,000} & \textbf{$\infty$} &\textbf{1,000}	 &\textbf{59,529} &\textbf{1,000}	\\
\hline
  \textbf {NY17}& & & & & &  \\
BIC & 47,688	&0,995&	  48,849&	0,996& 48,706 &	0,996\\
DPID& 49,218&0,998& 49,212&	0,998&	 48,706&	0,997	\\
L$_0$ & 36,461 	& 0,972	 & 39,043&	0,979  & oom & oom \\
d-LCI & 55,497	&	0,999 &\textbf{$\infty$}	&	\textbf{1,000}&58,497&	 \textbf{1,000}\\
d-VPI & \textbf{57,632}	&	\textbf{1,000} &\textbf{$\infty$}	&	\textbf{1,000}&\textbf{ 64,042}&	 \textbf{1,000}\\
\hline
\textbf {NY96}& & & & & & \\
BIC & 46,261&	0,996	 & 47,371	&0,997 & 47,189	&	0,996\\
DPID& 48,187&	0,998 & 48,398 &0,998& 47,901&	0,998\\
L$_0$ & 35,686 &	0,974  & oom & oom &  oom & oom \\
d-LCI & 55,227	& 0,999& \textbf{$\infty$} &\textbf{1,000}&58,306	 &	 \textbf{1,000}\\
d-VPI & \textbf{57,315}	&	\textbf{1,000} &\textbf{$\infty$}	&	\textbf{1,000}&\textbf{ 63,800}&	 \textbf{1,000}\\
\hline
\textbf {URBAN100}& & & & & &  \\
BIC & 37,071&	0,989	 &37,414&0,990 & 37,344	&	0,989\\
DPID& 40,648&	0,996& 40.858 &0,996& 40,192&	0,995\\
L$_0$ & 28,215 &0,951	 & 32,608 & 0,969 & 35,185 &  0,973\\
d-LCI & 53,800	& 0,999& \textbf{$\infty$} &\textbf{1,000}&56,560	 &	 \textbf{1,000}\\
d-VPI & \textbf{54,703}	&	\textbf{1,000}&\textbf{$\infty$}	&	\textbf{1,000}&\textbf{ 59,730}&	 \textbf{1,000}\\
\hline
\textbf {PEXELS300}& & & & & &  \\
 BIC  &  46,803 &	0,997	 &  47,834  & 0,997& 47,675&0,997	\\
 DPID & 48,193 &  0,998  & 48,388 &0,998 & 47,935&	0,998\\
L$0$ & 35,388 & 0,976	&37,693 & 0,980  &38,799 & 0,981   \\
d-LCI & 55,741& \textbf{1,000} &  \textbf{$\infty$} & \textbf{1,000} & 58,069 & \textbf{1,000}\\
 d-VPI &  \textbf{57,780}	&	 \textbf{1,000}  &\textbf{$\infty$}	&	 \textbf{1,000}& \textbf{ 63,595}&	 \textbf{1,000}\\
\hline
\end{tabular}
}
\end{table}

From the displayed average results, we observe the following.
\vspace{.3cm}\newline
$\Box$ Concerning upscaling:
\begin{enumerate}
\item[u.1]
On the datasets displayed in Table \ref{up234}, employing the methods with BIC input images, u-VPI has a slightly higher performance than BIC and u-LCI in terms of the visual quality values;
\item[u.2]
On the DIV2k dataset, providing the input images from the datasets therein included, we observe that the previous trend for BIC, u-LCI, and u-VPI is confirmed. However, we also find SCN that gives the best visual quality values  when the input images are generated by BIC (i.e.,  belonging to DIV2k-T-B and DIV2k-V-B datasets). Nevertheless, in the case of input images classified  unknown (i.e., belonging to  DIV2k-T-u  and DIV2k-V-u datasets),  slightly higher performance values are given by u-VPI, except in one case. Indeed, SCN has the best performance  on DIV2k-T-u  when s=4. On the contrary, SCN always provides the lowest PSNR and SSIM values when the input images are classified as both difficult and mild (see Table \ref{DIV2k-up4}). In this case, u-VPI continues to provide the best quality values for the Train images (i.e.,  for input images in DIV2k-T-d and DIV2k-T-m). However, BIC outperforms it for the Valid images (i.e.,  with input images from DIV2k-V-d and DIV2k-V-m). Finally,  no change regards the comparison u-VPI / u-LCI, where u-VPI always gives slightly higher values.
\end{enumerate}
\begin{table}[!htbp]
\caption {Average of the optimal values of $\theta$ resulting  from supervised-VPI with BIC input images. The downscaling $:3$ case is missing since it is independent of $\theta$ }
\label{tab:0}
\scriptsize{
\begin{center}
\begin{tabular}{r|r|r|r|r|r}
\hline
  &{:2}  & {:4} &{x2} & {x3} & {x4}\\
\hline
\textbf {BSDS500} & 0,279&0,649 &0,198 &0,203 &0,543   \\
\textbf {13US} &0,250 &0,250 &0,135 &0,288 &0,142 \\
\textbf {NY17} &0,374 &0,688 &0,132 &0,150 &0,153  \\
\textbf {NY96} &0,363 &0,678 &0,146 &0,193 &0,183  \\
\textbf {Urban100} &0,251 &0,596 &0,127 &0,264 &0,135  \\
\textbf {Pexels300} &0,370 & 0,596&0,118 &0,122 &0,121 \\
\textbf {Set 5} & & &0,110 & 0,250&0,120  \\
\textbf {Set 14} & & &0,121 &0,108 &0,188  \\
\textbf {DIV2k-T-B} & & &0,113 &0,125 &0,125  \\
\textbf {DIV2k-V-B} & & &0,114 &0,123 &0,121  \\
\textbf {DIV2k-T-u} & & &0,607 &0,098 &0,717  \\
\textbf {DIV2k-V-u} & & &0,609 &0,093 &0,722  \\
\textbf {DIV2k-T-d} & & & & &0,894  \\
\textbf {DIV2k-V-d} & & & & &0,896  \\
\textbf {DIV2k-T-m} & & & & & 0,892 \\
\textbf {DIV2k-V-m} & & & & & 0,912 \\
\hline
\end{tabular}
\end{center}
}
\end{table}

\begin{figure}[!htbp]
\begin{center}
\includegraphics [height=4.5cm,width=8.5cm]{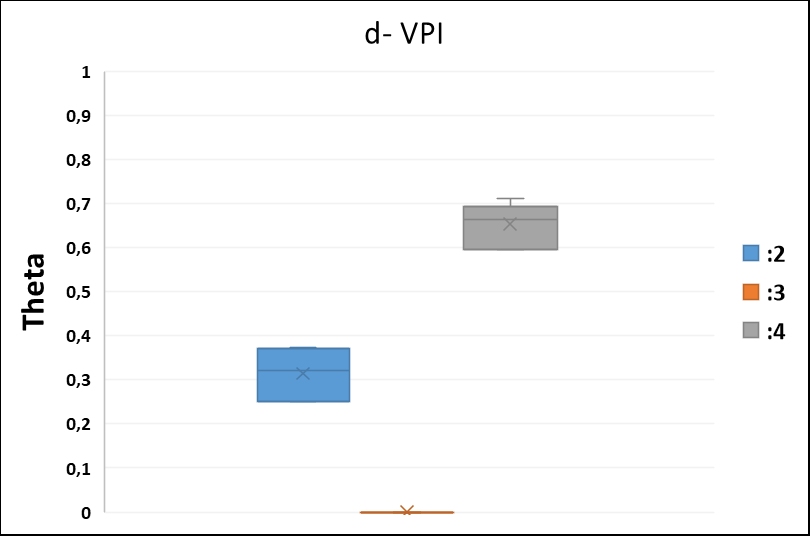}
\noindent
\newline
\newline
\includegraphics [height=4.5cm,width=8.5cm]{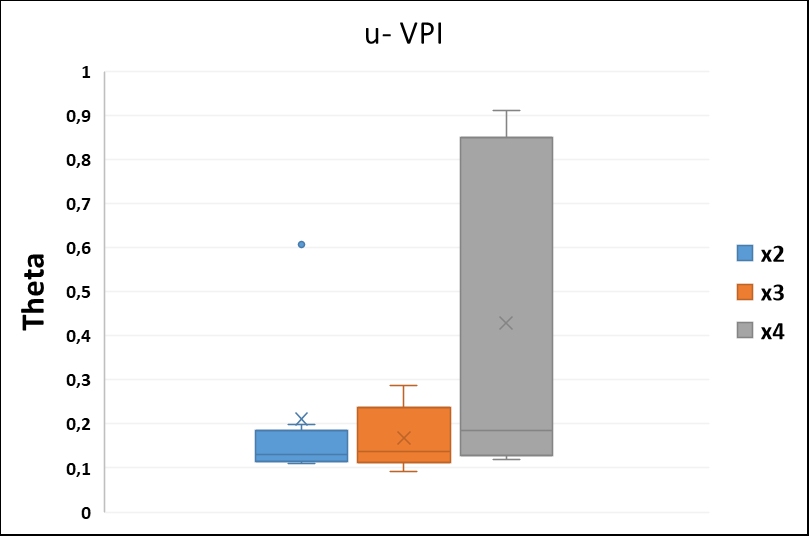}
\caption{Boxplot derived from Table \ref{tab:0}}
\label{fig:bp}
\end{center}
\end{figure}

\noindent
$\Box$ Concerning downscaling (with BIC input images):

\begin{enumerate}
 \item[d.1]
The best PSNR and SSIM values are achieved by d-VPI, followed in order by d-LCI ({\it ex-aequo} in some cases), DPID, BIC, and L$_0$. For all datasets  and scale factors displayed in Table \ref{up234}, there is a consistent gap between the PSNR values  by  d-VPI and those provided by BIC, DPID, and $L_0$. According to the results in \cite{Lagrange}, also d-LCI provides good values, but generally, d-VPI outperforms it, even if with a smaller gap.
  \item[d.2] The demo version of $L_0$ has memory problems with input images of large size. In fact, in the case of target images from $NY17$ and $NY96$ datasets, $L_0$ does not give any output for the scale factor $s=2,3$ that, according to (\ref{input-size}), generate a large input size.
  \item[d.3]
When $s=3$, d-VPI confirms the optimal performance proved in  Proposition 1 for odd scaling factors. We note that this case is missing in Figures \ref{fig:12a}--\ref{fig:11a} since (\ref{dim_ssim}) holds for both d-LCI and d-VPI. Nevertheless, we point out that  (\ref{dim_ssim})  has been proved under the ideal assumptions that the required LR sampling satisfies the Nyquist--Shannon theorem and that the initial data are not corrupted. Hence it is not always true. Indeed, we have verified (\ref{dim_ssim}) continues to hold starting from HR input images generated  by the Nearest-Neighbor and Bilinear methods \cite{NN-NIL2} (using Matlab \texttt{imresize} with \texttt{nearest} and \texttt{bilinear} option respectively), but in Subsection 5.2.3 we show it does not hold if we use SCN to generate the input HR image.
\end{enumerate}
 \subsubsection{Parameter modulation}
\label{sub4}
In this subsection, we use the previous quantitative analysis to hint to the user for setting the parameter $\theta$ in practice when the target image is unavailable.

 To this aim, in the previous tests employing supervised VPI with BIC input images, for each dataset, we compute  the average of   the optimal values of $\theta$ corresponding to the output images presenting the minimal MSE.  For both upscaling and downscaling, we report these results in Table \ref{tab:0} and show the relative boxplot in Figure \ref{fig:bp}. We remark that the downscaling case with scale factor $s=3$ is missing since, as previously stated, the d-VPI output is independent of $\theta$.

\begin{table}[!htbp]
\caption{Average CPU time in the upscaling cases of Table \ref{up234}}
\label{tempi-up}
\scriptsize{
\begin{center}
\begin{tabular}{r|r|r|r}
\hline\hline
& & &  \\
 & {x2}  & {x3}  & {x4} \\ \hline\hline
 \textbf{BSDS500} & & &  \\
BIC   &  0,003      &   0,002     &  0,003    \\
u-LCI &  0,014      &   0,010     &  0,008      \\
u-VPI &  0,013      &   0,012     &  0,010       \\
\hline
  \textbf {13US} & & &  \\
BIC   &   0,002      &    0,002    &        0,002       \\
u-LCI &    0,012     &   0,010     &        0,009  \\
u-VPI &    0,014     &   0,009     &        0,012 \\
\hline
 \textbf{NY17} & & &  \\
BIC   &  0,091     &   0,074     &    0,071  \\
u-LCI &  1,357     &    0,839    &    0,634  \\
u-VPI &  1,314     &   0,930     &   0,732             \\
\hline
  \textbf{NY96} & & &  \\
   BIC   &  0,056    &  0,055   &  0,051               \\
   u-LCI &  0,812     &   0,567 &  0,459                \\
   u-VPI &   0,864          &  0,606               &  0,487               \\
   \hline
  \textbf{URBAN100} & & &  \\
 BIC   &      0,009 	&  0,008  &  0,008                \\
 u-LCI &     0,051     &  0,051   &  0,059              \\
 u-VPI &     0,076        &   0,058          &     0,050             \\
 \hline
 \textbf{PEXELS300} & & &  \\
BIC   &   0,041& 0,037 & 0,035                \\
u-LCI & 0,300 & 0,216  &	0,185                   \\
u-VPI & 0,366            & 0,300             & 0,255                   \\
\hline
 \textbf{SET5} & & &  \\
BIC   &         0,003    &	  0,003    &	 0,002        \\
u-LCI &       0,014    &	  0,008   &	 0,006                 \\
u-VPI &       0,015      &  0,009      & 0,009                           \\
   \hline
  \textbf{SET14} & & &  \\
  BIC   &   0,003&  0,003   &	0,005            \\
  u-LCI &    0,020&  0,014   &	0,013                \\
  u-VPI &    0,022         &   0,017          &  0,015                      \\

\end{tabular}
\end{center}
}
\end{table}

\begin{table}[!htbp]
\caption{Average CPU time in the upscaling cases of Tables \ref{DIV2k-up234}--\ref{DIV2k-up4}}
\label{tempi-up-DIV2k}
\scriptsize{
\begin{center}
\begin{tabular}{r|r|r|r}
\hline\hline
& & &  \\
 & {x2}  & {x3}  & {x4} \\ \hline\hline

   \textbf{DIV2k-T-B} & & &  \\
   BIC   &   0,036 &	 0,031 &	 0,032    \\
   SCN   &   	14,353 &	 30,972 & 	17,879\\
   u-LCI &   	0,245  &	0,175  &	0,165    \\
   u-VPI &      0,318       & 0,225    & 0,187\\
       \hline
      \textbf{DIV2k-V-B} & & &  \\
   BIC   &   0,035  &		0,029 &	0,030       \\
   SCN   &   14,822 &	32,872    &	18,855    \\
   u-LCI &    	0,252 &	0,197  &	0,152       \\
   u-VPI &     0,315        &  0,234           &  0,192           \\
       \hline

         \textbf{DIV2k-T-u} & & &  \\
   BIC   &   0,035 &	0,025 &	 0,023        \\
   SCN   &   	14,442 &	29,500  &	 17,568    \\
   u-LCI &   	0,233 & 	0,176 &	0,149        \\
   u-VPI &      0,322       &  0,224           & 0,190             \\
       \hline

      \textbf{DIV2k-V-u} & & &  \\
   BIC   &     0,029 &	0,029 &0,027              \\
   SCN   &     14,976 &	32,270 &	18,362           \\
   u-LCI &     0,248 &		0,179  &	0,147            \\
   u-VPI &     0,333        &  0,231           & 0,214            \\

  \hline
      \textbf{DIV2k-T-d} & & &  \\
   BIC   &      &	 &  0,022 \\
   SCN   &     &	&	17,842  \\
   u-LCI &      &  &	 0,150           \\
   u-VPI &             &   & 0,206  \\

\hline
      \textbf{DIV2k-V-d} & & &  \\
   BIC   &      &	 &  0,023            \\
   SCN   &     &	&	17,608          \\
   u-LCI &      &		  &	  0,151          \\
   u-VPI &             &             & 0,211  \\

\hline
      \textbf{DIV2k-T-m} & & &  \\
   BIC   &      &	 &  0,024          \\
   SCN   &     &	&	18,532          \\
   u-LCI &      &		  &	 0,150           \\
   u-VPI &             &             & 0,206  \\

\hline
      \textbf{DIV2k-V-m} & & &  \\
   BIC   &      &	 &  0,024            \\
   SCN   &     &	&	 18,496         \\
   u-LCI &      &		  &	  0,150          \\
   u-VPI &             &             & 0,213  \\

\end{tabular}
\end{center}
}
\end{table}

\begin{table}[!htbp]
\caption{Average CPU time in downscaling tests of Table \ref{down234} (oom denotes "out of memory").}
\label{tempi-down}
\scriptsize{
\begin{center}
\begin{tabular}{r|r|r|r}
\hline\hline
& & &  \\
 & {:2}  & {:3}  & {:4} \\ \hline\hline
   \textbf{BSDS500} & & &  \\
 BIC &  0,006   & 0,009       & 0,017   \\
 DPID &  7,696  &   12,246    & 18,615 \\
 L0    &  3,647 & 8,020       & 14,295     \\
 d-LCI &  0,057 &  0,001      & 0,137 \\
 d-VPI & 0,046  & 0,001      & 0,117 \\
\hline
   \textbf{13US} & & &  \\
 BIC &    0,005  & 0,009   &   0,013   \\
 DPID &   5,593  & 8,905   &   13,819    \\
 L0    &  2,572  & 5,706   &   9,939      \\
 d-LCI &  0,042  & 0,001   &   0,108   \\
 d-VPI &  0,034  & 0,001  &   0,086     \\
\hline
  \textbf{NY17} & & &  \\
 BIC   &  0,229   &  0,325   &  0,639  \\
 DPID  & 448,023  & 731,498  &  1.098,113      \\
 L0    &  208,574 & 617,386  &    oom        \\
 d-LCI &  6,614   & 0,023   &   22,859      \\
 d-VPI & 7,246    &  0,023   & 	  20,194   \\
\hline
 \textbf{NY96} & & &  \\
BIC   & 0,155   &	 0,219    &    0,422\\
DPID  & 291,68   &	479,638   &   714,908\\
L0    & 133,19   & oom        &  om \\
d-LCI & 4,698    &	0,016     &	13,898\\
d-VPI & 4,704    & 0,016     &	13,949\\
\hline
 \textbf{URBAN100} & & &  \\
BIC   & {0,027}  &	 {0,041}   & 	  0,068\\
DPID  & 37,592   &	60,834     &     	93,996\\
L0    &	13,267   &  28,845     &        50,312 \\
d-LCI & 0,234    &	0,002      &		0,618\\
d-VPI & 0,259    & 0,002      &   	0,732\\
\hline
 \textbf{PEXELS300} & & &  \\
BIC   &    0,088 & 0,120 & 0,225   \\
DPID  & 151,800	&246,573 &374,155   \\
L0    & 	56,454  &125,413 &228,683     \\
d-LCI &    1,560 &	 0,009 &4,977          \\
d-VPI & 1,689 & 0,009 &   4,406\\
\end{tabular}
\end{center}
}
\end{table}

\subsubsection{CPU time analysis}
\label{sub5}
The CPU time required by each scaling method is also an important aspect to consider in the quantitative performance evaluation. For this reason, in the previous experiments, besides PSNR and SSIM, we measure the CPU time each method takes to produce the output images. Tables \ref{tempi-up}--\ref{tempi-down} show the average CPU time values we computed, for each scaling factor and dataset, by employing the displayed methods, in upscaling (Tables \ref{tempi-up}, \ref{tempi-up-DIV2k}) and downscaling (Table \ref{tempi-down}).

Regarding VPI, we point out that in Tables \ref{DIV2k-up234}--\ref{down234} we tested supervised VPI that is optimally structured to produce 19 resized images corresponding to unsupervised VPI called with 19 equidistant values of the input parameter $\theta$. For this reason, in comparison with the benchmark methods, we report in Tables \ref{tempi-up}--\ref{tempi-down} the average CPU time that unsupervised VPI takes, providing in input the average values of $\theta$ reported in Table \ref{tab:0} for each employed datasets and scale factors 2,3,4. For other values of $\theta$, we did not observe significant variations w.r.t. the displayed results.

Inspecting Tables \ref{tempi-up}--\ref{tempi-down}, we observe the following.
\vspace{.2cm}\newline
$\Box$ Concerning upscaling:
\vspace{.2cm}\newline
The method requiring the least CPU time is BIC. At a short distance, we find u-LCI and u-VPI with CPU times very close to each other. Much higher computation time is required by SCN, which always requires the longest CPU time. Table \ref{tempi-up-DIV2k} shows that this trend is independent of how the input image is generated (i.e., -B, -u, -d, -m).
\vspace{.2cm}\newline
$\Box$ Concerning downscaling:
\vspace{.2cm}\newline
Even in this case, the method requiring the least computation time is  BIC, closely followed by d-LCI and d-VPI that coincide and are much faster when the scale factor is 3, due to (\ref{RI-tilde}).
In the ranking, $L_0$ and DPID follow with much higher computation time than d-VPI,  especially on datasets characterized by larger image sizes (such as NY17, NY96, and PEXELS300). In particular, $L_0$ does not give any output for target images in NY96 and NY17 with scale factors $s\ge2$ and $s\ge 3$, respectively, while it is faster than DPID in the remaining cases.

\begin{table*}[!htbp]
\caption {Average performance results of upscaling methods (first column) on PEXELS300 dataset with input images generated by BIC, u-LCI, u-VPI, DPID and $L_0$ }
\label{cambioGin-up}
\scriptsize{
\begin{center}
\begin{tabular}{l||cc||cc||cc||cc||cc}
\hline
 &\multicolumn{2}{c|}{\bf BIC input image} & \multicolumn{2}{|c|}{\bf LCI input image} &
\multicolumn{2}{|c|}{\bf VPI input image}& \multicolumn{2}{|c|}{\bf DPID input image}
& \multicolumn{2}{|c}{\bf $L_0$ input image}
\\ \hline
& PSNR  & SSIM  & PSNR  & SSIM  & PSNR & SSIM & PSNR & SSIM & PSNR & SSIM \\
\hline
{\bf x2} &&&&&&&& \\
{BIC} &
36.249 & 0.962 & 36.274 & 0.963 & 37.047 & 0.968 & 36.336 & {\bf 0.966} & 35.194 & {\bf 0.964}  \\
{u-LCI} &
37.067 & 0.966 & 35.302 & 0.950 & 36.436 & 0.959 & 35.567 & 0.957 & 33.589 & 0.945 \\
{u-VPI} &
37.128 & 0.966 & {\bf 36.482} & {\bf 0.964} & {\bf 37.397} & {\bf 0.969} & {\bf 36.427} & {\bf 0.966} & {\bf 35.230} & {\bf 0.964} \\
{SCN} &
{\bf 37.916} & {\bf 0.971} & 33.939 & 0.945 & 35.184 & 0.957 & 34.325 & 0.953 & 31.818 & 0.936 \\
\hline
{\bf x3} &&&&&&&& \\
{BIC} &
33.147 & 0.932 & 32.409 & 0.929 & 32.409 & 0.929 & 32.988 & {\bf 0.935} & {\bf 32.055} & {\bf 0.929}  \\
{u-LCI} &
33.622 & 0.935 & 31.444 & 0.908 & 31.444 & 0.908 & 32.494 & 0.924 & 31.173 & 0.913 \\
{u-VPI} &
33.671 & 0.936 & {\bf 32.535} & {\bf 0.930} & {\bf 32.535} & {\bf 0.930} & {\bf 33.050} & {\bf 0.935} & 32.048 & 0.928 \\
{SCN} &
{\bf 34.462} & {\bf 0.944} & 30.245 & 0.904 & 30.245 & 0.904 & 31.849 & 0.926 & 30.145 & 0.906 \\
\hline
{\bf x4} &&&&&&&& \\
{BIC} &31.374 &0.908& 30.333 &0.903& 30.589 &0.907& 31.113 & {\bf 0.910}& 31.406 & {\bf 0.913} \\
{u-LCI} &31.741 &0.910& 29.414 &0.880& 29.721 &0.885& 30.689 &0.899& 31.001&0.901 \\
{u-VPI} &31.786 &0.912& {\bf 30.458} &{\bf 0.905}& {\bf 30.683} &{\bf 0.908}& {\bf 31.151} &{\bf 0.910}& {\bf 31.473}& {\bf 0.913} \\
{SCN} &{\bf 32.551} &{\bf 0.922}& 28.147 &0.872& 28.585 &0.882& 30.098 &0.902& 30.619 &0.907\\
\hline
\end{tabular}
\end{center}
}
\end{table*}

\begin{table*}[!htbp]
\caption {Average performance results of downscaling methods (first column) on BSDS500 dataset with input images generated by BIC, d-LCI, d-VPI and SCN}
\label{cambioGin-down}
\scriptsize{
\begin{center}
\begin{tabular}{l||cc||cc||cc||cc}
\hline\\
 &\multicolumn{2}{c}{\bf BIC input image} & \multicolumn{2}{|c}{\bf LCI input image} &
\multicolumn{2}{|c}{\bf VPI input image}& \multicolumn{2}{|c}{\bf SCN input image}
\\
\hline\hline
& PSNR  & SSIM  & PSNR  & SSIM  & PSNR & SSIM & PSNR & SSIM \\
\hline
{\bf :2} &&&&&&&& \\
{BIC} &
40.080 & 0.992 & 42.906 & 0.996 & 40.610 & 0.993 & {\bf 47.910} & {\bf 0.999}    \\
{d-LCI} &
54.840 & {\bf 1.000} & 57.392 & {\bf 1.000} & 58.509 & {\bf 1.000} & 36.931 & 0.987 \\
{d-VPI} &
{\bf 55.993} & {\bf 1.000} & {\bf 62.467} & {\bf 1.000} & {\bf 61.901} & {\bf 1.000} & 46.983 & {\bf 0.999}   \\
{DPID} &
41.065 & 0.991 & 40.256 & 0.990 & 40.942 & 0.991 & 37.173 & 0.985  \\
{$L_0$} &
37.539 & 0.990 & 36.943 & 0.989 & 37.250 & 0.989 & 32.057 & 0.966  \\
\hline
{\bf :3} &&&&&&&& \\
{BIC} &
40.452 & 0.993 & 43.325 & 0.996 & 40.860 & 0.994 & {\bf 47.453} & {\bf 0.999 }   \\
{d-LCI} &
$\mathbf\infty$ & {\bf 1.000} & $\mathbf\infty$ & {\bf 1.000} & $\mathbf\infty$ & {\bf 1.000} & 36.643 & 0.986 \\
{d-VPI} &
$\mathbf\infty$ & {\bf 1.000} & $\mathbf\infty$ & {\bf 1.000} & $\mathbf\infty$ & {\bf 1.000} & 36.643 & 0.986   \\
{DPID} &
42.282 & 0.994 & 40.981 & 0.992 & 42.021 & 0.994 & 38.212 & 0.988  \\
{$L_0$} &
36.828 & 0.987 & 36.348 & 0.985 & 36.680 & 0.986 & 33.027 & 0.969  \\
\hline
{\bf  :4} &&&&&&&& \\
{BIC} &
40.383 & 0.993 & 43.218 & 0.996 & 40.806 &0.994& {\bf 47.501} & {\bf 0.999}  \\
{d-LCI} &
56.846 &{\bf 1.000}& 59.981 &{\bf 1.000}& 60.319 &{\bf 1.000}& 35.911 & 0.984 \\
{d-VPI} &
{\bf 60.861} &{\bf 1.000}& $\mathbf\infty$ &{\bf 1.000}& {\bf 71.533} &{\bf 1.000}&  40.619 & 0.994 \\
{DPID} &
42.281 &0.994& 40.735 &0.993& 41.991 &0.994& 38.165 & 0.989  \\
{$L_0$} &
45.123 &0.997& 46.087 &0.998& 45.449 &0.998& 38.712 &0.991 \\
\hline
\end{tabular}
\end{center}
}
\end{table*}
\subsubsection{Input image dependency}
\label{sub3}
In this subsection, we study the dependency of the VPI performance on the way the input images are generated.   To this aim, we repeat the previous quantitative analysis computing the average PSNR and SSIM  values for supervised VPI and the benchmark methods, but we  let vary the scaling method generating the input images from the target ones in the dataset. More precisely, in downscaling we provide input HR images generated from BIC, u-LCI, u-VPI, and SCN upscaling methods, while in upscaling we get the input LR image by applying the downscaling methods BIC, d-LCI, d-VPI, DPID, and $L_0$. We point out that whenever u-VPI or d-VPI are employed to generate the input image, we use the unsupervised mode with the default value $\theta  =0.5$.

As above, we consider the scale factors $s=2,3,4$ and we require for the input images  the size $n_1\times n_2$,  determined by (\ref{input-size}), where $N_1\times N_2$ is the size of the target images in the dataset.

Since we have experimented that demo codes of $L_0$ and SCN have problems in processing images with a large size, we do not consider all datasets for this test, but we focus on PEXELS300 dataset in upscaling and on BSDS500 dataset in downscaling. The average performance results are shown in Tables \ref{cambioGin-up} and \ref{cambioGin-down}, respectively.

From these tables, we observe the following.
\vspace{.2cm}\newline
$\Box$ Concerning upscaling:

\begin{itemize}
\item SCN provides the highest quality measures only in the case of BIC input images, confirming the trend displayed in Table \ref{DIV2k-up234}.
\item In the case of input images generated by downscaling methods different from BIC, SCN always provides the lowest values and the best performance is attained by u-VPI except in the upscaling x4 case with $L_0$ input images when BIC presents  slightly higher performance values than u-VPI.
\item Similarly to BIC, u-VPI has a more stable behavior with respect to variations of the input image. The quality measures by u-VPI are always higher than those by u-LCI that behave better than BIC only with BIC input images.
\item In upscaling (x3), we note the same performance values for u-LCI and u-VPI input images (3th and 4th columns of Table \ref{cambioGin-up}), confirming that in downscaling (:3), both d-LCI and d-VPI generate the same input images.
\end{itemize}

$\Box$ Concerning downscaling:

\begin{itemize}
\item For even scale factors ($s=2,4$), d-VPI always provides much higher performance values than DPID and $L_0,$ which always presents the lowest quality measures. The d-VPI method followed by d-LCI achieves the highest performance, unless in the case of SCN input images where BIC holds the record, followed in order by d-VPI, DPID, d-LCI, and $ L_0 $
For odd scale factors, it is confirmed that d-VPI reduces to d-LCI reaching the optimal quality measures in the case of input images generated by BIC, u-LCI, or u-VPI. Nevertheless, for SCN input images, the ranking of the even cases $s = 2,4$ is confirmed, i.e., the best performance is given by BIC, followed in order by DPID, d-LCI = d-VPI, and $L_0.$
\end{itemize}

\subsection{Qualitative evaluation}
\label{PE}
\begin{figure*}[!htbp]
\begin{center}
\normalsize{  \hspace{2cm} \textbf{x2} \hspace{5cm} \textbf{x3}\hspace{5cm}\textbf{x4}}
\newline
\newline
\includegraphics[height=3.5cm,width=5.1cm]{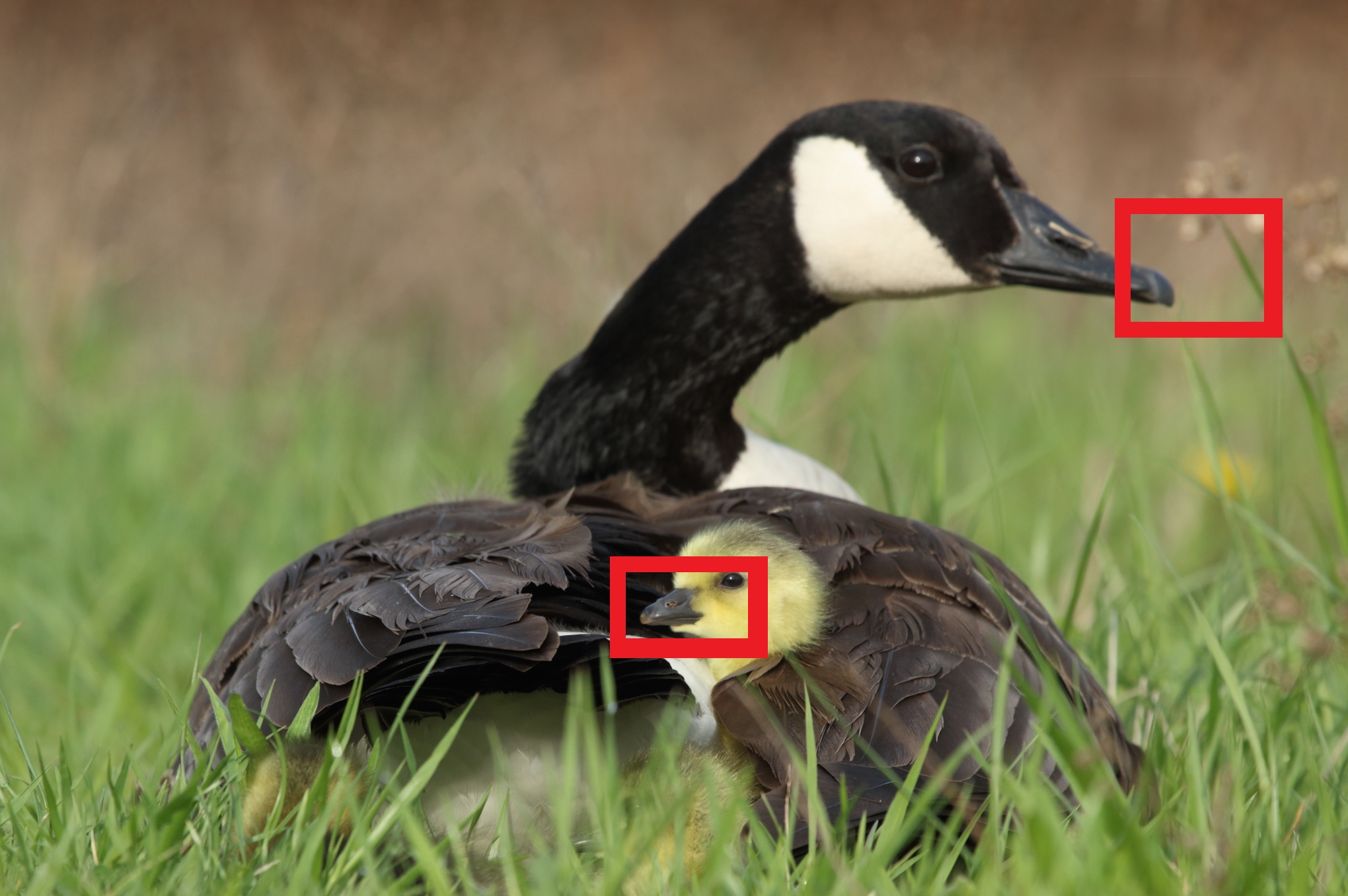}
\includegraphics[height=3.5cm,width=5.1cm]{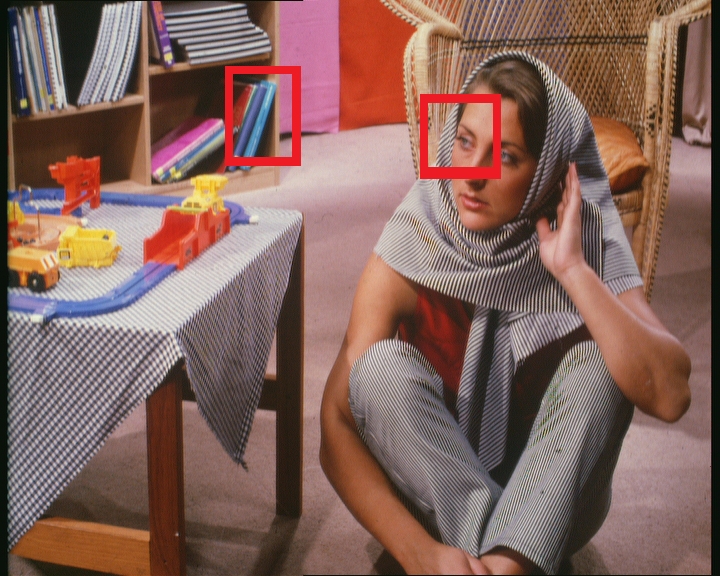}
\includegraphics[height=3.5cm,width=5.1cm]{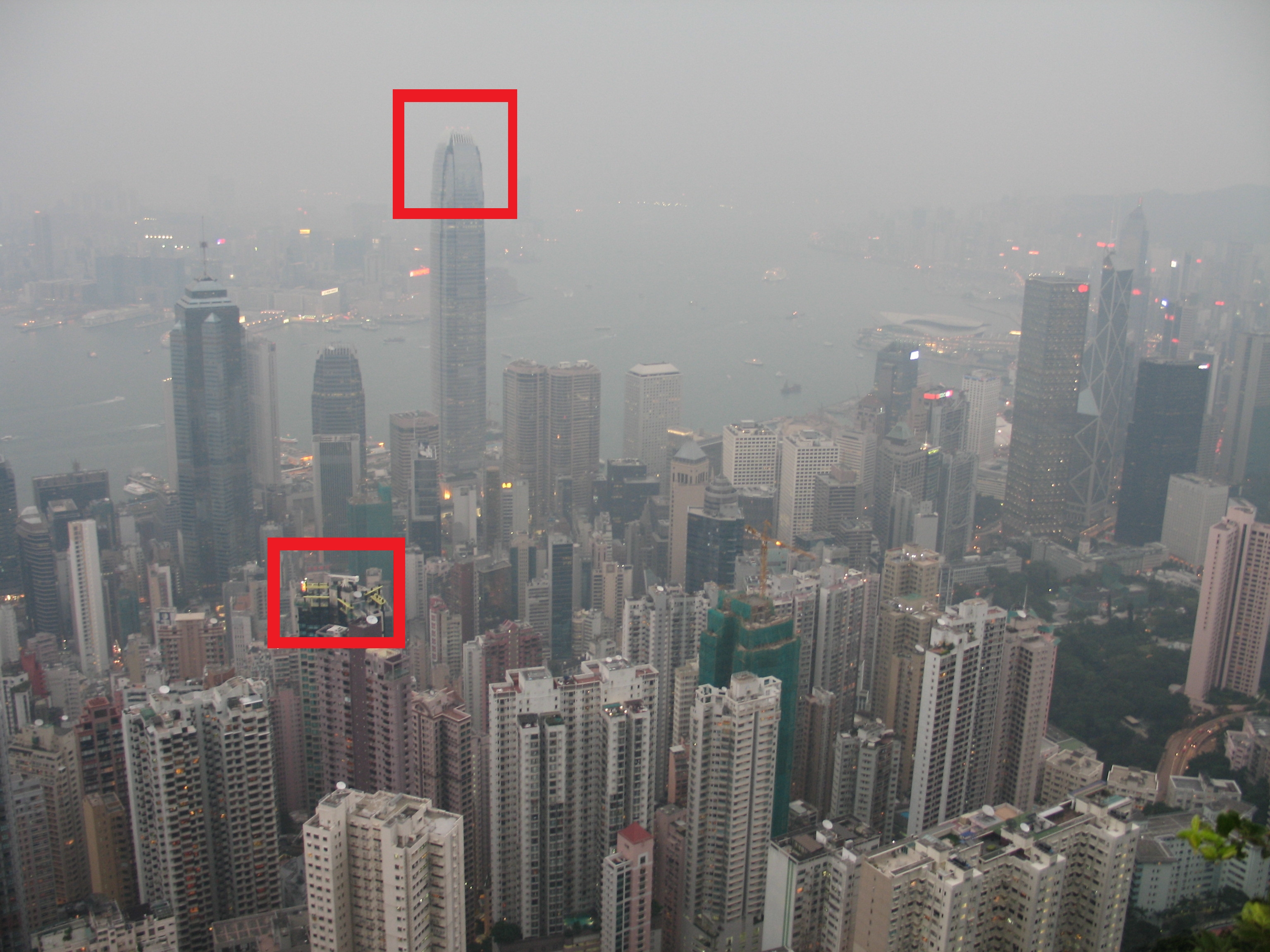}\\
\includegraphics[height=1cm, keepaspectratio]{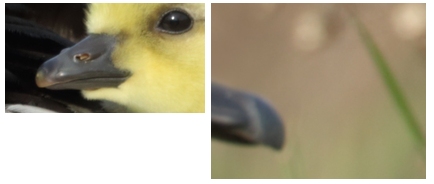} \hspace{3.2cm}
\includegraphics[height=1cm, keepaspectratio]{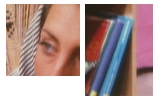}\hspace{3.3cm}
\includegraphics[height=1cm, keepaspectratio]{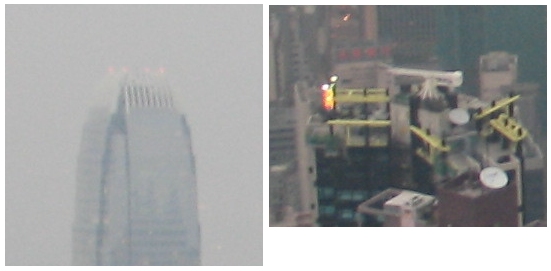}
\end{center}

\begin{tabular}{lll}
\hspace *{1.9cm}\textbf{Target image} & \hspace *{1.8cm} &\hspace *{1.8cm}\\
\hspace *{1.8cm} \textbf{0023 (1356$\times$2040) } & \hspace *{1.8cm} \textbf{S14.6 (576$\times$720)} &\hspace *{1.8cm} \textbf{N2 (2304$\times$3072)}\\
\hspace *{1.8cm} \textbf{from DIV2k} &  \hspace *{1.8cm} \textbf{from Set14} &\hspace *{1.8cm}  \textbf{from NY17}\\
\end{tabular}

\begin{center}
\includegraphics[height=3.5cm,width=5.1cm]{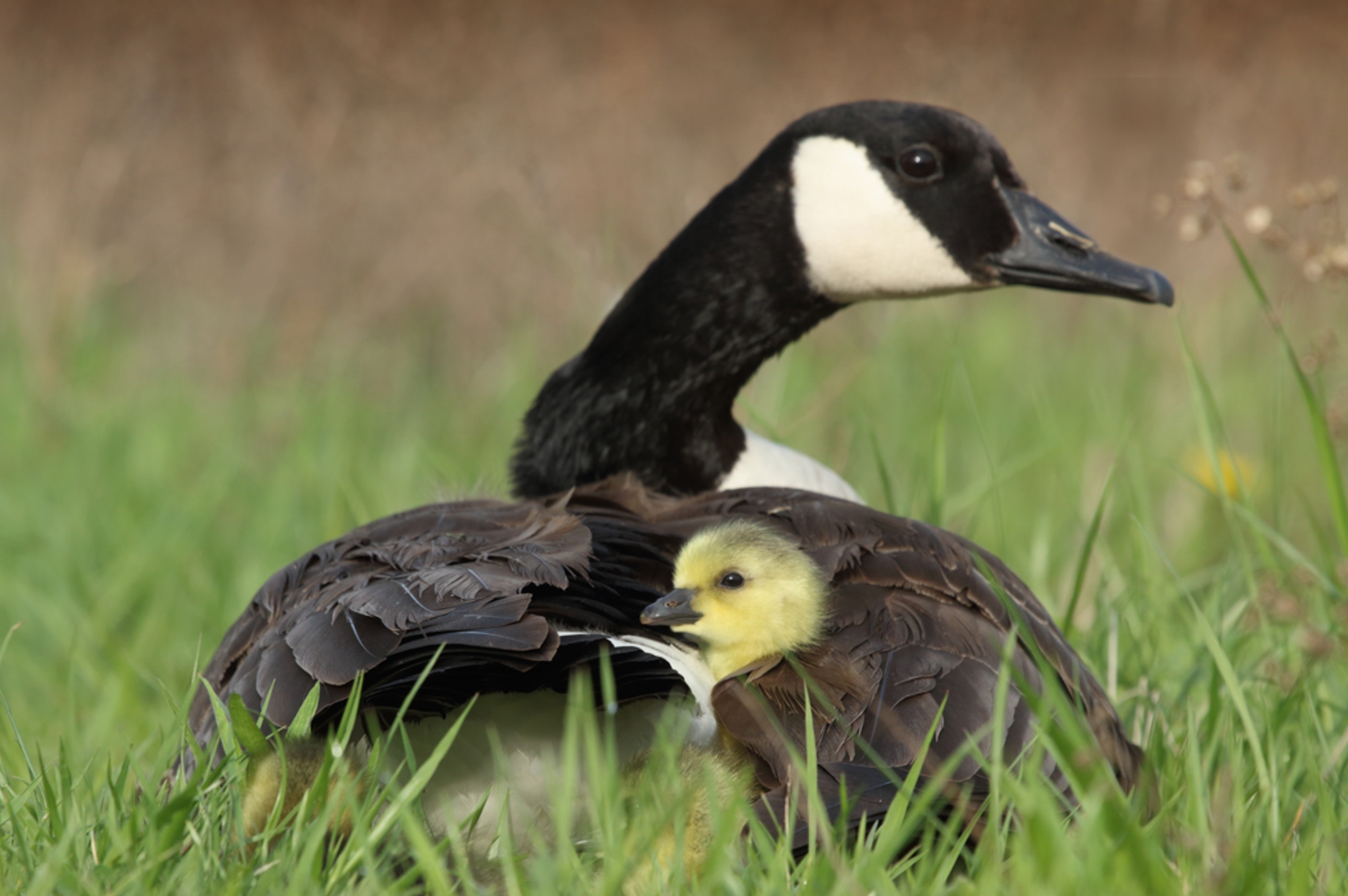}
\includegraphics[height=3.5cm,width=5.1cm]{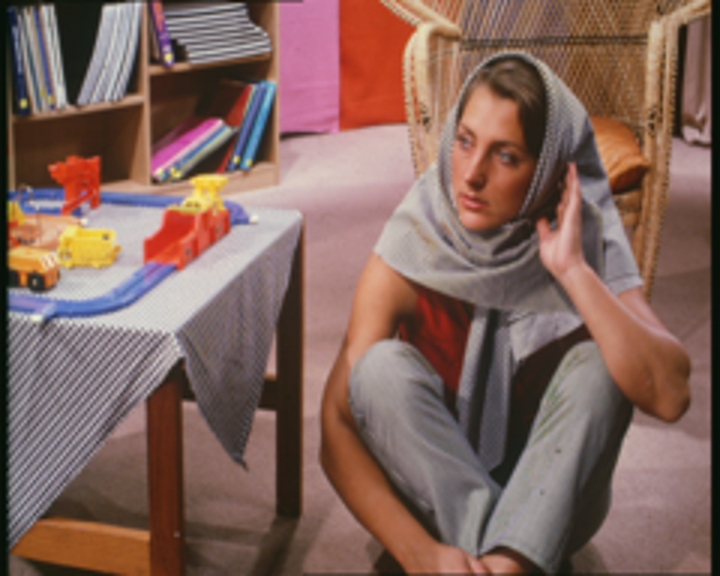}
 \includegraphics[height=3.5cm,width=5.1cm]{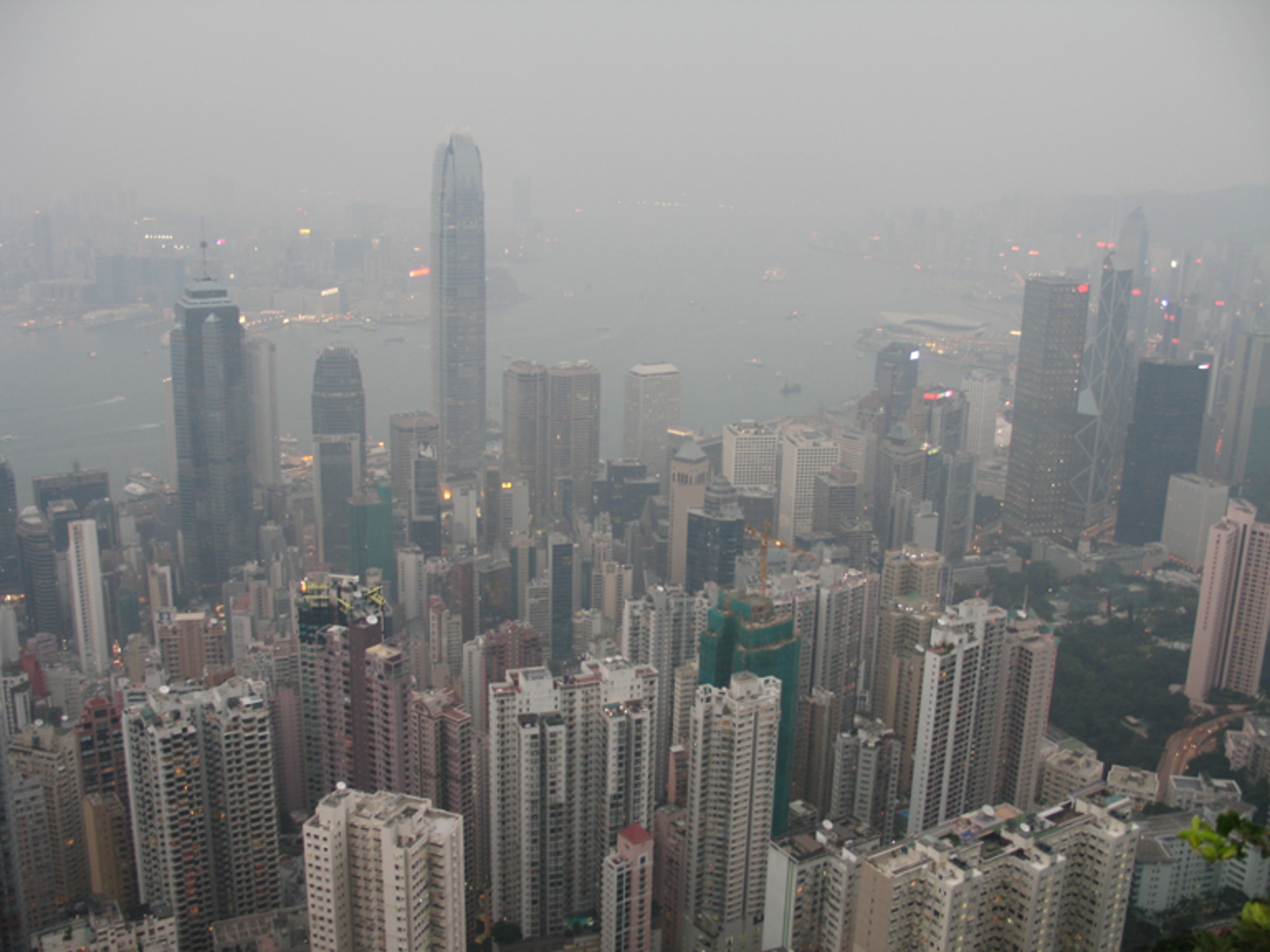}\\
\includegraphics[height=1cm, keepaspectratio]{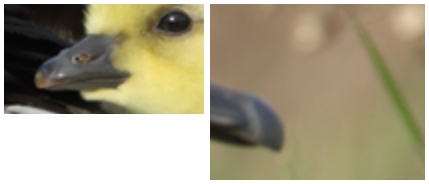} \hspace{3.2cm}
\includegraphics[height=1cm, keepaspectratio]{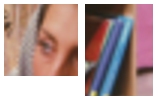}\hspace{3.3cm}
\includegraphics[height=1cm, keepaspectratio]{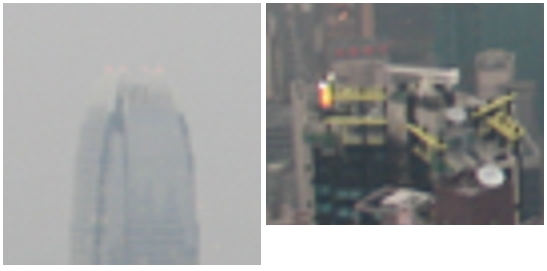}
\end{center}

\begin{tabular}{lll}
\hspace *{1.9cm}\textbf{BIC} & \hspace *{1.8cm} &\hspace *{1.8cm}\\
\hspace *{1.8cm} \textbf{PSNR=45,773} & \hspace *{2.2cm} \textbf{PSNR=26,204} &\hspace *{2.1cm} \textbf{ PSNR=39,220}\\
\hspace *{1.8cm} \textbf{SSIM=0,995} &  \hspace *{2.2cm} \textbf{SSIM=0,811} &\hspace *{2.2cm}  \textbf{SSIM=0,952}\\
\end{tabular}

\begin{center}
 \includegraphics[height=3.5cm,width=5.1cm]{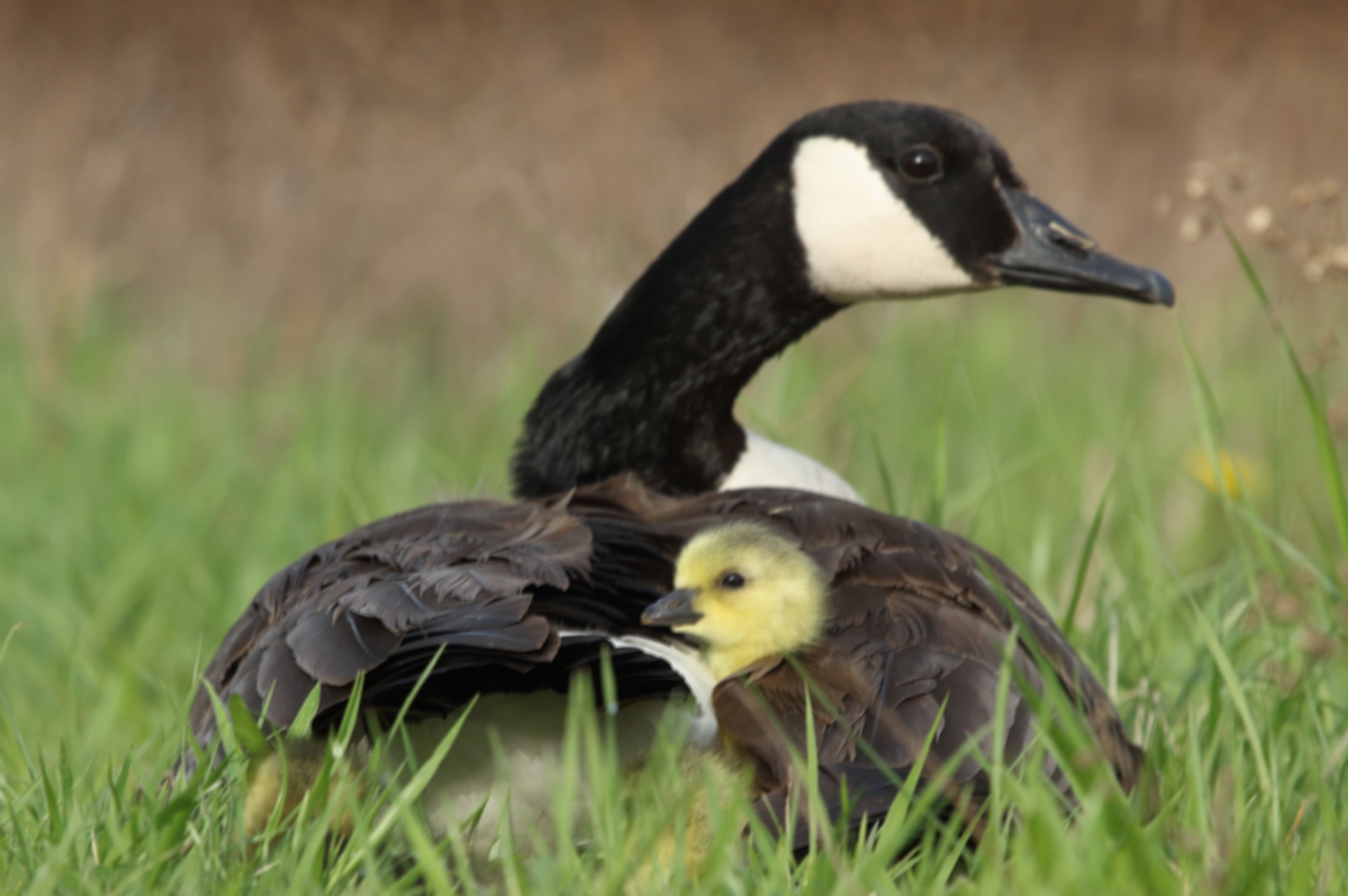}
\includegraphics[height=3.5cm,width=5.1cm]{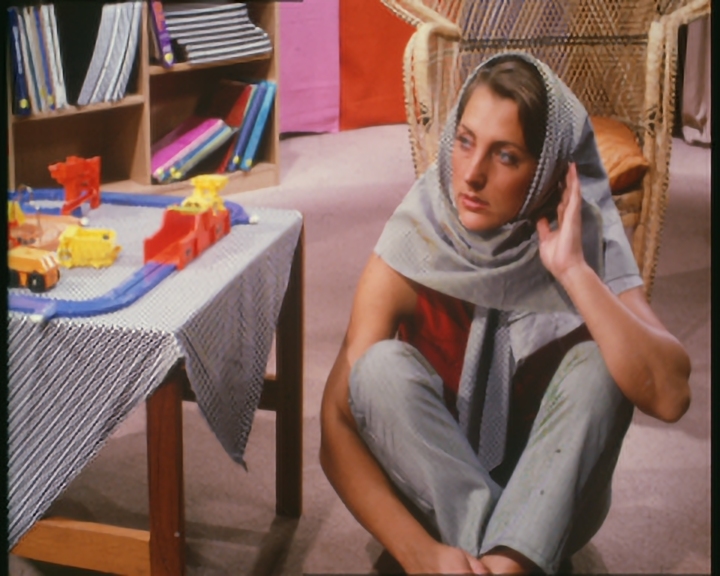}
\includegraphics[height=3.5cm,width=5.1cm]{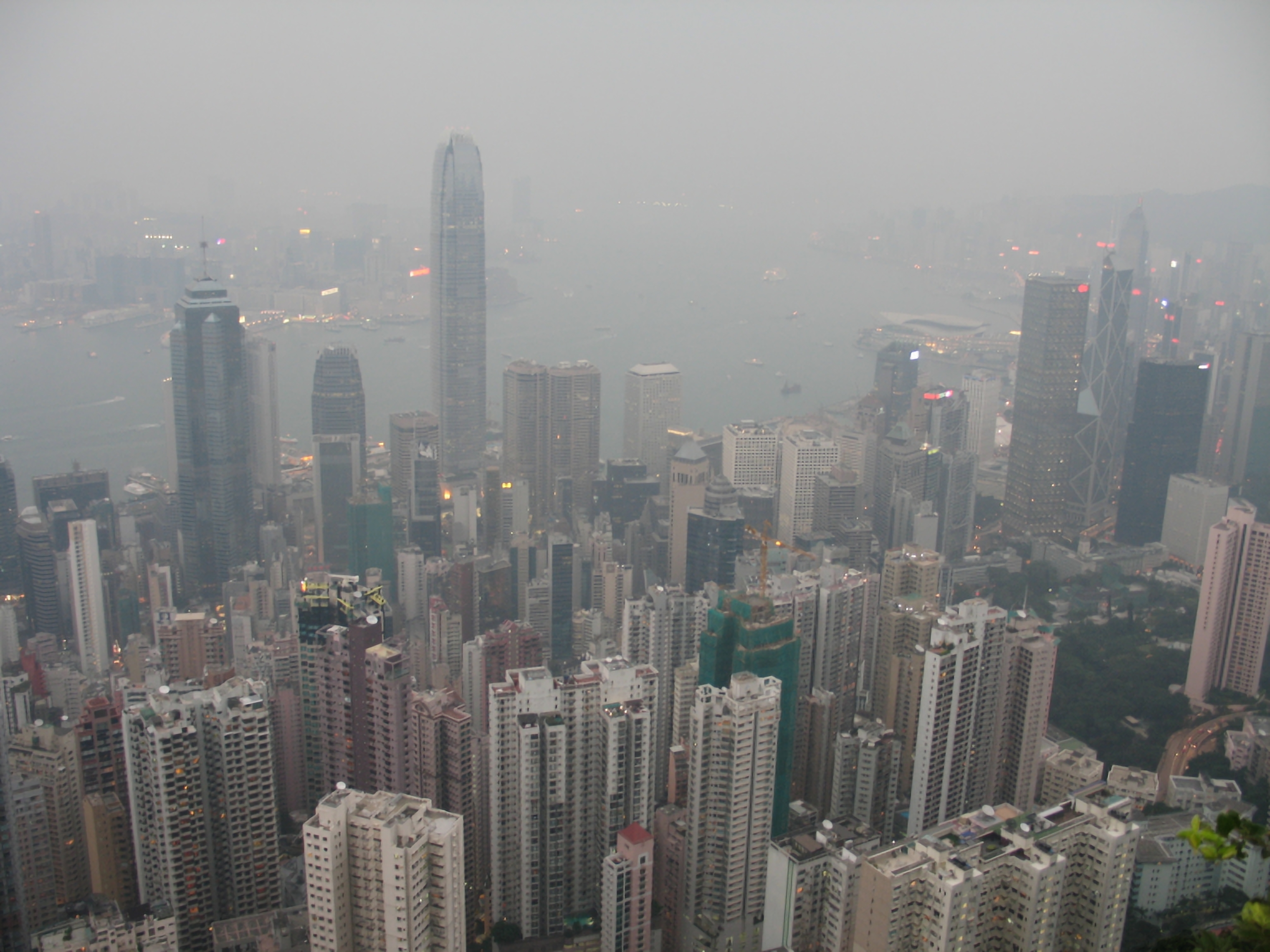}\\
\includegraphics[height=1cm, keepaspectratio]{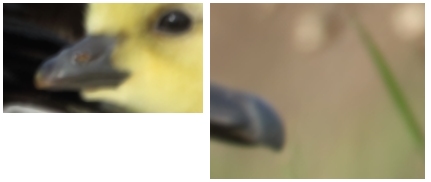} \hspace{3.2cm}
\includegraphics[height=1cm, keepaspectratio]{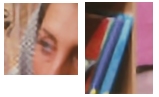}\hspace{3.3cm}
\includegraphics[height=1cm, keepaspectratio]{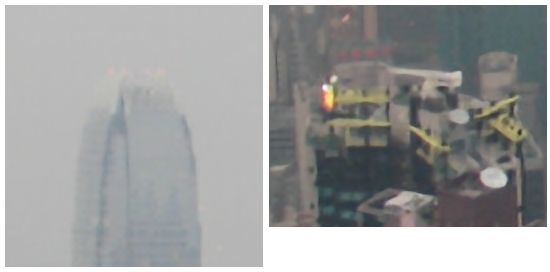}
\end{center}

\begin{tabular}{lll}
\hspace *{1.9cm}\textbf{SCN} & \hspace *{1.8cm} &\hspace *{1.8cm}\\
\hspace *{1.8cm} \textbf{PSNR=36,143} & \hspace *{2.2cm} \textbf{PSNR=26,274} &\hspace *{2.1cm} \textbf{ PSNR=31,637}\\
\hspace *{1.8cm} \textbf{SSIM=0,975} &  \hspace *{2.2cm} \textbf{SSIM=0,821} &\hspace *{2.2cm}  \textbf{SSIM=0,958}\\
\end{tabular}

\caption{Examples of supervised upscaling performance results at the scale factor 2 (left),  at the scale factor 3 (middle), at the scale factor 4 (right).}
\label{fig:6}
\end{figure*}

\begin{figure*}[!htbp]
\begin{center}
\normalsize{  \hspace{2cm} \textbf{x2} \hspace{5cm} \textbf{x3}\hspace{5cm}\textbf{x4}}
\newline
\newline
 \includegraphics[height=3.5cm,width=5.1cm]{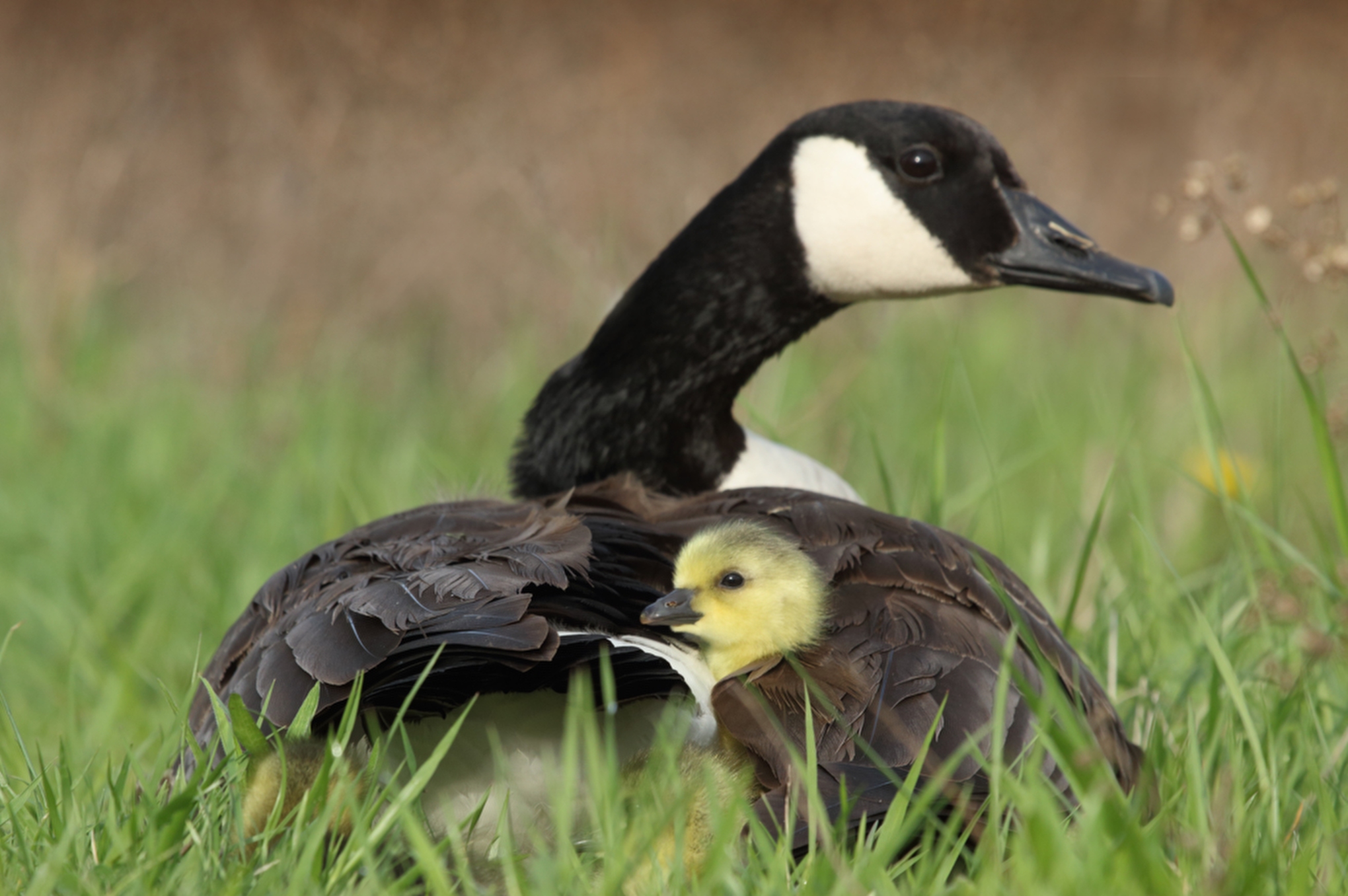}
\includegraphics[height=3.5cm,width=5.1cm]{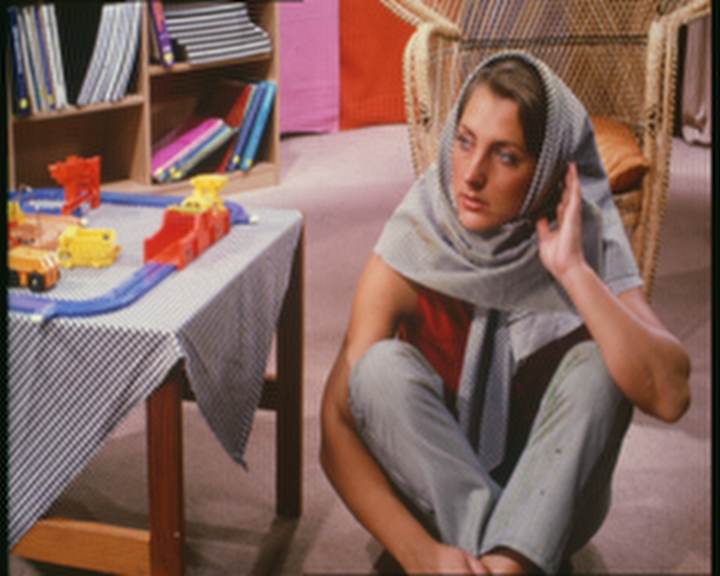}
\includegraphics[height=3.5cm,width=5.1cm]{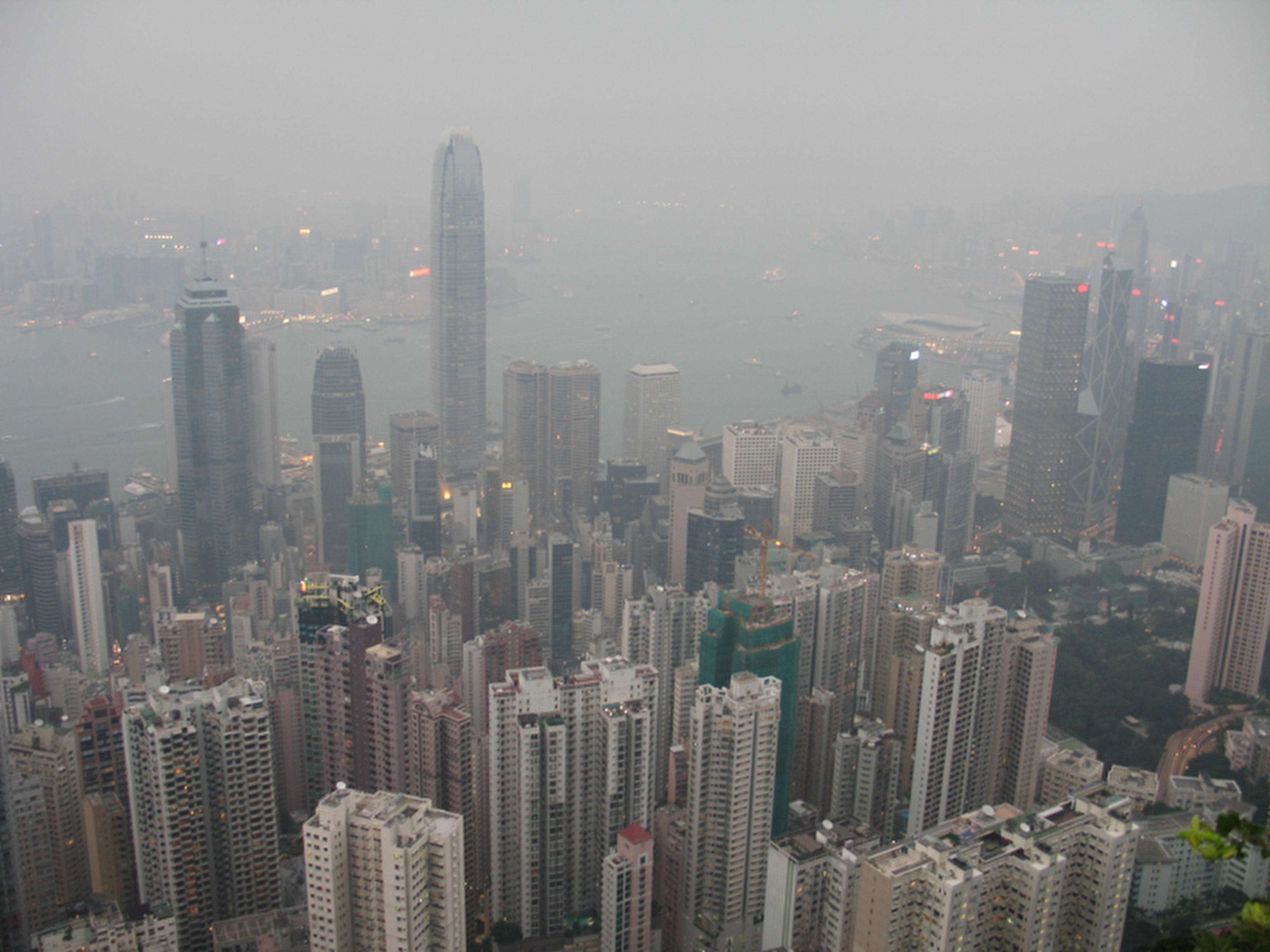}\\
\includegraphics[height=1cm, keepaspectratio]{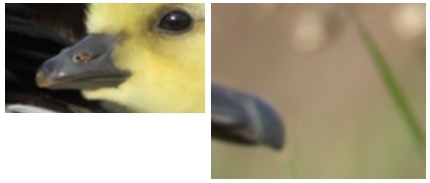} \hspace{3.2cm}
\includegraphics[height=1cm, keepaspectratio]{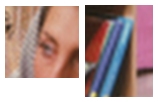}\hspace{3.3cm}
\includegraphics[height=1cm, keepaspectratio]{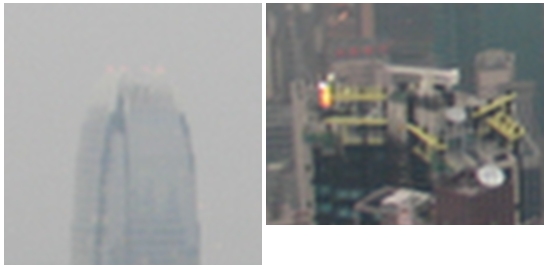}
\end{center}

\begin{tabular}{lll}
\hspace *{1.9cm}\textbf{LCI} & \hspace *{1.8cm} &\hspace *{1.8cm}\\
\hspace *{1.8cm} \textbf{PSNR=46,688} & \hspace *{2.1cm} \textbf{PSNR=26,585 } &\hspace *{2.1cm} \textbf{ PSNR=40,068}\\
\hspace *{1.8cm} \textbf{SSIM=0,996} &  \hspace *{2.1cm} \textbf{SSIM=0,822} &\hspace *{2.2cm}  \textbf{SSIM=0,957}\\
\end{tabular}

\begin{center}
 \includegraphics[height=3.5cm,width=5.1cm]{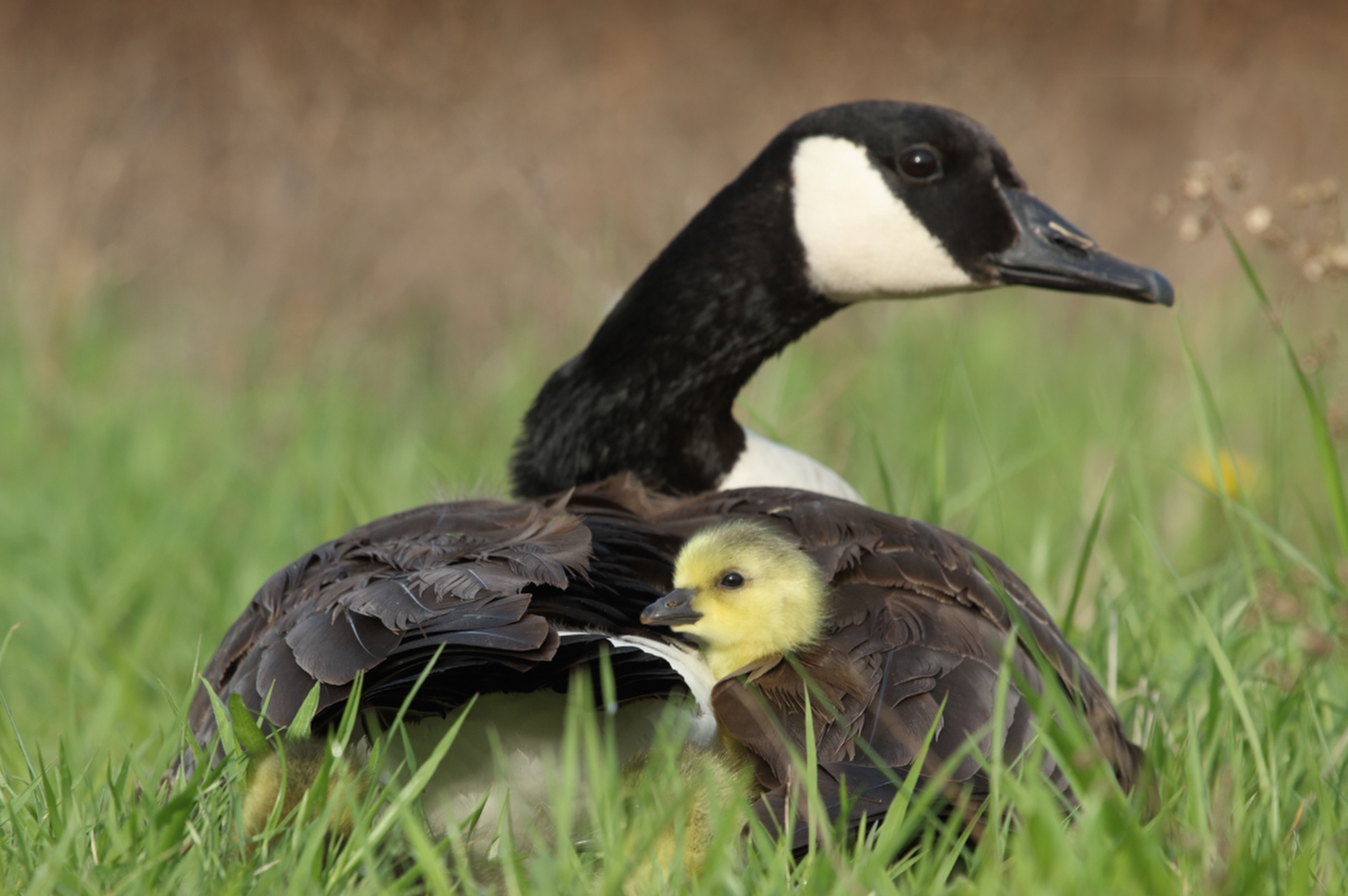}
\includegraphics[height=3.5cm,width=5.1cm]{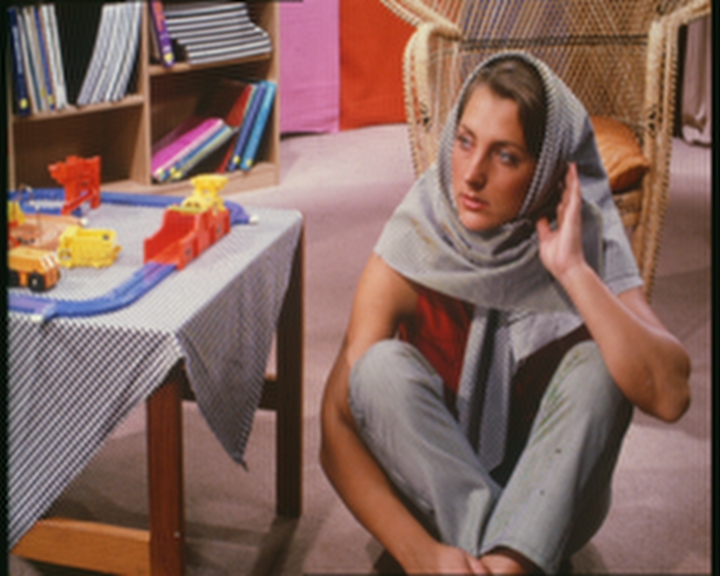}
\includegraphics[height=3.5cm,width=5.1cm]{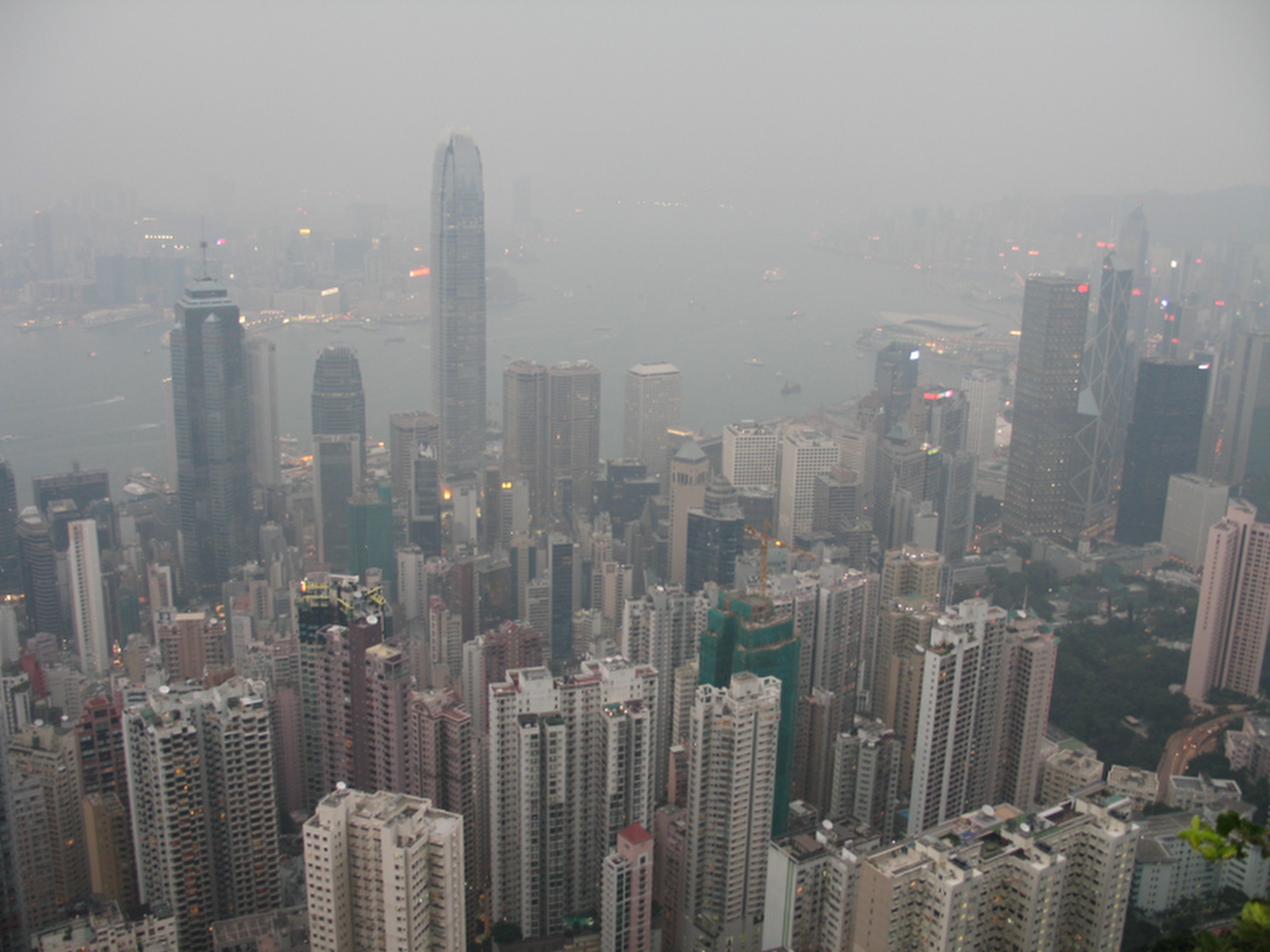}\\
 \includegraphics[height=1cm, keepaspectratio]{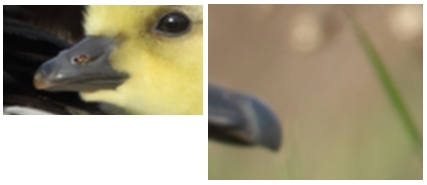} \hspace{3.2cm}
\includegraphics[height=1cm, keepaspectratio]{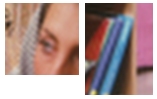}\hspace{3.3cm}
\includegraphics[height=1cm, keepaspectratio]{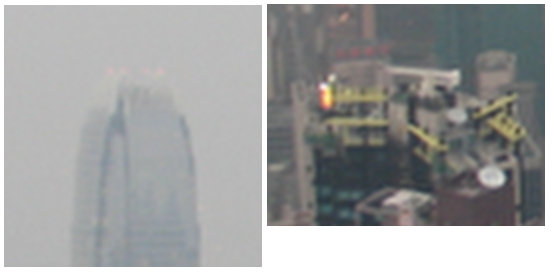}
\end{center}

\begin{tabular}{lll}
\hspace *{1.9cm}\textbf{VPI} & \hspace *{1.8cm} &\hspace *{1.8cm}\\
\hspace *{1.8cm} \textbf{PSNR=46,745} & \hspace *{2.1cm} \textbf{PSNR=26,590 } &\hspace *{2.1cm} \textbf{ PSNR=40,113}\\
\hspace *{1.8cm} \textbf{SSIM=0,996} &  \hspace *{2.1cm} \textbf{SSIM=0,822} &\hspace *{2.2cm}  \textbf{SSIM=0,957}\\
\end{tabular}

\caption{Examples of supervised upscaling performance results at the scale factor 2 (left),  at the scale factor 3 (middle), at the scale factor 4 (right).}
\label{fig:7}
\end{figure*}

\begin{figure*}[!htbp]
\begin{center}
\normalsize{  \hspace{2cm} \textbf{:2} \hspace{5cm} \textbf{:3}\hspace{5cm}\textbf{:4}}
\newline
\newline
\includegraphics[height=3.5cm, width=5.1cm]{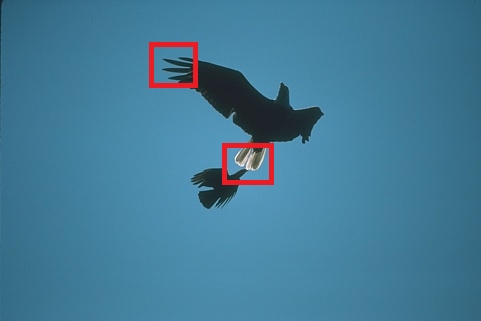}
\includegraphics[height=3.5cm, width=5.1cm]{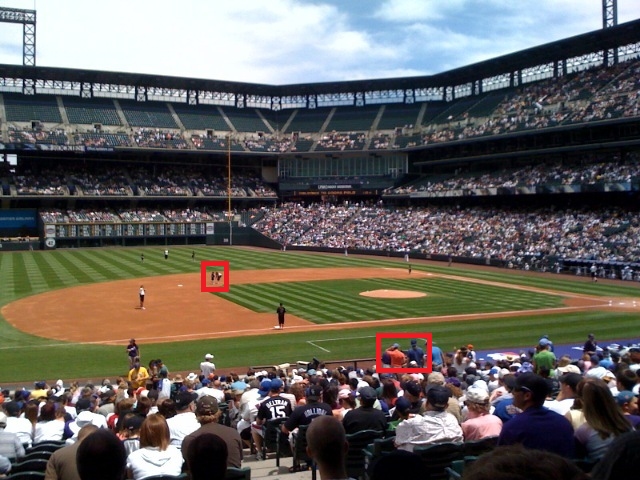}
\includegraphics[height=3.5cm, width=5.1cm]{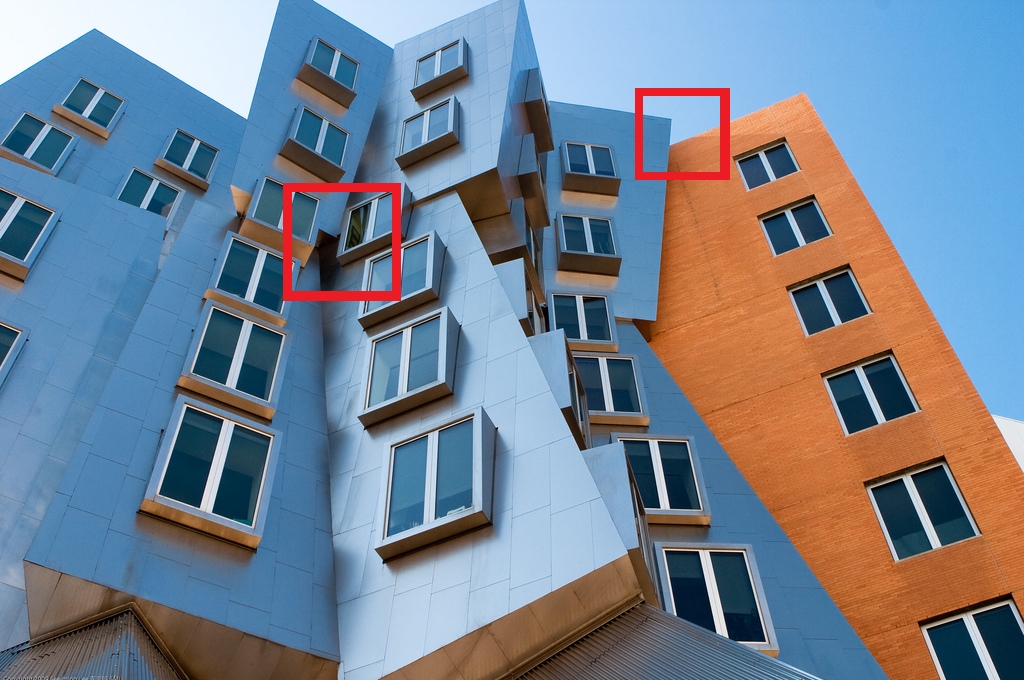}\\
\includegraphics[height=1cm, keepaspectratio]{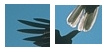} \hspace{3.2cm}
\includegraphics[height=1cm, keepaspectratio]{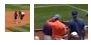}\hspace{3.3cm}
\includegraphics[height=1cm, keepaspectratio]{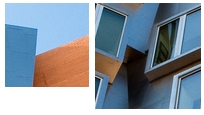}
\end{center}

\begin{tabular}{lll}
\hspace *{1.9cm}\textbf{Target image} & \hspace *{1.8cm} &\hspace *{1.8cm}\\
\hspace *{1.8cm} \textbf{135069 (321$\times$481)} & \hspace *{1.8cm} \textbf{NA38 (480$\times$640)} &\hspace *{1.8cm} \textbf{img075 (680$\times$1024)}\\
\hspace *{1.8cm} \textbf{from BSDS500} &  \hspace *{1.8cm} \textbf{from NY96} &\hspace *{1.8cm}  \textbf{from Urban100}\\
\end{tabular}

\begin{center}
\includegraphics[height=3.5cm,width=5.1cm]{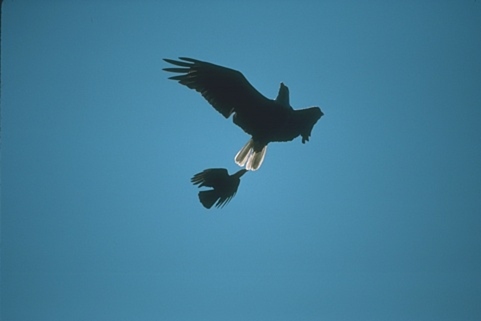}
\includegraphics[height=3.5cm, width=5.1cm]{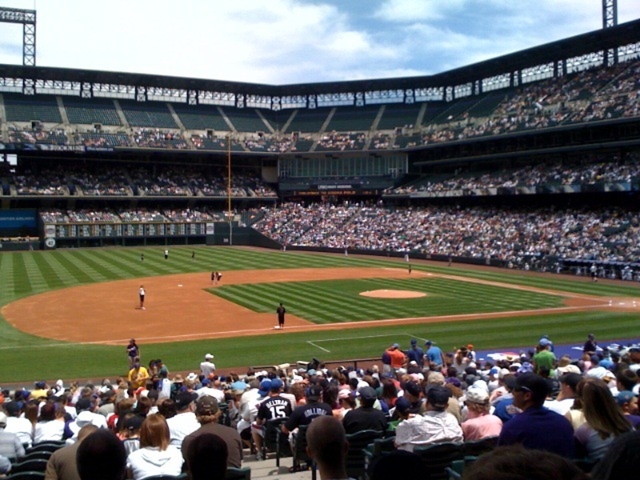}
\includegraphics[height=3.5cm, width=5.1cm]{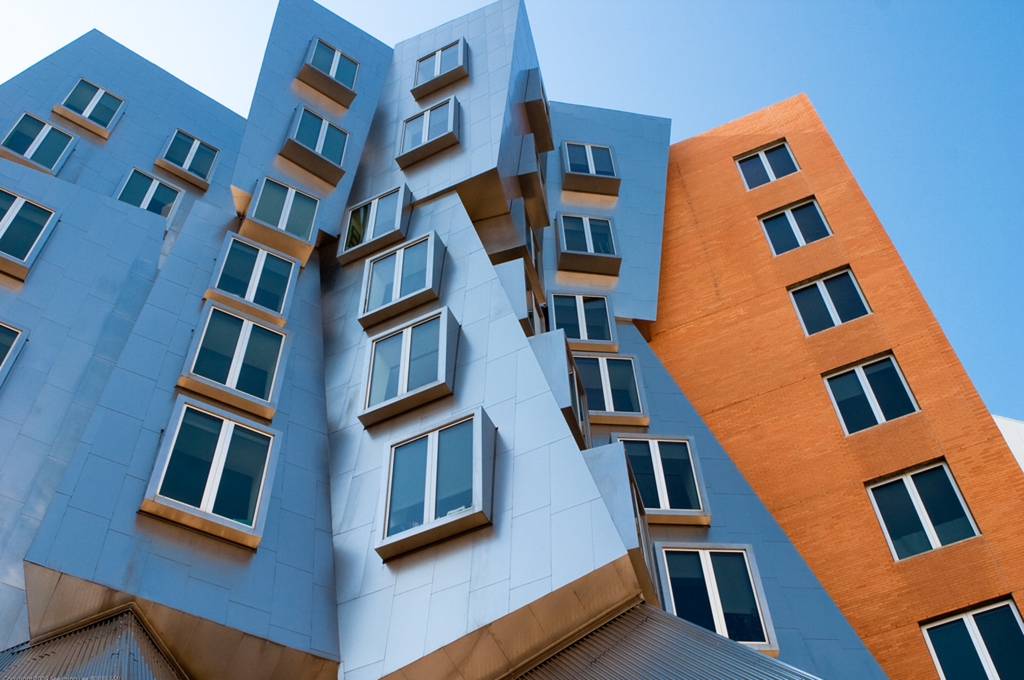}\\
\includegraphics[height=1cm, keepaspectratio]{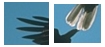} \hspace{3.2cm}
\includegraphics[height=1cm, keepaspectratio]{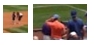}\hspace{3.3cm}
\includegraphics[height=1cm, keepaspectratio]{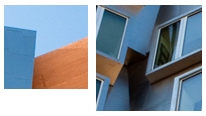}
\end{center}

\begin{tabular}{lll}
\hspace *{1.9cm}\textbf{BIC} & \hspace *{1.8cm} &\hspace *{1.8cm}\\
\hspace *{1.8cm} \textbf{PSNR=50,266} & \hspace *{2.1cm} \textbf{PSNR=39,885} &\hspace *{2.3cm} \textbf{PSNR=41,035}\\
\hspace *{1.8cm} \textbf{SSIM=1,000} &  \hspace *{2.1cm} \textbf{SSIM=0,995} &\hspace *{2.3cm}  \textbf{SSIM=0,997}\\
\end{tabular}

\begin{center}
\includegraphics[height=3.5cm,width=5.1cm]{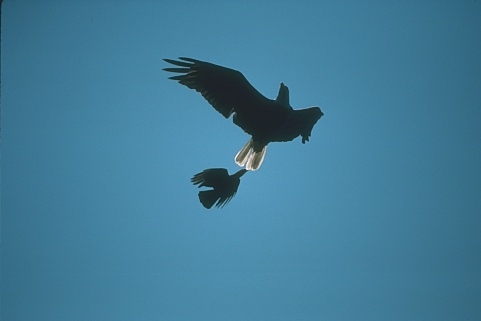}
\includegraphics[height=3.5cm,width=5.1cm]{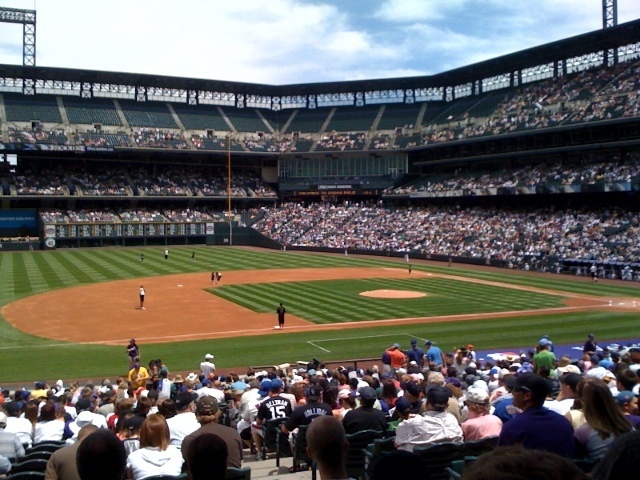}
\includegraphics[height=3.5cm,width=5.1cm]{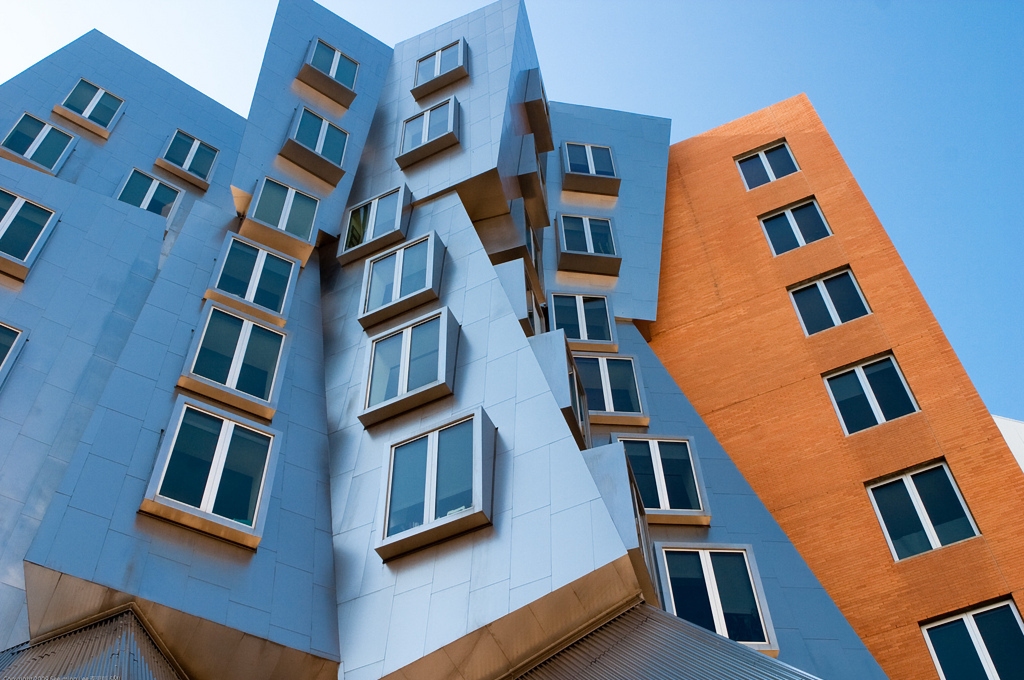}\\
\includegraphics[height=1cm, keepaspectratio]{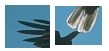}\hspace{3.2cm}
\includegraphics[height=1cm, keepaspectratio]{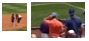}\hspace{3.3cm}
\includegraphics[height=1cm, keepaspectratio]{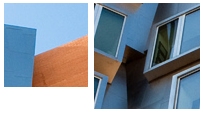}
\end{center}

\begin{tabular}{lll}
\hspace *{1.9cm}\textbf{DPID} & \hspace *{1.8cm} &\hspace *{1.8cm}\\
\hspace *{1.8cm} \textbf{PSNR=54,342 } & \hspace *{2.1cm} \textbf{PSNR=42,037} &\hspace *{2.3cm} \textbf{ PSNR=44,545}\\
\hspace *{1.8cm} \textbf{SSIM=1,000} &  \hspace *{2.1cm} \textbf{SSIM=0,997} &\hspace *{2.4cm}  \textbf{SSIM=0,999}\\
\end{tabular}

\caption{Examples of supervised downscaling performance results at the scale factor 2 (left),  at the scale factor 3 (middle), at the scale factor 4 (right). }
\label{fig:8}

\end{figure*}

\begin{figure*}[!htbp]
\begin{center}
\normalsize{  \hspace{2cm} \textbf{:2} \hspace{5cm} \textbf{:3}\hspace{5cm}\textbf{:4}}
\newline
\newline
\includegraphics[height=3.5cm,width=5.1cm]{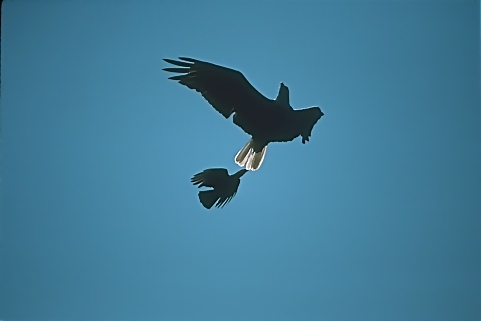}
\includegraphics[height=3.5cm,width=5.1cm]{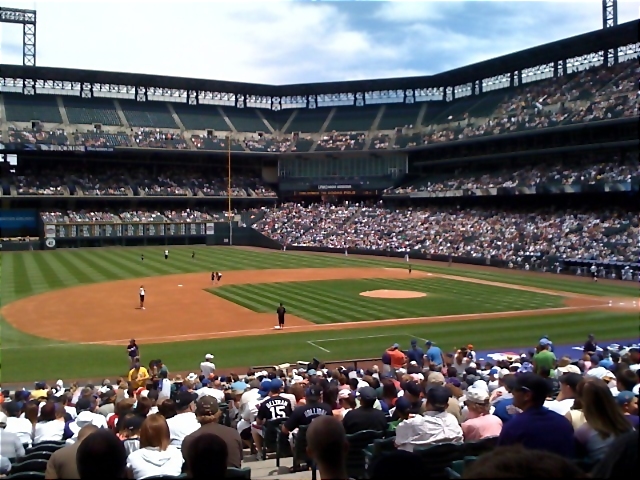}
\includegraphics[height=3.5cm,width=5.1cm]{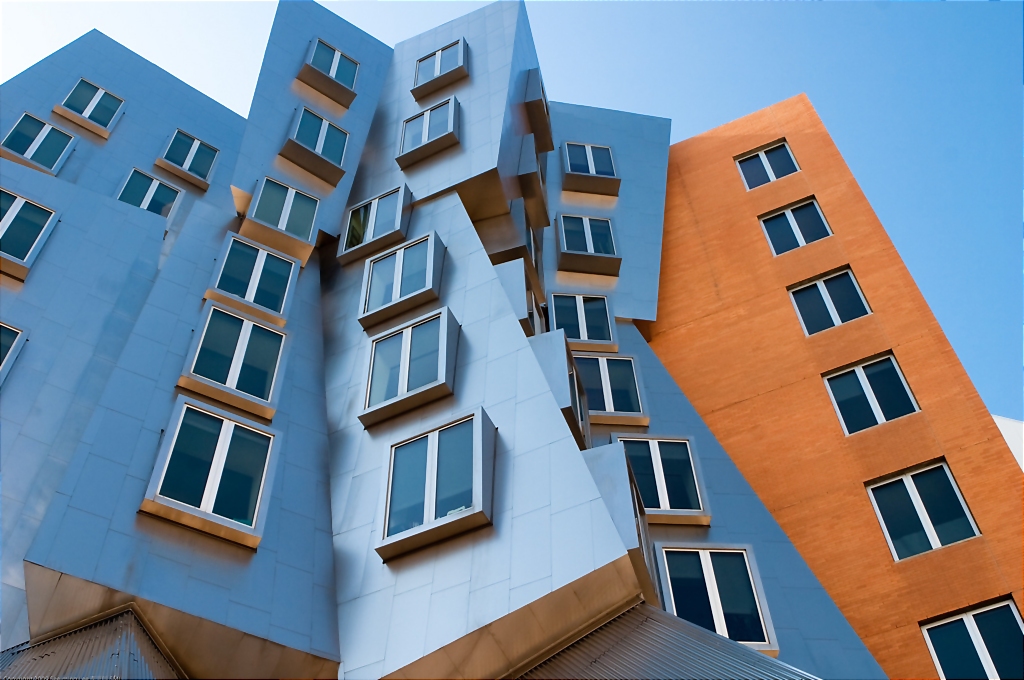}\\
\includegraphics[height=1cm, keepaspectratio]{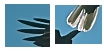}\hspace{3.2cm}
\includegraphics[height=1cm, keepaspectratio]{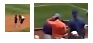}\hspace{3.3cm}
\includegraphics[height=1cm, keepaspectratio]{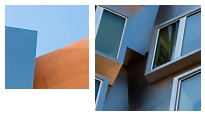}
\end{center}

\begin{tabular}{lll}
\hspace *{1.8cm}\textbf{ L$_0$} & \hspace *{1.8cm} &\hspace *{1.8cm}\\
\hspace *{1.8cm} \textbf{PSNR=39,771 } & \hspace *{2.1cm} \textbf{PSNR=33,064} &\hspace *{2.1cm} \textbf{ PSNR=37,198}\\
\hspace *{1.8cm} \textbf{SSIM=0,997} &  \hspace *{2.1cm} \textbf{SSIM=0,971} &\hspace *{2.2cm}  \textbf{SSIM=0,991}\\
\end{tabular}

\begin{center}
\includegraphics[height=3.5cm,width=5.1cm]{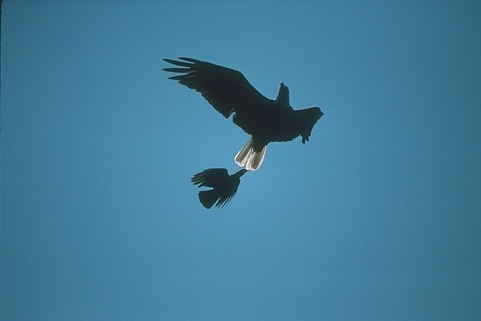}
\includegraphics[height=3.5cm,width=5.1cm]{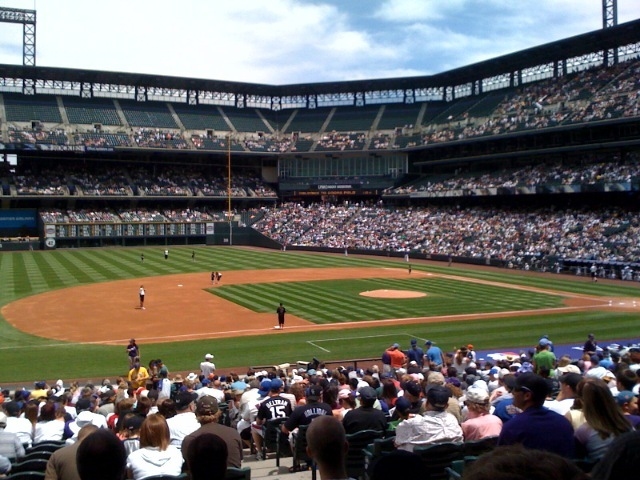}
\includegraphics[height=3.5cm,width=5.1cm]{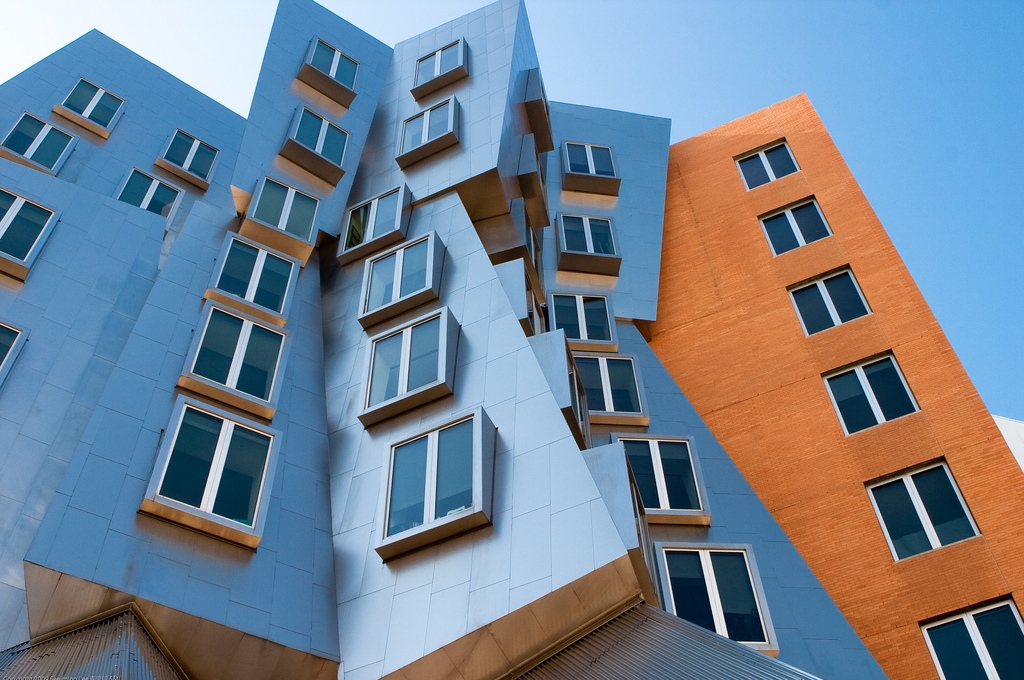}\\
\includegraphics[height=1cm, keepaspectratio]{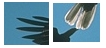}\hspace{3.2cm}
\includegraphics[height=1cm, keepaspectratio]{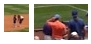}\hspace{3.3cm}
\includegraphics[height=1cm, keepaspectratio]{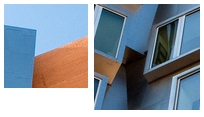}
\end{center}

\begin{tabular}{lll}
\hspace *{1.9cm}\textbf{LCI} & \hspace *{1.8cm} &\hspace *{1.8cm}\\
\hspace *{1.8cm} \textbf{PSNR=56,984} & \hspace *{2.1cm} \textbf{PSNR={$\infty$} } &\hspace *{2.3cm} \textbf{ PSNR=57,600}\\
\hspace *{1.8cm} \textbf{SSIM=1,000} &  \hspace *{2.1cm} \textbf{SSIM=1,000} &\hspace *{2.4cm}  \textbf{SSIM=1,000}\\
\end{tabular}

\begin{center}
\includegraphics[height=3.5cm,width=5.1cm]{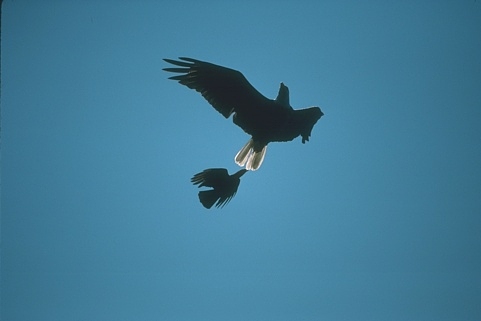}
\includegraphics[height=3.5cm,width=5.1cm]{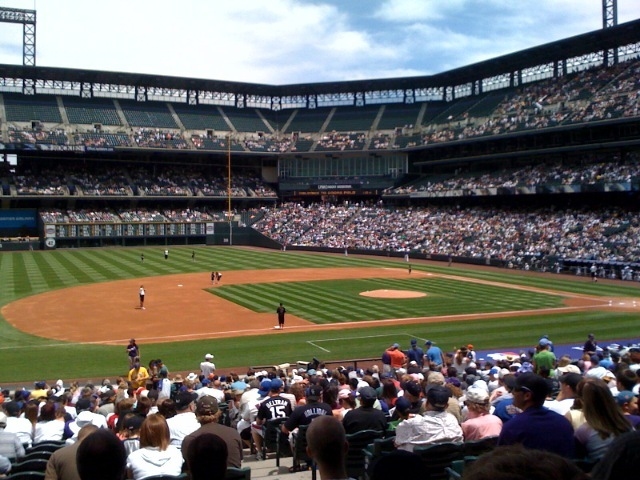}
\includegraphics[height=3.5cm,width=5.1cm]{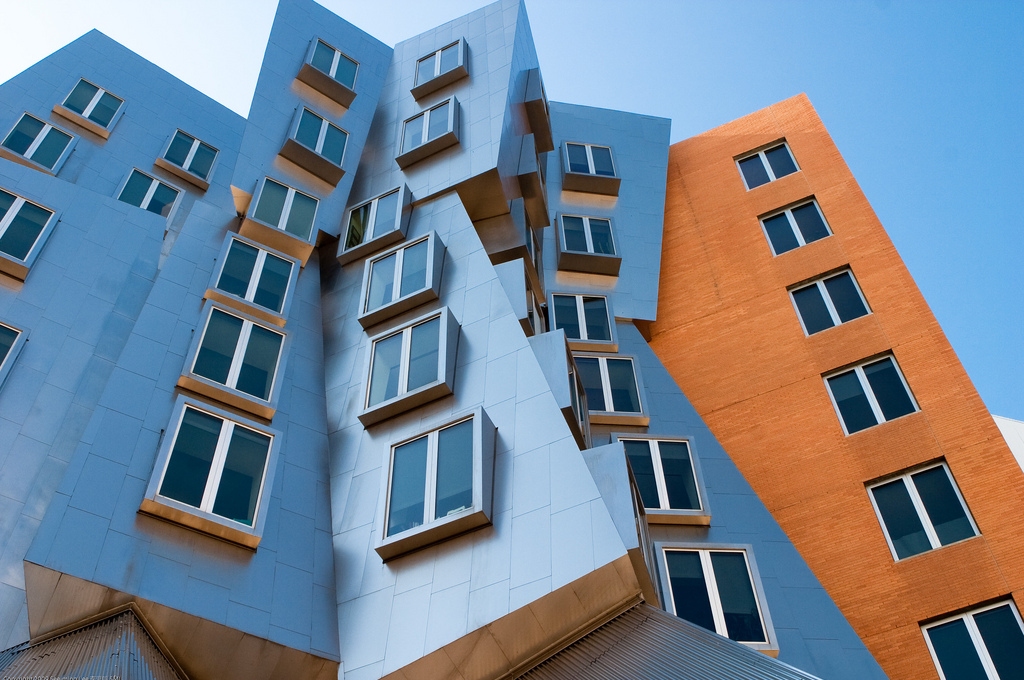}\\
\includegraphics[height=1cm, keepaspectratio]{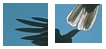}\hspace{3.2cm}
\includegraphics[height=1cm, keepaspectratio]{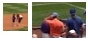}\hspace{3.3cm}
\includegraphics[height=1cm, keepaspectratio]{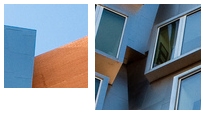}
\end{center}

\begin{tabular}{lll}
\hspace *{1.9cm}\textbf{VPI} & \hspace *{1.8cm} &\hspace *{1.8cm}\\
\hspace *{1.8cm} \textbf{PSNR=62.452} & \hspace *{2.1cm} \textbf{PSNR={$\infty$} } &\hspace *{2.3cm} \textbf{ PSNR=62,536}\\
\hspace *{1.8cm} \textbf{SSIM=1,000} &  \hspace *{2.1cm} \textbf{SSIM=1,000} &\hspace *{2.4cm}  \textbf{SSIM=1,000}\\
\end{tabular}

\caption{Examples of supervised downscaling performance results at the scale factor 2 (left),  at the scale factor 3 (middle), at the scale factor 4 (right). For layout reasons, the performance results both for d-LCI and d-VPI are reported although they coincide for the downscaling factor 3.}
\label{fig:9}

\end{figure*}

We test VPI and the benchmark methods for scale factors varying from 2 to very large values both for supervised and unsupervised mode. In this subsection, some visual results of the numerous performed tests are given. 
\vspace{.2cm}\newline
$\bullet$  Concerning the supervised mode:
\vspace{.2cm}\newline
 we show some examples of  performance results in Figures \ref{fig:6}-\ref{fig:7} for upscaling and in Figures \ref{fig:8}-\ref{fig:9} for downscaling with different BIC input images and scale factors 2, 3, 4. In these figures, some Regions of Interest (ROI) are shown in order to highlight the results at a perceptual level. The visual inspection of these performance results confirms the quantitative evaluation in terms of PSSNR and SSIM exhibited in Subsection \ref{QE}. Hence, we deduce that: a) the observable structure of the objects is captured; b) local contrast and luminance of the input image are preserved; c) small details and most of the salient edges are maintained; d) the presence of ringing and over smoothing artifacts is very limited; e) the resized image is sufficiently not blurred.
\vspace{.2cm}\newline
$\bullet$ Concerning the unsupervised mode:
\vspace{.2cm}\newline
we set, as default, the free parameter $\theta$  equal to 0.5 and take the input images directly from the datasets. Unlike the supervised mode, we cannot compute the PSNR and SSIM quality measures since the target image is missing. Consequently, in the following, we evaluate the performance according to our absolute human perceptual ability taking into account the CPU time (briefly denoted by T) when the results are almost equivalent in terms of perceived quality.

\begin{figure*}[!htbp]

\begin{center}
\includegraphics[height=3.5cm, width=5.1cm]{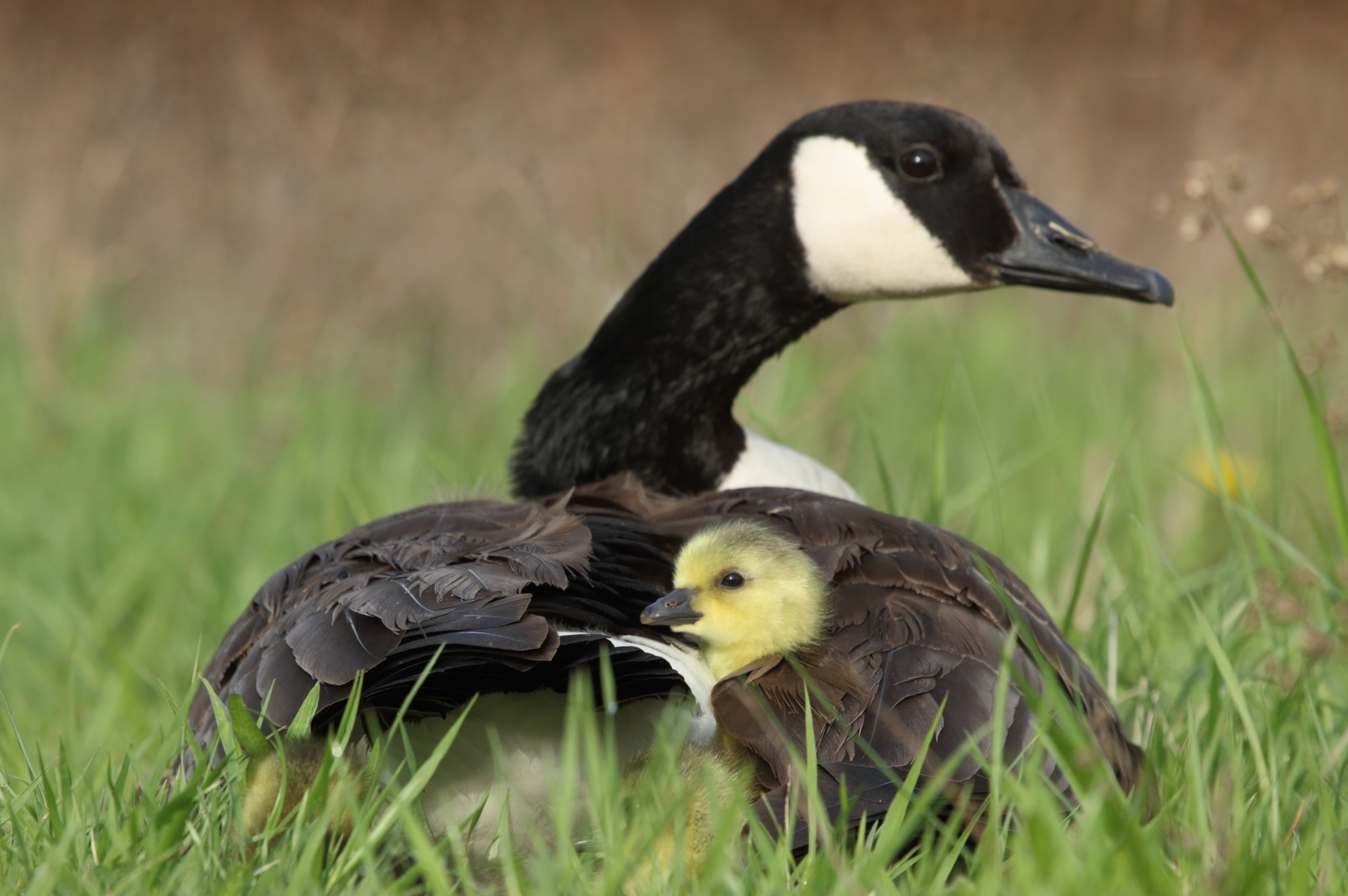}
\hspace *{1cm}
\includegraphics[height=3.5cm, width=5.1cm]{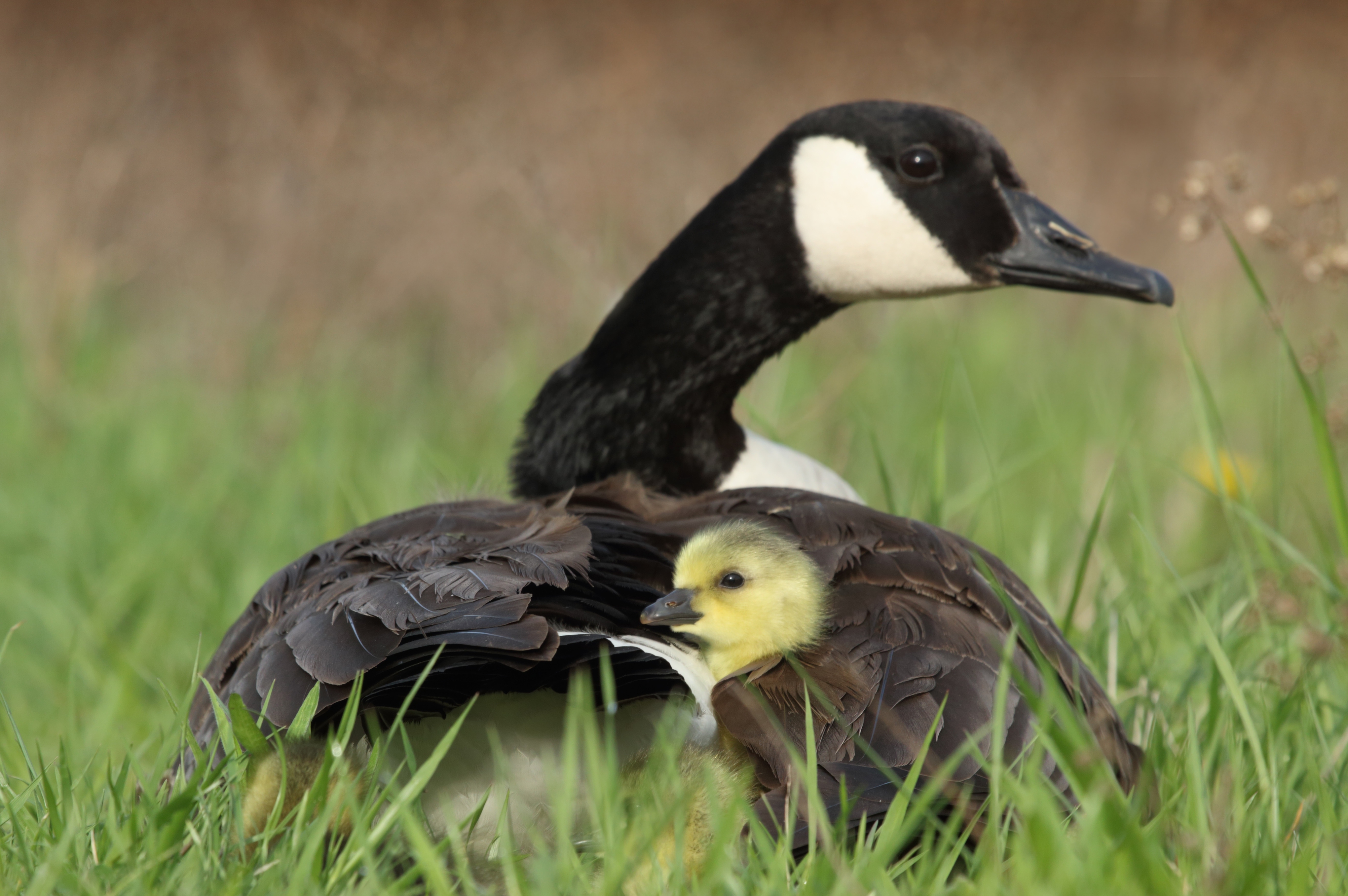}
\end{center}

\begin{tabular}{ll}
\hspace *{3cm} \textbf{BIC (T=0.124)} &\hspace *{3.2cm}  \textbf{SCN (T=42,497)}\\
\end{tabular}

\begin{center}
\includegraphics[height=3.5cm, width=5.1cm]{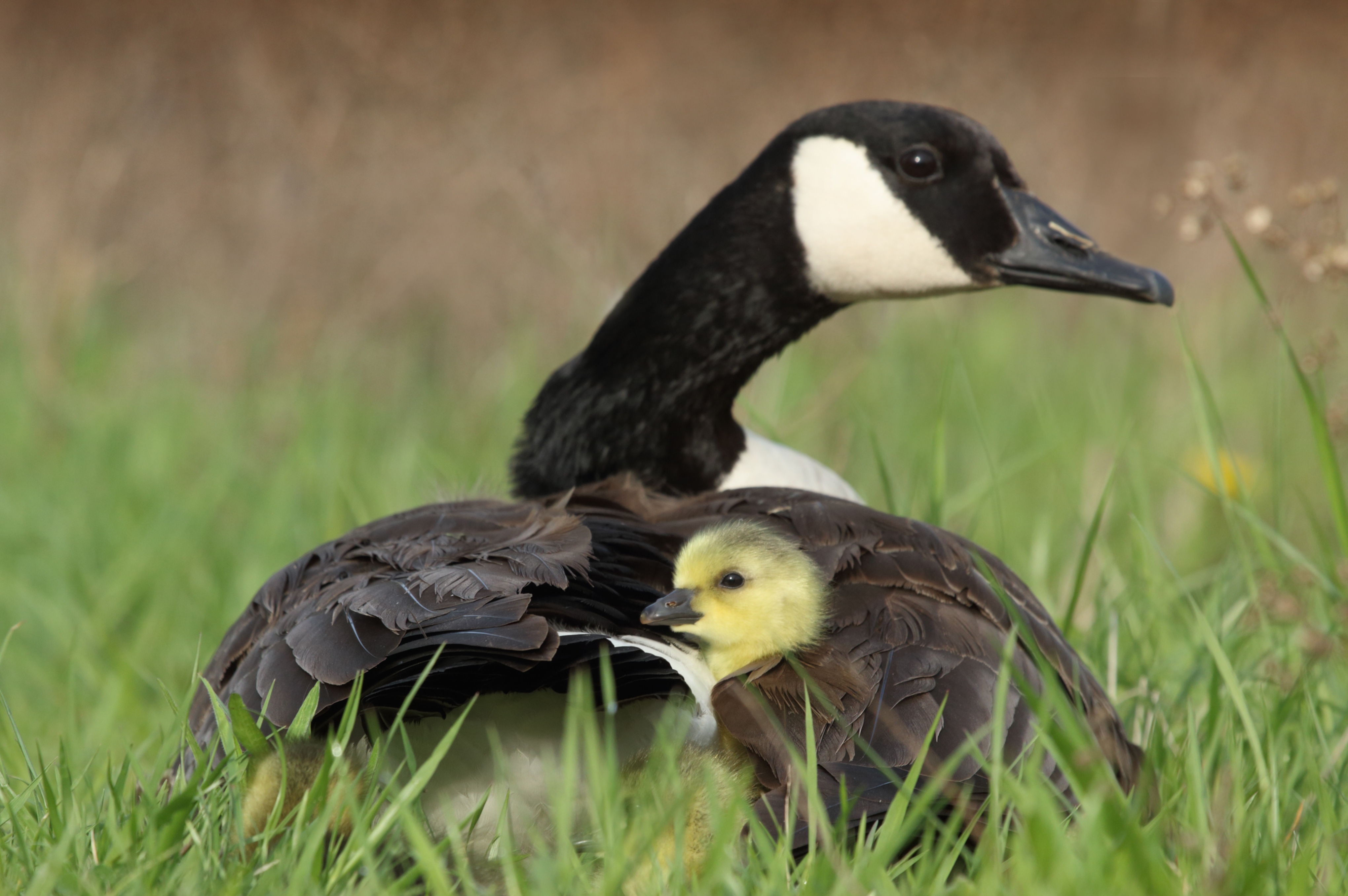}
\hspace *{1cm}
\includegraphics[height=3.5cm, width=5.1cm]{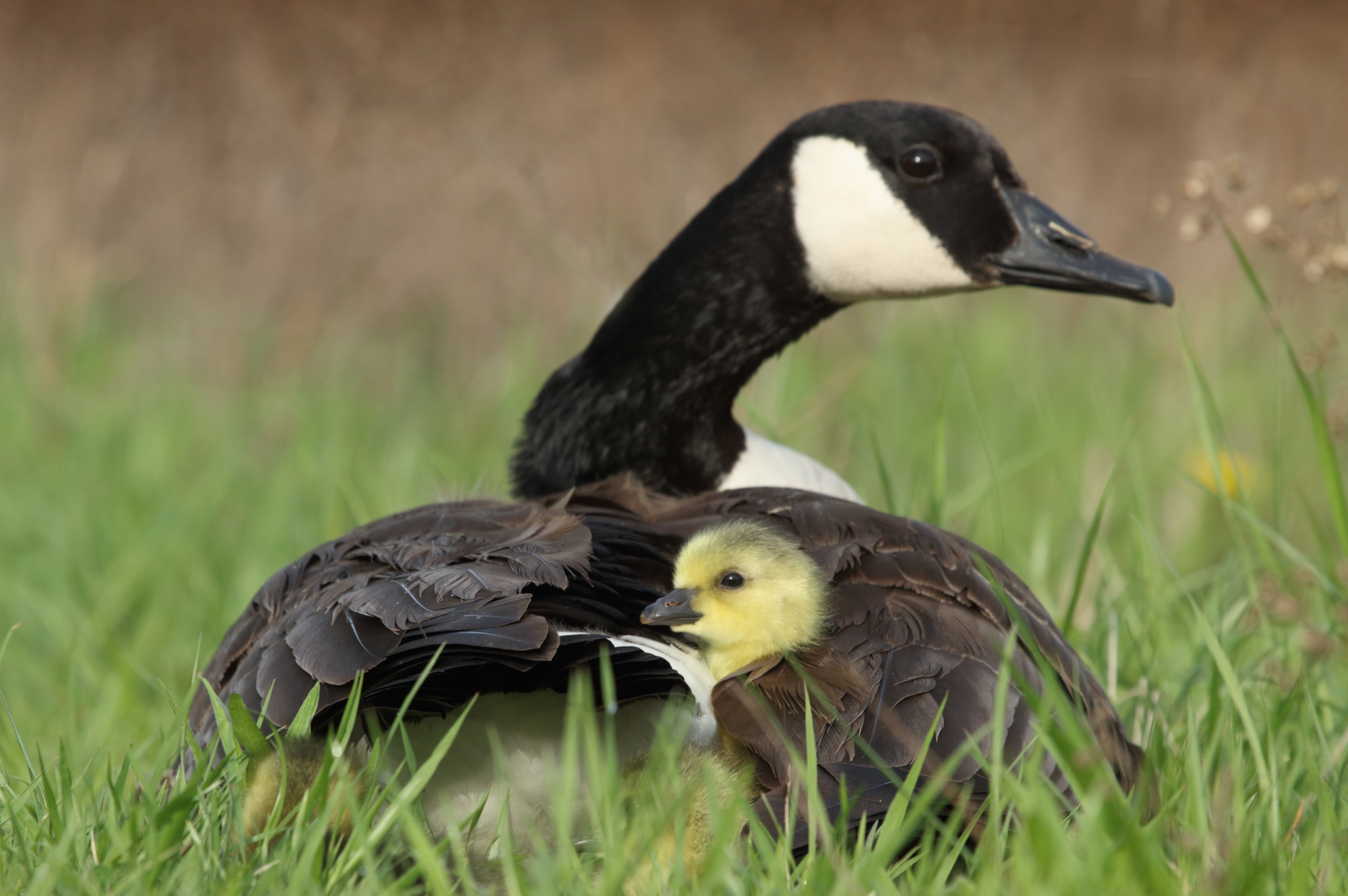}
\end{center}

\begin{tabular}{ll}
\hspace *{3cm} \textbf{u-LCI (T=1.462)} &\hspace *{3cm}  \textbf{u-VPI (T=1,698)}\\
\end{tabular}

\caption{An example of unsupervised upscaling performance results at the scale factor 2 on an image extracted from DIV2k (size=1356$\times$2040). }
\label{fig:10}

\end{figure*}

\begin{figure*}[!htbp]

\begin{center}
\includegraphics[height=3.5cm, width=5.1cm]{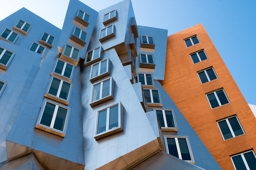}
\hspace *{0.5cm}
\includegraphics[height=3.5cm, width=5.1cm]{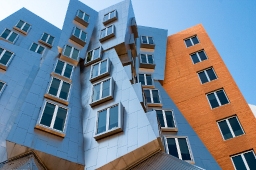}
\hspace *{0.5cm}
\includegraphics[height=3.5cm, width=5.1cm]{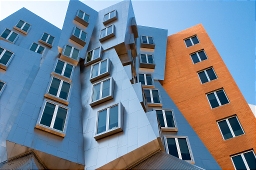}
\end{center}

\begin{tabular}{lll}
\hspace *{0.2cm} \textbf{BIC (T=0,018)} &\hspace *{2.9cm}  \textbf{DPID (T=5,476)}&\hspace *{2.5cm}  \textbf{ L$_0$ (T=2,008)}\\
\end{tabular}

\begin{center}
\includegraphics[height=3.5cm, width=5.1cm]{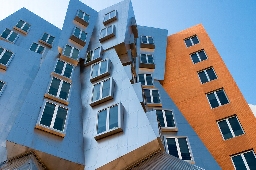}
\hspace *{0.5cm}
\includegraphics[height=3.5cm, width=5.1cm]{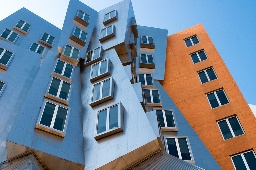}
\hspace *{0.5cm}
\includegraphics[height=3.5cm, width=5.1cm]{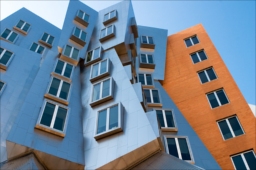}
\end{center}

\begin{tabular}{lll}
\hspace *{0.2cm} \textbf{d-LCI (T=0,032)} &\hspace *{2.8cm}  \textbf{d-VPI (T=0,033)}&\hspace *{2.5cm}  \textbf{f-d-VPI (T=0,133)}\\
\end{tabular}

\caption{An example of unsupervised downscaling performance results at the scale factor 4 on a image extracted from Urban100 (size=680$\times$1024). In f-d-VPI a circular averaging filter with size 3 is employed.}
\label{fig:11}

\end{figure*}

Firstly, we consider as input two images already displayed for the supervised mode and used in that case as target images (see Figure \ref{fig:6} and Figure \ref{fig:8}). We show the performance results at the scale factor 2 (upscaling) in Figure \ref{fig:10} and at the scale factor 4 (downscaling) in Figure \ref{fig:11}. A careful examination of Figure \ref{fig:10} does not highlight significant perceptual visual differences in the upscaled images produced by all methods. However, the required CPU times for u-VPI, u-LCI, and BIC are very close, while SCN takes much more processing time. On the other hand, since the input image in Figure \ref{fig:11} has too high-frequency details, differently from Figure \ref{fig:8} achieved by BIC input image, aliasing effects are visually detectable for all methods. In particular, d-LCI and d-VPI present more evident aliasing effects than the other methods, although with a minor CPU time with respect to L$_0$ and DPID. 

Even if avoiding aliasing effects is an important part of downscaling methods, this is out of the aim of the present paper, which intends to show a different point of view for both upscaling and downscaling by using a specific Approximation Theory tool in the Image Processing framework. Consequently, in Figure \ref{fig:11}, as well as in the remaining experiments, we just show the performance result obtained by a pre-filtering combined with d-VPI (denoted as f-d-VPI). We point out that in f-d-VPI, the type of filter to employ is selected based on the feature images. Our selection includes the following 2-D filters: a) averaging filter ('average'); b) circular averaging filter ('disk'); c) Gaussian filter ('gaussian'); d) motion ('motion'), they all implemented in Matlab using \texttt{hspecial} to specify the filter type.

As mentioned in Section \ref{method}, this solution only partially reduces the aliasing effect in Figure \ref{fig:11}. It does not affect the processing time too much since the CPU time of f-d-VPI  is much smaller than that of L$_0$ and DPID, and very close to the CPU time of BIC. 

Intending to highlight the aliasing influence, in the sequel, we consider different kinds of input images extracted from PEXELS300 dataset (so having the same input size 1800$\times$1800), and we apply to them unsupervised d-VPI and the benchmark methods with the same downscaling factor. 

\begin{figure*}[!htbp]

\begin{tabular}{l}
\hspace *{7.3cm} \textbf{Input images}\\
\end {tabular}

\begin{center}
  \includegraphics[height=5.2cm, width=5.2cm]{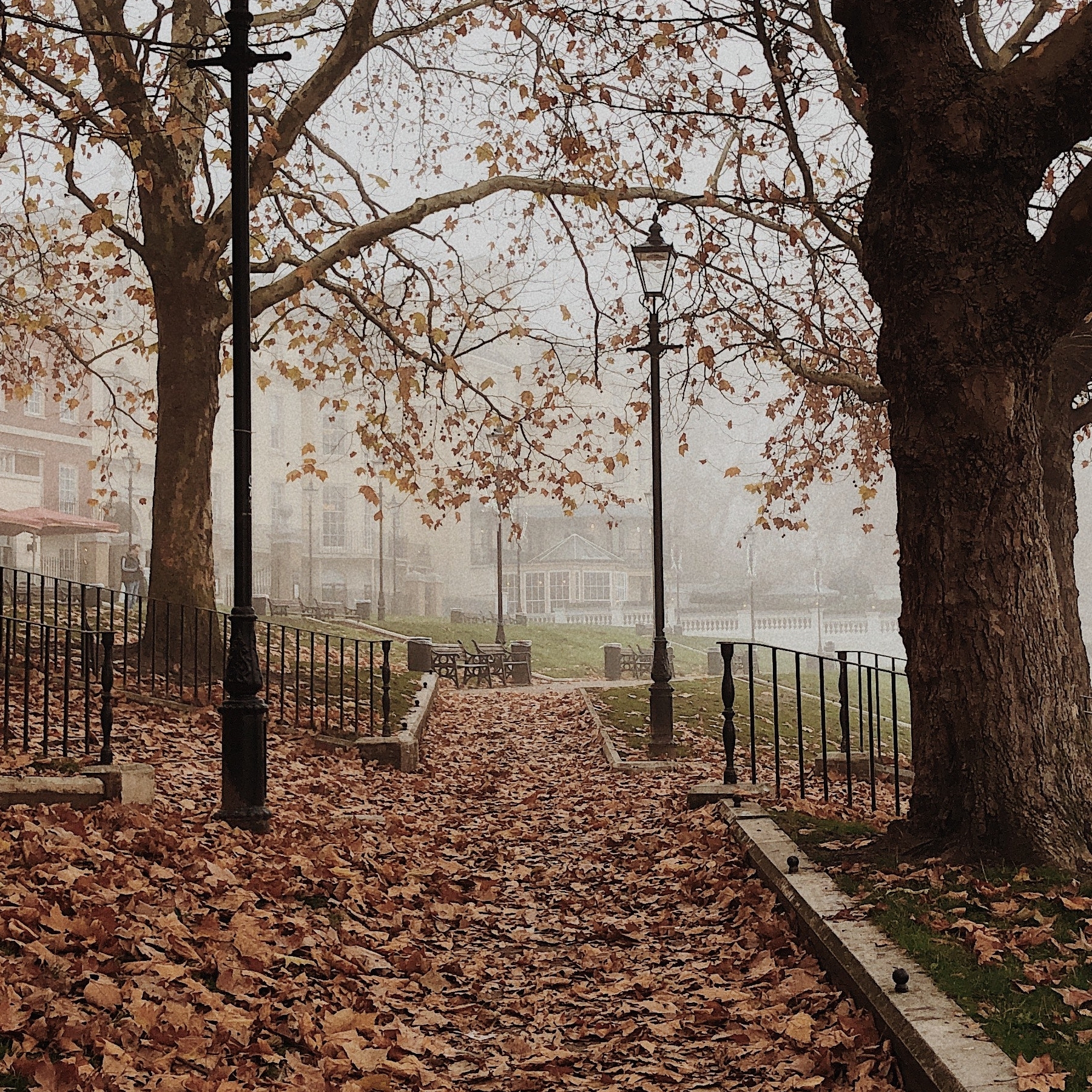}
\hspace *{2cm}
  \includegraphics[height=5.2cm, width=5.2cm]{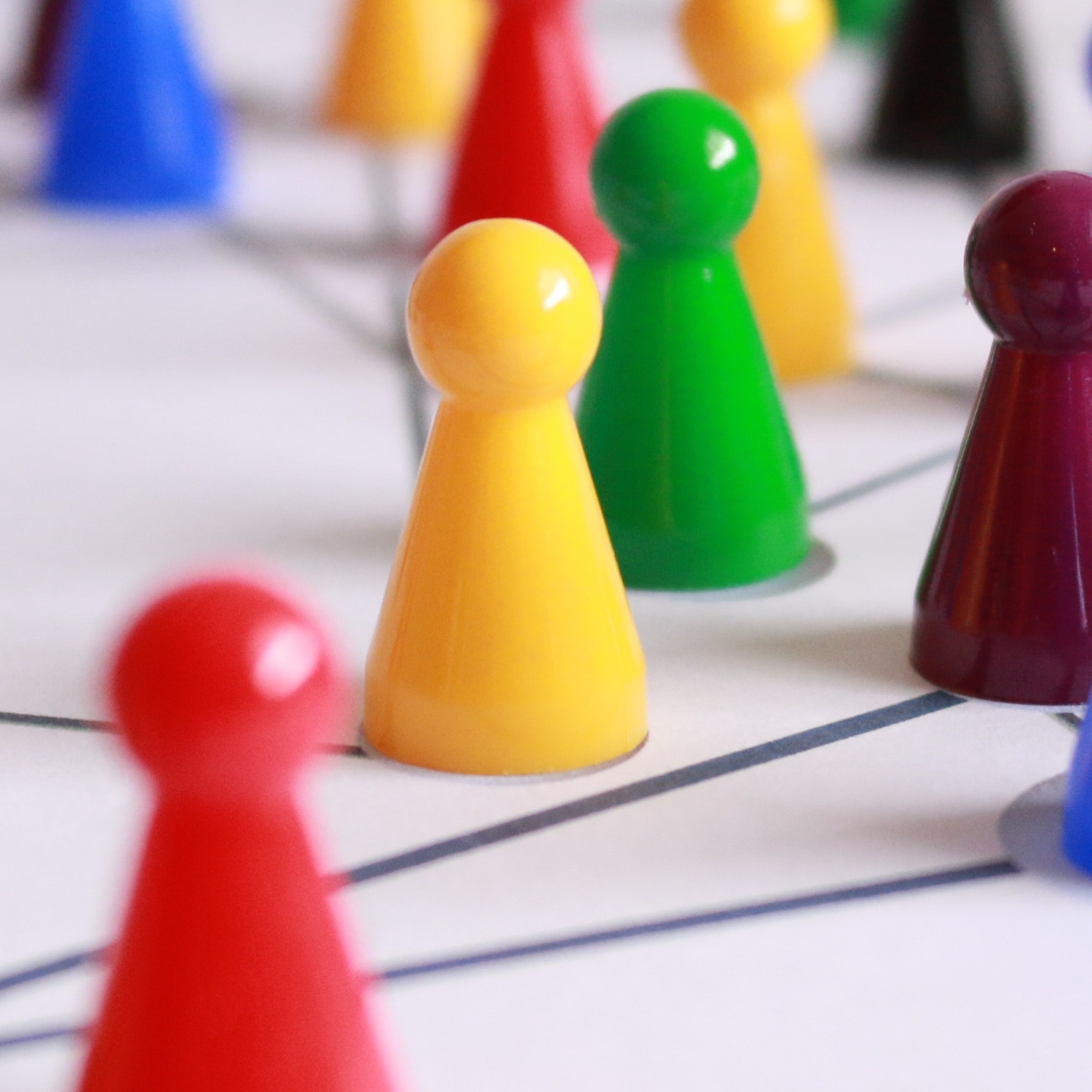}
\end{center}

\begin{tabular}{ll}
\hspace *{4cm} \textbf{1640882} &\hspace *{5.3cm}  \textbf{163064}\\
\newline
\newline
\newline
\end{tabular}

\begin{tabular}{lllll}
\newline
 \hspace *{0.3cm} \textbf{BIC} &\hspace *{2.1cm} \textbf{ DPID}&\hspace *{1.7cm} \textbf{ L$_0$}  &\hspace *{2.3cm} \textbf{ d-VPI $\equiv$ d-LCI}&\hspace *{0.5cm} \textbf{f-d-VPI}\\
\end{tabular}

\begin{center}
\includegraphics[height=3.2cm,width=3.2cm]{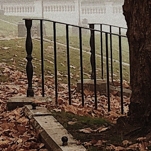}
\includegraphics[height=3.2cm,width=3.2cm]{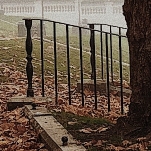}
\includegraphics[height=3.2 cm,width=3.2cm]{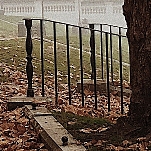}
\includegraphics[height=3.2cm,width=3.2cm]{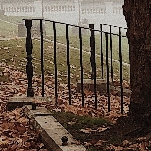}
\includegraphics[height=3.2cm,width=3.2cm]{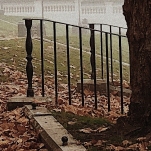}
\end{center}

\begin{tabular}{lllll}
 \hspace *{0.2cm} \textbf{(T=0,029)} &\hspace *{1.4cm} \textbf{(T=30,251)}&\hspace *{1cm} \textbf{(T=12,368)} & \hspace *{1cm} \textbf{(T=0,002)} &\hspace *{1.1cm} \textbf{(T=0,037)}\\
\end{tabular}

\begin{center}
\includegraphics[height=3.2cm,width=3.2cm]{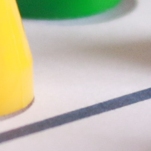}
\includegraphics[height=3.2cm,width=3.2cm]{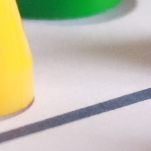}
\includegraphics[height=3.2 cm,width=3.2cm]{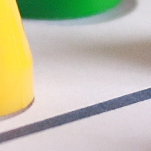}
\includegraphics[height=3.2cm,width=3.2cm]{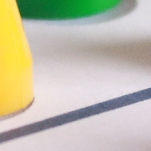}
\includegraphics[height=3.2cm,width=3.2cm]{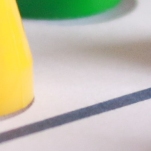}
\end{center}

\begin{tabular}{lllll}
 \hspace *{0.2cm} \textbf{(T=0,090)} &\hspace *{1.4cm} \textbf{(T=30,116)}&\hspace *{1cm} \textbf{(T=13,126)} & \hspace *{1cm} \textbf{(T=0,004)} &\hspace *{1.1cm} \textbf{(T=0,125)}\\
\end{tabular}

\caption{Two input images extracted from PEXELS300 (top) and ROIs of unsupervised downscaling performance results at the scale factor 3  (size: 1800$\times$1800) (middle and bottom). In f-d-VPI an average filter is employed. }
\label{fig:12}
\end{figure*}

In Figure \ref{fig:12} downscaling with scaling factor 3 is applied to the images (1640882 and 163064 from PEXELS300) displayed at the top. Some ROIs of the resulting output images are shown in the middle and bottom in order to emphasize the aliasing phenomenon. By visually inspecting of these ROIs, we can check a different behavior of d-VPI that, in this case, coincides with d-LCI being the LR image computed by (\ref{ih}). Indeed, we note that for the input image on the  right (163064) the aliasing effect is not appreciable (see, for instance, the diagonal line in the ROIs) while it becomes visible for the input image on the left (1640882). In the latter case, we observe aliasing occurs for all methods to a different extent, but BIC,  L$_0$, and DPID have better performance since the downscaled images are affected by aliasing to a lesser extent than d-VPI (see the vertical elements of the railing). However, in the resized image by f-d-VPI, the aliasing effect  is equally present with respect to the other methods without a significant computational burden. It results to be the second-fastest method and is competitive with BIC  (L$_0$ and DPID have a greater CPU time).

\begin{figure*}[!htbp]

\begin{tabular}{l}
\hspace *{7.3cm} \textbf{Input image}\\
\end {tabular}

\begin{center}
  \includegraphics[height=5.2cm, width=5.2cm]{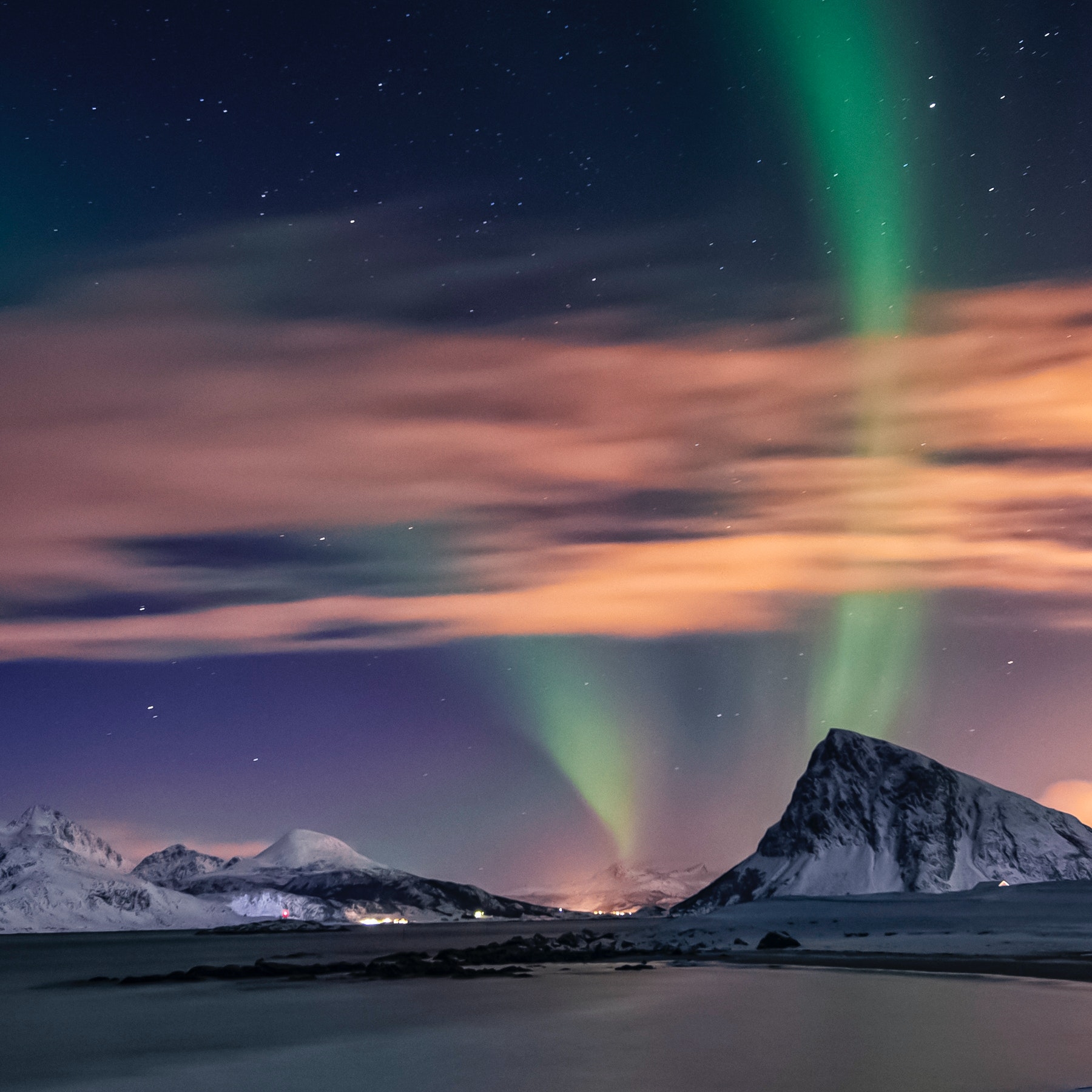}
\end{center}

\begin{tabular}{l}
\hspace *{7.5cm} \textbf{3472764 }\\
\newline
\newline
\end{tabular}

\begin{center}
\includegraphics[height=5.2cm,width=5.2cm]{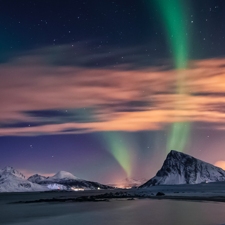}
\includegraphics[height=5.2cm,width=5.2cm]{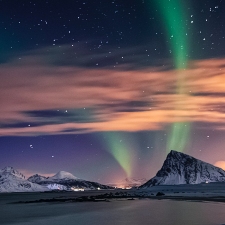}
\includegraphics[height=5.2 cm,width=5.2cm]{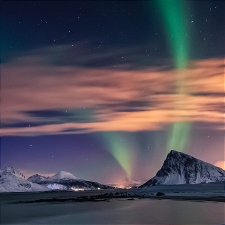}
\end {center}

\begin{tabular}{lll}
\hspace *{0.6cm} \textbf{BIC (T=0,031)} &\hspace *{2.5cm}  \textbf{DPID (T=22.100)}&\hspace *{2.1cm}  \textbf{ L$_0$ (T=12,996)}\\
\end{tabular}

\begin{center}
\includegraphics[height=5.2cm,width=5.2cm]{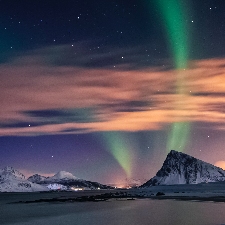}
\includegraphics[height=5.2cm,width=5.2cm]{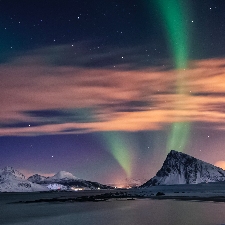}
\includegraphics[height=5.2cm,width=5.2cm]{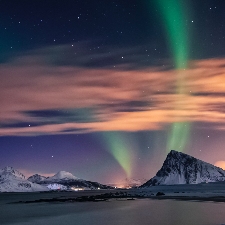}
\end{center}

\begin{tabular}{lll}
\hspace *{0.6cm} \textbf{d-LCI (T=0,091)} &\hspace *{2.5cm}  \textbf{d-VPI (T=0,113)}&\hspace *{2.1cm}  \textbf{f-d-VPI (T=0,137)}\\
\end{tabular}

\caption{Examples of unsupervised downscaling performance results at the scale factor 8 on an input image extracted from PEXELS300 (size 1800$\times$1800). The input image is shown with the same printing size of the resulting images to facilitate the visual comparison. In f-d-VPI an average filter is employed.}
\label{fig:13}
\end{figure*}

Finally, in Figure \ref{fig:13}, we test all downscaling methods at the scale factor 8 on the input image displayed on the top  (3472764 from PEXELS300). In this case, d-VPI and d-LCI produce better visual performance results since the stars in the sky are more adequately preserved in terms of numbers and shape. BIC and  L$_0$ reduce too much the number of stars and introduce a blurring effect, while DPID provides a downscaled image where the stars are almost all reshaped and doubled. Moreover, f-d-VPI seems not to give new insights. Note that  the aliasing is visible in other areas of the image (see, for example, the mountain ridge area) with almost the same intensity for all methods. About CPU time, also in this case, DPID and L$_0$ are the most expensive.

In conclusion, we point out that the aliasing effect does not always occur at the same scale factor and does not always influence the downscaling performance similarly. Moreover, in some contexts, even the downscaling methods designed to reduce the aliasing can result inadequate to manage this problem and can introduce distortions even greater than aliasing itself. In these cases, as well as when the aliasing is not visible and when the quality of the downscaled image is visually equivalent, d-VPI may prove to be preferable since it provides a good compromise in terms of quality and CPU time.


\section{Conclusions}
\label{concl}
This paper proposes a new image scaling method, VPI, which is based on non-uniform sampling grids and employs the filtered de la Vall\'ee Poussin type polynomial interpolation at Chebyshev zeros of 1st kind.

The VPI method is simple to implement and highly flexible since it can be applied to resize arbitrary digital images both in upscaling and downscaling by specifying the scale factor or the desired size.

VPI depends on an additional input parameter $\theta\in [0,1]$ that, if necessary, can be suitably modulated to improve the approximation. In particular, taking $\theta=0$, VPI reduces to the LCI method that has been introduced by the authors in \cite{Lagrange} and is based on classical Lagrange interpolation at the same nodes. Nevertheless, for any $\theta\in ]0,1]$ VPI improves the LCI performance and proves to be more stable than the latter due to the uniform boundedness of Lebesgue constants corresponding to the de la Vall\'ee Poussin type interpolation.

The VPI performance has been evaluated using two commonly adopted quality measures, PSNR and SSIM, and measuring the required CPU time too. Comparisons with other recent resizing methods (also specialized in only upscaling or downscaling) have been carried on a wide number of images belonging to several, commonly available, datasets and characterized by different contents and sizes ranging from small to large scale. During the VPI validation procedure, also the modulation of the free parameter $\theta$ has been observed experimentally. Further, the dependency on the input image has been considered by applying to the target images in the datasets different scaling methods in order to generate the input images.

The experimental results confirm that VPI has a competitive and satisfactory performance, with quality measures generally higher and more stable than those of the benchmark methods. Moreover, VPI results much faster than the methods specialized in only downscaling or upscaling, with CPU time close to the one required by LCI and {\rm imresize}, the Matlab optimized version of bicubic interpolation method (BIC).

At a visual level, VPI captures the object's visual structure by preserving the salient details, the local contrast, and the luminance of the input image, with well--balanced colors and limited presence of artifacts.

One limitation of VPI concerns downscaling performance when HR images have high--frequency details. Indeed, in downscaling with odd scale factors, VPI  produces the same LR image of LCI for any value of $\theta$. In this case, if the Nyquist limit is satisfied, we give a theoretical estimate for the MSE, which, in particular, is null if the input image (or some crucial pixels of it) are "not corrupted". Nevertheless, even starting from "exact" HR images, in downscaling VPI can suffer from aliasing problems when the frequency content of the image and the required size for the LR image are such to violate the Nyquist--Shannon theorem.  In our experiments, we report cases when aliasing does not occur and cases when aliasing occurs. In the latter case, we just apply an appropriate filter to the input image before running d-VPI. However, reducing aliasing effects for d-VPI remains an open problem to further investigations.

\vspace{0.7cm}

\textbf{Funding}

The research has been accomplished within RITA (Research ITalian network on Approximation), and UMI (Unione Matematica Italiana) research groups: TAA\-UMI (Approximation Theory and Applications) and AI\&ML\&MAT\-UMI (Mathematics for Artificial Intelligence and Machine Learning). It has been partially supported by GNCS\-INdAM and University of Basilicata (local funds).

 \textbf{Acknowledgements}
The authors thank the anonymous reviewers for their helpful remarks that allowed to improve the quality of the paper.

\textbf{Code and supplementary materials}

The source code and the dataset PEXELS300 are openly available at the following link:\\
  https://github.com/ImgScaling/VPIscaling.

\end{document}